\DeclareUnicodeCharacter{2212}{-}

\documentclass[journal]{IEEEtran}
\usepackage[utf8]{inputenc}
\usepackage{amsthm}
\usepackage{amssymb}
\usepackage{threeparttable}
\usepackage{amsmath}
\usepackage{verbatim}
\usepackage{tikz}
\usepackage{color}
\usepackage{xcolor}
\usepackage{colortbl}
\usepackage{arydshln}
\usepackage{hyperref,tablefootnote,footnotehyper}

\definecolor{ciano}{rgb}{0.854,0.909,0.988}
\definecolor{cinza}{rgb}{0.7529,0.7529,0.7529}

\usepackage{booktabs}

\definecolor{tabwhite}{rgb}{1,1,1}
\definecolor{taborange}{rgb}{0.92,0.6,0.46}
\definecolor{tabgrey}{rgb}{0.4,0.4,0.4}

\def\BibTeX{{\rm B\kern-.05em{\sc i\kern-.025em b}\kern-.08emT\kern-.1667em\lower.7ex\hbox{E}\kern-.125emX}}
    
\begin{document}

\title{Neural Attention Models in Deep Learning: Survey and Taxonomy}



\author{Alana~de~Santana~Correia,
        and~Esther~Luna~Colombini,~\IEEEmembership{Member,~Brazil}
\thanks{Laboratory of Robotics and Cogntive Systems (LaRoCS)
Institute of Computing, University of Campinas, Av. Albert Einstein, 1251 - Campinas, SP - Brazil e-mail: \{alana.correia, esther\}@ic.unicamp.br.}
}


\maketitle

\begin{abstract}
Attention is a state of arousal capable of dealing with limited processing bottlenecks in human beings by focusing selectively on one piece of information while ignoring other perceptible information \cite{colombini2014attentional}. For decades, concepts and functions of attention have been studied in philosophy, psychology, neuroscience, and computing. Currently, this property has been widely explored in deep neural networks. Many different neural attention models are now available and have been a very active research area over the past six years. From the theoretical standpoint of attention, this survey provides a critical analysis of major neural attention models. Here we propose a taxonomy that corroborates with theoretical aspects that predate Deep Learning. Our taxonomy provides an organizational structure that asks new questions and structures the understanding of existing attentional mechanisms. In particular, 17 criteria derived from psychology and neuroscience classic studies are formulated for qualitative comparison and critical analysis on the 51 main models found on a set of more than 650 papers analyzed. Also, we highlight several theoretical issues that have not yet been explored, including discussions about biological plausibility, highlight current research trends, and provide insights for the future.
\end{abstract}

\begin{IEEEkeywords}
Survey, Taxonomy, Attention Mechanism, Neural Networks, Deep Learning, Attention Models.
\end{IEEEkeywords}

\IEEEpeerreviewmaketitle

%

%
\maketitle

\section{Introduction}
\label{sec:introduction}

Attention is a state of arousal capable of dealing with limited processing bottlenecks by focusing selectively on one piece of information while ignoring other perceptible information \cite{colombini2014attentional}. According to James \cite{James1890JAMPOP} attention can be considered as an internal force that spontaneously or voluntarily creates a mental expectation of a sensory or motor nature, favoring the perception of stimuli and the production of responses. This internal force can also be understood as a cognitive need, given that at any moment, the environment presents more perceptual information than can be supported, and it is impossible to perform all motor actions simultaneously in response to all external stimuli. In nature, attention is an essential activity concerning the survival of all forms of life, resulting from a long process of cognitive evolution of living beings. Among the beings that occupy the lowest evolutionary scale positions, attention acts mainly on perception, selecting, and modulating relevant stimuli from the environment. This mechanism is decisive for the perpetuation and evolution of species, as it is characterized as the ability to settle in points of interest in the environment and recognize possible prey, predators, or rivals. In humans, attention is intrinsically present in the brain throughout the cognitive cycle, acting from the perception of stimuli, organization of complex mental processes to decision making.

For decades, several areas of science have been concerned with understanding the role of attention. In psychology, studies dating back to 1890, looking for behavioral correlates that reflect the performance of attentional processes in the human brain, such as surveillance time \cite{mackworth1948breakdown}, inattentional blindness \cite{simons1999gorillas}, attentional blink \cite{raymond1992temporary}, reaction time in cognitive processing \cite{sanders1998elements}, and the selective ability to filter external stimuli \cite{broadbent2013perception}. Cognitive neuroscience studies have employed invasive and non-invasive approaches, such as neuroanatomical/neurophysiological techniques, electroencephalography, positron emission tomography (PET), and functional magnetic resonance imaging (fMRI), to capture insights about attentional disorders \cite{driver2001neurobiological}. Neurophysiologists seek to study how neurons respond to represent external stimuli of interest \cite{treue2001neural}. Finally, computational neuroscientists capture all the insights from the different perspectives experienced and support realistic computational models to simulate and explain attentional behaviors, seeking to understand how, where and when attention processes occur or are needed \cite{rolls2006attention}.

Inspired by these studies, computer scientists in the 1990s proposed the first attentional mechanisms for computer systems to resolve performance limitations inherent in the high computational complexity of the algorithms existing at the time. Initially, several attentional vision models for object recognition \cite{salah2002selective}, image compression \cite{ouerhani2004visual}, image matching \cite{walther2006interactions}, image segmentation \cite{ouerhani2004visual}, object tracking \cite{walther2004detection}, active vision \cite{clark1988modal}, and recognition \cite{salah2002selective} emerged inspired by Feature Integration Theory - one of the first theories to formalize visual attention to perception - in which a set of simple features is extracted from a scene observed by the system separately, and in subsequent steps the integration of the stimuli occurs supporting the identification of relevant objects in the environment. Subsequently, visual attention emerged as a tool capable of providing essential information on the environment for robotic agents' decision-making in the world. So, several robotic navigation systems \cite{baluja1997expectation}, SLAM\cite{frintrop2008attentional}, and and human-robot interaction \cite{breazeal1999context} integrated attention to improve the performance of these autonomous agents.

Artificial intelligence scientists have noticed attention as a fundamental concept for improving deep neural networks' performance in the last decade. In Deep Learning, Attention introduces a new form of computing inspired by the human brain quite different from what neural networks do today. Attentional mechanisms make networks more scalable, simpler, facilitate multimodality, and reduce information bottlenecks from long spatial and temporal dependencies. Attentional interfaces currently focus on two major fronts of development and research as small modules that can be easily plugged into classic DL architectures and end-to-end attention networks where attention is intrinsically present throughout the architecture. The attentional interfaces usually complement convolutional and recurrence operations allowing the control of the dynamic flow of resources and internal or external information, coming from specific parts of the neural structure or other external cognitive elements (e.g., external memories, pre-trained layers). End-to-end attention networks represent major advances in Deep Learning. State-of-art approaches in Natural Language Processing \cite{vaswani_attention_2017} \cite{sukhbaatar2015end}, multimodal learning, and unstructured data learning via graph neural networks use end-to-end attention approaches \cite{velickovic_graph_2018}\cite{zhang2019hyperbolic}. Currently, much of the research aimed at DL uses attentional structures in the most diverse application domains, so that we have been able to map more than 6,000 works published in the area since 2014 in the main publication repositories.


Despite the high extent of research in various areas of computing, psychology, neuroscience, and even philosophy, the historical problem is that attention is omnipresent in the brain, as there is no single attention center, which makes concepts and aspects of study quite abstract and difficult to validate. When a group of related concepts and ideas becomes challenging to manage, a taxonomy is useful. Through taxonomy, it is possible to group different aspects and systematically study them. In psychology and neuroscience, this problem is still present, but there are already several theories and taxonomies widely accepted by several cognitive-behavioral researchers \cite{chun2011taxonomy}.

Specifically, in Deep Learning, there is no a taxonomy based on theoretical concepts of attention, given that the few that exist are far from theoretical concepts and are quite specific in a given scope \cite{chaudhari2019attentive}\cite{galassi2020attention}. In this sense, a unified attention framework supported by a taxonomy with concepts founded on psychology and neuroscience is necessary to elucidate how the different attentional mechanisms act in DL, facilitating the visualization of new research opportunities. Thus, we aim in this work to present the reader with a taxonomy of attention for neural networks based on various theoretical insights on the attention and several relevant researches that precedes Deep Learning \cite{colombini2014attentional}. We formulated a very broad and generic taxonomy around five main dimensions: components (Section \ref{sec:components}), function of the process (Section \ref{sec:selectiveperception}), stimulus' nature (Section \ref{sec:space_based}), process' nature according to the stimulus (Section \ref{sec:bottom-up}), and continuity (Section \ref{sec:hard_soft}).

\subsection{Contributions}

This survey presents a taxonomy that corroborates theoretical aspects of attention that predate Deep Learning. Based on our taxonomy, we present and discuss the main neural attention models in the field.

As the main contributions of our work, we highlight:

\begin{enumerate}
    \item We proposed a unified attention framework;
    \item We introduced the first survey with a taxonomy based on theoretical concepts of attention;
    \item We systematically review the main neural attention models and provide all of our search and filtering tools to help researchers in future searches on the topic \footnote{Download link: \url{https://github.com/larocs/attention_dl}};   
    \item We select the main works through exhaustive search and filtering techniques and choose the most relevant among more than 650 papers critically analyzed;
    \item Based on our taxonomy, we discuss the biological plausibility of the main attentional mechanisms; 
    \item Based on the concepts presented in our taxonomy, we describe in detail the main attentional systems of the field;
    \item Finally, we present a broad description of trends and research opportunities.
\end{enumerate}

\subsection{Organization}
This survey is structured as follows. In Section ~\ref{sec:concept} we present the concept of attention. Section \ref{sec:attention_model} contains an unified attention model. In section ~\ref{sec:our_taxonomy} we introduce and discuss our taxonomy. Section \ref{sub:main_architectures}, we present the main architectures and discuss the central models from our taxonomy perspective. Finally, in Section ~\ref{sec:trends} we discuss limitations, open challenges, current trends, and future directions in the area, concluding our work in section ~\ref{sec:conclusion}.

\section{The concept of attention}
\label{sec:concept}

Attention is difficult to define formally and universally. In psychology, Attention is the act or state of attending, primarily through applying the mind to an object of sense or thought or even as a condition of readiness that selectively narrows or focuses consciousness and receptivity \cite{colombini2014attentional}. However, from a Deep Learning standpoint, a clear definition of attention is needed. In our definition, \textbf{Attention is a system composed of one or multiple modules, which allocate structural or temporal resources, select or modulate signals to perform a task}. Each module consists of a function or multiple non-linear functions trained in conjunction with the neural network. Specifically, each module outputs a selective or modulating mask for an input signal. The structural resources allocated are elements of the architecture (e.g., number of neurons, number of layers), time resources refer to computation per step, number of time steps, processing time in modules of the architectures or frameworks. The task is the goal application (e.g., classification, regression, segmentation, object recognition, control, among others), and signals are given at any abstraction level (e.g., features, visual information, audio, text, memories, latent space vectors). Our definition is supported by the following sections of this work, mainly for our taxonomy (Section ~\ref{sec:our_taxonomy}).


\section{An unified Attention Framework}
\label{sec:attention_model}

\begin{figure}[htb]
    \centering
    \includegraphics[width=\linewidth]{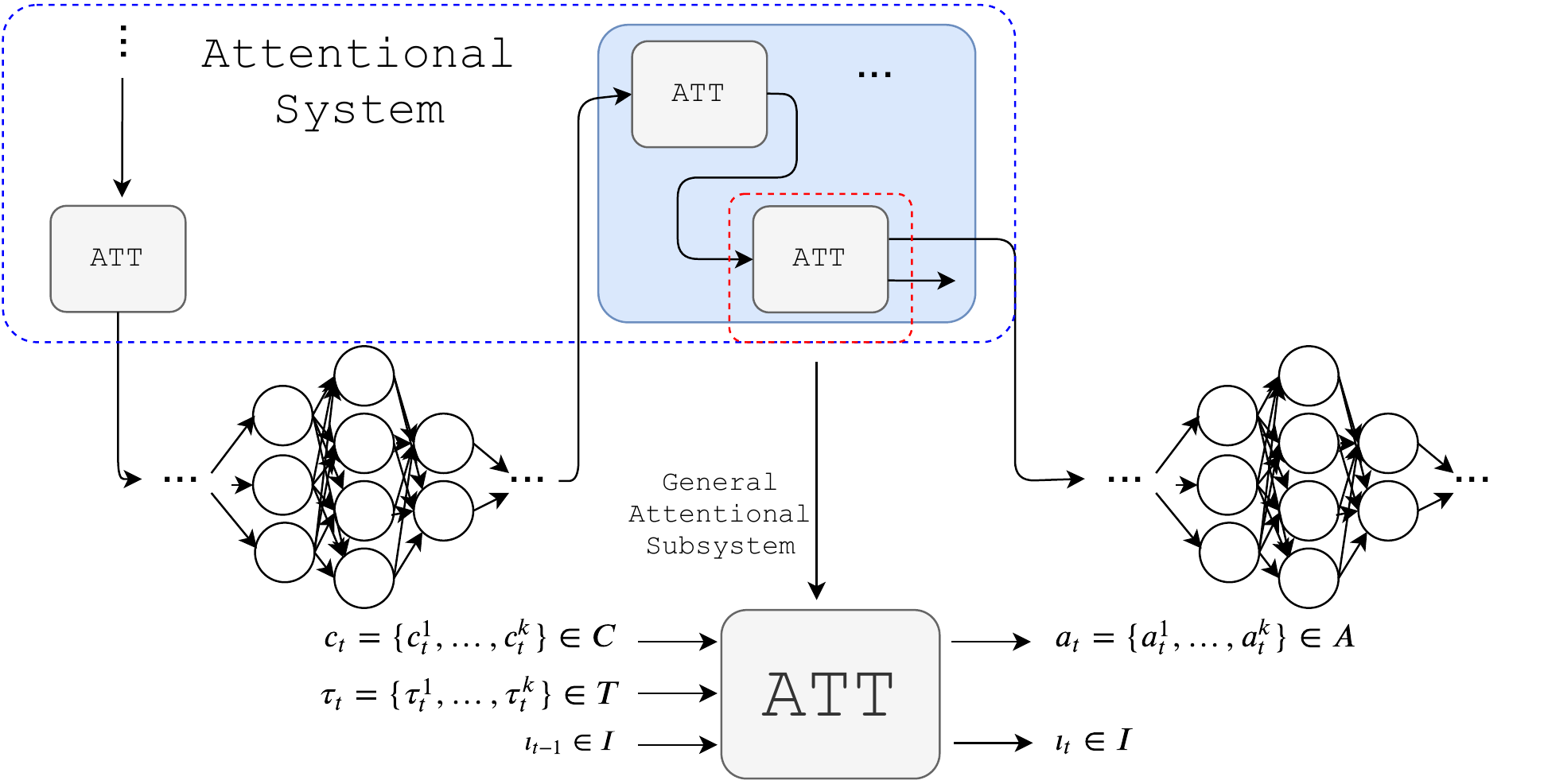}
    \caption{Illustration of our attentional framework for DL, in which several attentional subsystems are coupled in the neural networks sequentially or recurrently. Each subsystem has a different profile based on the input data's structure and sensory modality. A single subsystem receives as the primary input the focus target (i.e., the stimulus to be filtered), and sometimes auxiliary inputs (e.g., contextual information and subsystem's previous internal state) to help the mechanism guide the focus in time.}
    \label{fig:att_module}
\end{figure}

In this section, we define a general and unified model of attention. Our model corroborates with theoretical aspects and is independent of the architecture and the application domain. Specifically, we consider that \textbf{an attentional system contains a set of attentional subsystems - even in a recursive manner - to allocate resources for processes}. An attentional subsystem, at each time step $ t $, receives as \textbf{input} a \textbf{contextual input} $c_{t}$, a \textbf{focus target} $\tau_{t}$, and \textbf{past inner state} $i_{t-1}$. And produces as output a \textbf{current inner state} $i_{t}$, and \textbf{current focus output} $ a_{t} $, as shown Figure \ref{fig:att_module}. The focused output is the main element of the subsystem because it assigns targets an importance score. Together, several attentional subsystems always perform actions to provide selection capabilities. The subsystem profile depends on the data structure and the desired output. We propose a general structure with some additional components that, although not universally present, are still found in most models in the literature. In Figure \ref{tab:notation_att_module}, we list all key components.

\begin{figure*}[!htb]
\begin{center}

\small
\fontfamily{pag}\selectfont
\setlength\arrayrulewidth{0.9pt}

\begin{tabular}{ll|}
\hline
\multicolumn{1}{|c|}{\cellcolor{ciano}\textbf{Symbol}} & \cellcolor{cinza}\textbf{Description}  \\ \hline \hline
\multicolumn{2}{l}{Context}                    \\ \hline
\multicolumn{1}{|c;{0.7pt/1pt}}{\cellcolor{ciano}$k$}               & \multicolumn{1}{l|}{\cellcolor{cinza}Sensory modality index.} \\ 
\multicolumn{1}{|c;{0.7pt/1pt}}{\cellcolor{ciano}$C$}               & \multicolumn{1}{l|}{\begin{tabular}[c]{@{}c@{}}\cellcolor{cinza}Contextual input set, $C = \{c_{t-1}, \ldots, c_{t}\}$, $C \in \mathbb{R}$, (e.g., hidden states, memory data, sensory data).\end{tabular}} \\ 
\multicolumn{1}{|c;{0.7pt/1pt}}{\cellcolor{ciano}$c_{t}$}               & \multicolumn{1}{l|}{\cellcolor{cinza}\begin{tabular}[c]{@{}l@{}}Contextual input at time $t$, $c_{t} = \{c_{t}^{1}, \ldots, c_{t}^{k}\} $, $c_{t} \in C$.\end{tabular}} \\
\multicolumn{1}{|c;{0.7pt/1pt}}{\cellcolor{ciano}$c^{k}_{t}$}                & \multicolumn{1}{l|}{\cellcolor{cinza}\begin{tabular}[c]{@{}l@{}}Contextual input from sensory modality $k$ at time $t$, $c_{t}^{k} = \{c_{t,1}^{k}, \ldots, c_{t,n_{ck}}^{k}\}$, $c_{t,j}^{k} \in \mathbb{R}^{F_{c}}$, where $F_{c}$ is\\amount of features.\end{tabular} } \\ \hline
\multicolumn{2}{l}{Focus target}                          \\ \hline
\multicolumn{1}{|c;{0.7pt/1pt}}{\cellcolor{ciano}$T$}                & \multicolumn{1}{l|}{\cellcolor{cinza}\begin{tabular}[c]{@{}c@{}}\cellcolor{cinza}Focus target set, $T = \{\tau_{t-1}, \ldots, \tau_{t}\}$, $T \in \mathbb{R}$.\end{tabular}} \\ 
\multicolumn{1}{|c;{0.7pt/1pt}}{\cellcolor{ciano}$\tau_{t}$}                & \multicolumn{1}{l|}{\cellcolor{cinza}\begin{tabular}[c]{@{}l@{}}\cellcolor{cinza}Focus target at time $t$, \cellcolor{cinza}$\tau_{t} = \{\tau_{t}^{1}, \ldots, \tau_{t}^{k}\}$, $\tau_{t} \in T$.\end{tabular}} \\ 
\multicolumn{1}{|c;{0.7pt/1pt}}{\cellcolor{ciano}$\tau_{t}^{k}$ }                & \multicolumn{1}{l|}{\cellcolor{cinza}\begin{tabular}[c]{@{}l@{}}Focus target from sensory modality $k$ \\\cellcolor{cinza}Features for $n_{\tau k}$ elements, if $\tau_{t}^{k}$ is a data, Hyperparameters or index, if $\tau_{t}^{k}$ is a program. \\$\tau_{t}^{k} = \{\tau_{t,1}^{k}, \ldots, \tau_{t,n_{\tau k}}^{k}\}$, $\tau_{t,j}^{k} \in $ $\mathbb{R}^{F_{\tau k}}$, where $F_{\tau k}$ is amount of features. \end{tabular}} \\ \hline
\multicolumn{2}{l}{Inner state}                           \\ \hline
\multicolumn{1}{|c;{0.7pt/1pt}}{\cellcolor{ciano}$I$}                & \multicolumn{1}{l|}{\cellcolor{cinza}\begin{tabular}[c]{@{}l@{}} Inner state set, $I = \{i_{t-1}, \ldots, i_{t}\}$, $I \in \mathbb{R}$. \end{tabular}}\\ 
\multicolumn{1}{|c;{0.7pt/1pt}}{\cellcolor{ciano}$i_t$}                & \multicolumn{1}{l|}{\cellcolor{cinza}Inner state at time $t$, $\iota_{t} \in I$.} \\ 
\multicolumn{1}{|c;{0.7pt/1pt}}{\cellcolor{ciano}$i_t-1$}                & \multicolumn{1}{l|}{\cellcolor{cinza}Past inner state at time $t-1$, $\iota_{t-1} \in I$.} \\ \hline
\multicolumn{2}{l}{Focus output}                          \\ \hline
\multicolumn{1}{|c;{0.7pt/1pt}}{\cellcolor{ciano}$A$}                & \multicolumn{1}{l|}{\cellcolor{cinza}\begin{tabular}[c]{@{}l@{}}Focus output set, $A = \{a_{t-1}, \ldots, a_{t}\}$, $A = \left \{ x \in \mathbb{R}: 0 < x < 1\right \}$ or $A = \left \{ x \in \mathbb{Z}: 0 \leq x \leq 1\right \}$.\end{tabular}} \\
\multicolumn{1}{|c;{0.7pt/1pt}}{\cellcolor{ciano}$a_{t}$}                & \multicolumn{1}{l|}{\cellcolor{cinza}\begin{tabular}[c]{@{}c@{}}Focus output at time $t$, $a_{t} = \{a_{t}^{1}, \ldots, a_{t}^{k}\} \in A$.\end{tabular}} \\ 
\multicolumn{1}{|c;{0.7pt/1pt}}{\cellcolor{ciano}$a_{t}^{k}$}                & \multicolumn{1}{l|}{\cellcolor{cinza}\begin{tabular}[c]{@{}l@{}}Focus output from sensory modality $k$ at time $t$, $a_{t}^{k} = \{a_{t,1}^{k}, \ldots, a_{t,n_{\tau k}}^{k}\}$ are attention scores,\\
$a_{t,j}^{k} \in \mathbb{R}^{F_{\tau k}}$ or $a_{t,j}^{k} \in \mathbb{R}$, $a_{t}^{k} \in \mathbb{R}^{n_{\tau k} \times F_{\tau k}}$ or $a_{t}^{k} \in \mathbb{R}^{n_{\tau k}}$.\end{tabular}} \\ \hline
\end{tabular}
\end{center}
\caption{Notation for unified attention model. Note the notation supports recurrence and multimodality.}
\label{tab:notation_att_module}
\end{figure*}

\section{A Taxonomy for Attention Models}
\label{sec:our_taxonomy}
This section introduces our taxonomy around 16 factors that will be used to categorize and discuss the main neural attention models summarized in Figure \ref{table:main_models}. These factors have their origins in behavioral and computational studies of attention. In section \ref{sec:components} we present the models of attention from the component perspective, whereas in section \ref{sec:selectiveperception} we discuss the process function presenting mechanisms of perceptual and cognitive selection. Section \ref{sec:space_based} presents the mechanisms according to the nature of the stimulus, while in section \ref{sec:bottom-up} we discuss the mechanisms analyzing the nature of the process according to the stimulus. Finally, in section \ref{sec:hard_soft} we present the mechanisms from the continuity standpoint.

\begin{figure*}[!htb]
\begin{center}
\small
\fontfamily{pag}\selectfont
\setlength\arrayrulewidth{0.9pt}

\begin{tabular}{p{2mm}p{38mm}p{2mm}p{2mm}p{2mm}p{2mm}p{2mm}p{2mm}p{2mm}p{6mm}p{2mm}p{2mm}p{2.4mm}p{2mm}p{2.4mm}p{2mm}p{2.4mm}p{2mm}}
\hline
\multicolumn{1}{|c|}{\textbf{\cellcolor{ciano}No}} & \cellcolor{ciano}\textbf{Model}    & \multicolumn{1}{r|}{\textbf{\cellcolor{ciano}Year}} & \textbf{\cellcolor{ciano}f1} & \textbf{\cellcolor{ciano}f2}  & \textbf{\cellcolor{ciano}f3} & \multicolumn{1}{c|}{\textbf{\cellcolor{ciano}f4}} & \textbf{\cellcolor{ciano}f5} & \multicolumn{1}{c|}{\textbf{\cellcolor{ciano}f6}} & \textbf{\cellcolor{ciano}f7} & \textbf{\cellcolor{ciano}f8} & \multicolumn{1}{c|}{\textbf{\cellcolor{ciano}f9}} & \textbf{\cellcolor{ciano}f10} & \multicolumn{1}{c|}{\textbf{\cellcolor{ciano}f11}} & \textbf{\cellcolor{ciano}f12} & \multicolumn{1}{c|}{\textbf{\cellcolor{ciano}f13}} & \textbf{\cellcolor{cinza}f14} & \multicolumn{1}{c|}{\textbf{\cellcolor{cinza}f15}} \\ \hline \hline

\multicolumn{15}{l}{\textbf{Bottom-up}} \\ \hline

\multicolumn{1}{|c|}{\cellcolor{ciano}\textbf{1}} & \cellcolor{ciano}STN \cite{jaderberg2015spatial} & \multicolumn{1}{r;{0.7pt/1pt}}{\textbf{\cellcolor{ciano}2015}} & \cellcolor{ciano}+ & \cellcolor{ciano}- & \cellcolor{ciano}- &  \multicolumn{1}{c;{0.7pt/1pt}}{\cellcolor{ciano}-} & \cellcolor{ciano}- & \multicolumn{1}{c;{0.7pt/1pt}}{\cellcolor{ciano}+} & \cellcolor{ciano}\textbf{FM} & \cellcolor{ciano}- & \multicolumn{1}{c;{0.7pt/1pt}}{\cellcolor{ciano}-} & \cellcolor{ciano}- & \multicolumn{1}{c;{0.7pt/1pt}}{\cellcolor{ciano}-} & \cellcolor{ciano}+ & \multicolumn{1}{c|}{\cellcolor{ciano}-} & \cellcolor{cinza}+ & \multicolumn{1}{c|}{\cellcolor{cinza}-}\\
\multicolumn{1}{|c|}{\cellcolor{ciano}\textbf{2}} & \cellcolor{ciano}Chen et al. \cite{chen2016latent} & \multicolumn{1}{r;{0.7pt/1pt}}{\textbf{\cellcolor{ciano}2016}} & \cellcolor{ciano}- & \cellcolor{ciano}+ & \cellcolor{ciano} -& \multicolumn{1}{c;{0.7pt/1pt}}{\cellcolor{ciano}-} & \cellcolor{ciano}- & \multicolumn{1}{c;{0.7pt/1pt}}{\cellcolor{ciano}+} & \cellcolor{ciano}- & \cellcolor{ciano}- & \multicolumn{1}{c;{0.7pt/1pt}}{\cellcolor{ciano}\textbf{L}} & \cellcolor{ciano}- & \multicolumn{1}{c;{0.7pt/1pt}}{\cellcolor{ciano}-} & \cellcolor{ciano}- & \multicolumn{1}{c|}{\cellcolor{ciano}+} & \cellcolor{cinza}- & \multicolumn{1}{c|}{\cellcolor{cinza}+} \\
\multicolumn{1}{|c|}{\cellcolor{ciano}\textbf{3}} & \cellcolor{ciano}AT \cite{zagoruyko2016paying} & \multicolumn{1}{r;{0.7pt/1pt}}{\textbf{\cellcolor{ciano}2016}} & \cellcolor{ciano}- & \cellcolor{ciano}+ & \cellcolor{ciano}- & \multicolumn{1}{c;{0.7pt/1pt}}{\cellcolor{ciano}-} & \cellcolor{ciano}- & \multicolumn{1}{c;{0.7pt/1pt}}{\cellcolor{ciano}+} & \cellcolor{ciano}\textbf{FM} & \cellcolor{ciano}- & \multicolumn{1}{c;{0.7pt/1pt}}{\cellcolor{ciano}-} & \cellcolor{ciano}- & \multicolumn{1}{c;{0.7pt/1pt}}{\cellcolor{ciano}-} & \cellcolor{ciano}+ & \multicolumn{1}{c|}{\cellcolor{ciano}-} & \cellcolor{cinza}- & \multicolumn{1}{c|}{\cellcolor{cinza}+} \\
\multicolumn{1}{|c|}{\cellcolor{ciano}\textbf{4}} & \cellcolor{ciano}SNAIL \cite{mishra_simple_2017} & \multicolumn{1}{r;{0.7pt/1pt}}{\textbf{\cellcolor{ciano}2017}} & \cellcolor{ciano}- & \cellcolor{ciano}+ & \cellcolor{ciano}+ & \multicolumn{1}{c;{0.7pt/1pt}}{\cellcolor{ciano}-} & \cellcolor{ciano}- & \multicolumn{1}{c;{0.7pt/1pt}}{\cellcolor{ciano}+} & \cellcolor{ciano}\textbf{H} & \cellcolor{ciano}- & \multicolumn{1}{c;{0.7pt/1pt}}{\cellcolor{ciano}-} & \cellcolor{ciano}- & \multicolumn{1}{c;{0.7pt/1pt}}{\cellcolor{ciano}-} & \cellcolor{ciano}- & \multicolumn{1}{c|}{\cellcolor{ciano}+} & \cellcolor{cinza}- & \multicolumn{1}{c|}{\cellcolor{cinza}+} \\
\multicolumn{1}{|c|}{\cellcolor{ciano}\textbf{5}} & \cellcolor{ciano}SENet \cite{hu_squeeze-and-excitation_2017} & \multicolumn{1}{r;{0.7pt/1pt}}{\textbf{\cellcolor{ciano}2018}} & \cellcolor{ciano}- & \cellcolor{ciano}+ & \cellcolor{ciano} -& \multicolumn{1}{c;{0.7pt/1pt}}{\cellcolor{ciano}-} & \cellcolor{ciano}- & \multicolumn{1}{c;{0.7pt/1pt}}{\cellcolor{ciano}+} & \cellcolor{ciano}- & \cellcolor{ciano}- & \multicolumn{1}{c;{0.7pt/1pt}}{\cellcolor{ciano}\textbf{V}} & \cellcolor{ciano}- & \multicolumn{1}{c;{0.7pt/1pt}}{\cellcolor{ciano}-} & \cellcolor{ciano}+ & \multicolumn{1}{c|}{\cellcolor{ciano}-} & \cellcolor{cinza}- & \multicolumn{1}{c|}{\cellcolor{cinza}+} \\
\multicolumn{1}{|c|}{\cellcolor{ciano}\textbf{6}} & \cellcolor{ciano}GAT \cite{velickovic_graph_2018} & \multicolumn{1}{r;{0.7pt/1pt}}{\textbf{\cellcolor{ciano}2018}} & \cellcolor{ciano}- & \cellcolor{ciano}+ & \cellcolor{ciano} -& \multicolumn{1}{c;{0.7pt/1pt}}{\cellcolor{ciano}-} & \cellcolor{ciano}- & \multicolumn{1}{c;{0.7pt/1pt}}{\cellcolor{ciano}+} & \cellcolor{ciano}\textbf{H} & \cellcolor{ciano}- & \multicolumn{1}{c;{0.7pt/1pt}}{\cellcolor{ciano}\textbf{H}} & \cellcolor{ciano}- & \multicolumn{1}{c;{0.7pt/1pt}}{\cellcolor{ciano}-} & \cellcolor{ciano}+ & \multicolumn{1}{c|}{\cellcolor{ciano}+} & \cellcolor{cinza}- & \multicolumn{1}{c|}{\cellcolor{cinza}+} \\
\multicolumn{1}{|c|}{\cellcolor{ciano}\textbf{7}} & \cellcolor{ciano}$A^{2}-Nets$ \cite{chen20182} & \multicolumn{1}{r;{0.7pt/1pt}}{\textbf{\cellcolor{ciano}2018}} & \cellcolor{ciano}- & \cellcolor{ciano}+ & \cellcolor{ciano} -& \multicolumn{1}{c;{0.7pt/1pt}}{\cellcolor{ciano}-} & \cellcolor{ciano}- & \multicolumn{1}{c;{0.7pt/1pt}}{\cellcolor{ciano}+} & \cellcolor{ciano}\textbf{FM} & \cellcolor{ciano}- & \multicolumn{1}{c;{0.7pt/1pt}}{\cellcolor{ciano}-} & \cellcolor{ciano}- & \multicolumn{1}{c;{0.7pt/1pt}}{\cellcolor{ciano}-} & \cellcolor{ciano}- & \multicolumn{1}{c|}{\cellcolor{ciano}+} & \cellcolor{cinza}- & \multicolumn{1}{c|}{\cellcolor{cinza}+} \\
\multicolumn{1}{|c|}{\cellcolor{ciano}\textbf{8}} & \cellcolor{ciano}DANet \cite{fu_dual_2018} & \multicolumn{1}{r;{0.7pt/1pt}}{\textbf{\cellcolor{ciano}2019}} & \cellcolor{ciano}- & \cellcolor{ciano}+ & \cellcolor{ciano}+ & \multicolumn{1}{c;{0.7pt/1pt}}{\cellcolor{ciano}-} & \cellcolor{ciano}- & \multicolumn{1}{c;{0.7pt/1pt}}{\cellcolor{ciano}+} & \cellcolor{ciano}\textbf{FM} & \cellcolor{ciano}- & \multicolumn{1}{c;{0.7pt/1pt}}{\cellcolor{ciano}\textbf{V}} & \cellcolor{ciano}- & \multicolumn{1}{c;{0.7pt/1pt}}{\cellcolor{ciano}-} & \cellcolor{ciano}+ & \multicolumn{1}{c|}{\cellcolor{ciano}-} & \cellcolor{cinza}- & \multicolumn{1}{c|}{\cellcolor{cinza}+} \\
\multicolumn{1}{|c|}{\cellcolor{ciano}\textbf{9}} & \cellcolor{ciano}HAN \cite{wang2019heterogeneous} & \multicolumn{1}{r;{0.7pt/1pt}}{\textbf{\cellcolor{ciano}2019}} & \cellcolor{ciano}- & \cellcolor{ciano}+ & \cellcolor{ciano}- & \multicolumn{1}{c;{0.7pt/1pt}}{\cellcolor{ciano}-} & \cellcolor{ciano}- & \multicolumn{1}{c;{0.7pt/1pt}}{\cellcolor{ciano}+} & \cellcolor{ciano}\textbf{H} & \cellcolor{ciano}- & \multicolumn{1}{c;{0.7pt/1pt}}{\cellcolor{ciano}\textbf{H}} & \cellcolor{ciano}- & \multicolumn{1}{c;{0.7pt/1pt}}{\cellcolor{ciano}-} & \cellcolor{ciano}+ & \multicolumn{1}{c|}{\cellcolor{ciano}+} & \cellcolor{cinza}- & \multicolumn{1}{c|}{\cellcolor{cinza}+} \\
\multicolumn{1}{|c|}{\cellcolor{ciano}\textbf{10}} & \cellcolor{ciano}TIM \cite{lamb2021transformers} & \multicolumn{1}{r;{0.7pt/1pt}}{\textbf{\cellcolor{ciano}2021}} & \cellcolor{ciano}- &  \cellcolor{ciano}+ & \cellcolor{ciano}+ & \multicolumn{1}{c;{0.7pt/1pt}}{\cellcolor{ciano}-} & \cellcolor{ciano}- & \multicolumn{1}{c;{0.7pt/1pt}}{\cellcolor{ciano}+} & \cellcolor{ciano}\textbf{H} & \cellcolor{ciano}- & \multicolumn{1}{c;{0.7pt/1pt}}{\cellcolor{ciano}-} & \cellcolor{ciano}- & \multicolumn{1}{c;{0.7pt/1pt}}{\cellcolor{ciano}-} & \cellcolor{ciano}- & \multicolumn{1}{c|}{\cellcolor{ciano}+} & \cellcolor{cinza}- & \multicolumn{1}{c|}{\cellcolor{cinza}+} \\ \hline
\multicolumn{15}{l}{\textbf{Top-down}} \\ \hline
\multicolumn{1}{|c|}{\cellcolor{ciano}\textbf{11}} & \cellcolor{ciano}RNNSearch \cite{bahdanau_neural_2014} & \multicolumn{1}{r;{0.7pt/1pt}}{\textbf{\cellcolor{ciano}2014}} & \cellcolor{ciano}- & \cellcolor{ciano}+ & \cellcolor{ciano}+ & \multicolumn{1}{c;{0.7pt/1pt}}{\cellcolor{ciano}-} & \cellcolor{ciano}- & \multicolumn{1}{c;{0.7pt/1pt}}{\cellcolor{ciano}+} & \cellcolor{ciano}\textbf{H} & \cellcolor{ciano}- & \multicolumn{1}{c;{0.7pt/1pt}}{\cellcolor{ciano}-} & \cellcolor{ciano}- & \multicolumn{1}{c;{0.7pt/1pt}}{\cellcolor{ciano}-} & \cellcolor{ciano}- & \multicolumn{1}{c|}{\cellcolor{ciano}+} & \cellcolor{cinza}- & \multicolumn{1}{c|}{\cellcolor{cinza}+} \\ \multicolumn{1}{|c|}{\cellcolor{ciano}\textbf{12}} & \cellcolor{ciano}Tang et al. \cite{tang_learning_2014} & \multicolumn{1}{r;{0.7pt/1pt}}{\textbf{\cellcolor{ciano}2014}} & \cellcolor{ciano}+ & \cellcolor{ciano}- & \cellcolor{ciano}- & \multicolumn{1}{c;{0.7pt/1pt}}{\cellcolor{ciano}+} & \cellcolor{ciano}+ & \multicolumn{1}{c;{0.7pt/1pt}}{\cellcolor{ciano}-} & \cellcolor{ciano}- & \cellcolor{ciano}+ & \multicolumn{1}{c;{0.7pt/1pt}}{\cellcolor{ciano}-} & \cellcolor{ciano}- & \multicolumn{1}{c;{0.7pt/1pt}}{\cellcolor{ciano}-} & \cellcolor{ciano}- & \multicolumn{1}{c|}{\cellcolor{ciano}+} & \cellcolor{cinza}+ & \multicolumn{1}{c|}{\cellcolor{cinza}-} \\
\multicolumn{1}{|c|}{\cellcolor{ciano}\textbf{13}} & \cellcolor{ciano}aNN \cite{wang2014attentional} & \multicolumn{1}{r;{0.7pt/1pt}}{\textbf{\cellcolor{ciano}2014}} & \cellcolor{ciano}- & \cellcolor{ciano}+ & \cellcolor{ciano}+ & \multicolumn{1}{c;{0.7pt/1pt}}{\cellcolor{ciano}-} & \cellcolor{ciano}- & \multicolumn{1}{c;{0.7pt/1pt}}{\cellcolor{ciano}+} & \cellcolor{ciano}- & \cellcolor{ciano}- & \multicolumn{1}{c;{0.7pt/1pt}}{\cellcolor{ciano}+} & \cellcolor{ciano}- & \multicolumn{1}{c;{0.7pt/1pt}}{\cellcolor{ciano}-} & \cellcolor{ciano}- & \multicolumn{1}{c|}{\cellcolor{ciano}+} & \cellcolor{cinza}- & \multicolumn{1}{c|}{\cellcolor{cinza}+} \\
\multicolumn{1}{|c|}{\cellcolor{ciano}\textbf{14}} & \cellcolor{ciano}NTM \cite{graves_neural_2014} & \multicolumn{1}{r;{0.7pt/1pt}}{\textbf{\cellcolor{ciano}2014}} & \cellcolor{ciano}- & \cellcolor{ciano}+ & \cellcolor{ciano}+ & \multicolumn{1}{c;{0.7pt/1pt}}{\cellcolor{ciano}-} & \cellcolor{ciano}- & \multicolumn{1}{c;{0.7pt/1pt}}{\cellcolor{ciano}+} & \cellcolor{ciano}\textbf{EM} & \cellcolor{ciano}- & \multicolumn{1}{c;{0.7pt/1pt}}{\cellcolor{ciano}\textbf{O}} & \cellcolor{ciano}- & \multicolumn{1}{c;{0.7pt/1pt}}{\cellcolor{ciano}-} & \cellcolor{ciano}- & \multicolumn{1}{c|}{\cellcolor{ciano}+} & \cellcolor{cinza}+ & \multicolumn{1}{c|}{\cellcolor{cinza}+} \\
\multicolumn{1}{|c|}{\cellcolor{ciano}\textbf{15}} & \cellcolor{ciano}RAM \cite{mnih_recurrent_2014} & \multicolumn{1}{r;{0.7pt/1pt}}{\textbf{\cellcolor{ciano}2014}} & \cellcolor{ciano}+ & \cellcolor{ciano}- & \cellcolor{ciano}+ & \multicolumn{1}{c;{0.7pt/1pt}}{\cellcolor{ciano}-} & \cellcolor{ciano}+ & \multicolumn{1}{c;{0.7pt/1pt}}{\cellcolor{ciano}-} & \cellcolor{ciano}\textbf{I} & \cellcolor{ciano}- & \multicolumn{1}{c;{0.7pt/1pt}}{\cellcolor{ciano}-} & \cellcolor{ciano}- & \multicolumn{1}{c;{0.7pt/1pt}}{\cellcolor{ciano}-} & \cellcolor{ciano}- & \multicolumn{1}{c|}{\cellcolor{ciano}+} & \cellcolor{cinza}+ & \multicolumn{1}{c|}{\cellcolor{cinza}-} \\ 
\multicolumn{1}{|c|}{\cellcolor{ciano}\textbf{16}} & \cellcolor{ciano}dasNet \cite{stollenga_deep_2014} & \multicolumn{1}{r;{0.7pt/1pt}}{\textbf{\cellcolor{ciano}2014}} & \cellcolor{ciano}- & \cellcolor{ciano}+ & \cellcolor{ciano}+ & \multicolumn{1}{c;{0.7pt/1pt}}{\cellcolor{ciano}-} & \cellcolor{ciano}- & \multicolumn{1}{c;{0.7pt/1pt}}{\cellcolor{ciano}+} & \cellcolor{ciano}- & \cellcolor{ciano}- & \multicolumn{1}{c;{0.7pt/1pt}}{\cellcolor{ciano}\textbf{V}} & \cellcolor{ciano}- & \multicolumn{1}{c;{0.7pt/1pt}}{\cellcolor{ciano}-} & \cellcolor{ciano}- & \multicolumn{1}{c|}{\cellcolor{ciano}+} & \cellcolor{cinza}- & \multicolumn{1}{c|}{\cellcolor{cinza}+} \\ 
\multicolumn{1}{|c|}{\cellcolor{ciano}\textbf{17}}           & \cellcolor{ciano}EMNet \cite{sukhbaatar2015end}                 & \multicolumn{1}{r;{0.7pt/1pt}}{\textbf{\cellcolor{ciano}2015}}     &     \cellcolor{ciano}-        &       \cellcolor{ciano}+  &  \cellcolor{ciano}-  & \multicolumn{1}{c;{0.7pt/1pt}}{\cellcolor{ciano}-}            &    \cellcolor{ciano}-         & \multicolumn{1}{c;{0.7pt/1pt}}{\cellcolor{ciano}+}            &     \cellcolor{ciano}\textbf{EM}        &      \cellcolor{ciano}-       & \multicolumn{1}{c;{0.7pt/1pt}}{\cellcolor{ciano}-} & \cellcolor{ciano}- & \multicolumn{1}{c;{0.7pt/1pt}}{\cellcolor{ciano}-} & \cellcolor{ciano}- & \multicolumn{1}{c|}{\cellcolor{ciano}+} &       \cellcolor{cinza}-      & \multicolumn{1}{c|}{\cellcolor{cinza}+}             \\
\multicolumn{1}{|c|}{\cellcolor{ciano}\textbf{18}}           & \cellcolor{ciano}DRAW \cite{draw}                 & \multicolumn{1}{r;{0.7pt/1pt}}{\textbf{\cellcolor{ciano}2015}}     &     \cellcolor{ciano}+        &       \cellcolor{ciano}+   &  \cellcolor{ciano}+   & \multicolumn{1}{c;{0.7pt/1pt}}{\cellcolor{ciano}-}            &    \cellcolor{ciano}+         & \multicolumn{1}{c;{0.7pt/1pt}}{\cellcolor{ciano}+}            &     \cellcolor{ciano}\textbf{I/H}        &      \cellcolor{ciano}-       & \multicolumn{1}{c;{0.7pt/1pt}}{\cellcolor{ciano}-} & \cellcolor{ciano}- & \multicolumn{1}{c;{0.7pt/1pt}}{\cellcolor{ciano}-} & \cellcolor{ciano}- & \multicolumn{1}{c|}{\cellcolor{ciano}+} &       \cellcolor{cinza}-      & \multicolumn{1}{c|}{\cellcolor{cinza}+}             \\ 
\multicolumn{1}{|c|}{\cellcolor{ciano}\textbf{19}}           & \cellcolor{ciano}Xu et al. \cite{xu_show_2015}                  & \multicolumn{1}{r;{0.7pt/1pt}}{\textbf{\cellcolor{ciano}2015}}     &     \cellcolor{ciano}-        &  \cellcolor{ciano}+   &   \cellcolor{ciano}+     & \multicolumn{1}{c;{0.7pt/1pt}}{\cellcolor{ciano}-}            &     \cellcolor{ciano}-      & \multicolumn{1}{c;{0.7pt/1pt}}{\cellcolor{ciano}+}            &    \cellcolor{ciano}\textbf{H}         &     \cellcolor{ciano}-        & \multicolumn{1}{c;{0.7pt/1pt}}{\cellcolor{ciano}-}            &      \cellcolor{ciano}-       & \multicolumn{1}{c;{0.7pt/1pt}}{\cellcolor{ciano}-}             &      \cellcolor{ciano}-        & \multicolumn{1}{c|}{\cellcolor{ciano}+}             &     \cellcolor{cinza}+         & \multicolumn{1}{c|}{\cellcolor{cinza}+}             \\
\multicolumn{1}{|c|}{\cellcolor{ciano}\textbf{20}}           & \cellcolor{ciano}Ptr-Net \cite{vinyals2015pointer}                  & \multicolumn{1}{r;{0.7pt/1pt}}{\textbf{\cellcolor{ciano}2015}}     &     \cellcolor{ciano}+        &  \cellcolor{ciano}+  &    \cellcolor{ciano}+     & \multicolumn{1}{c;{0.7pt/1pt}}{\cellcolor{ciano}-}            &       \cellcolor{ciano}-      & \multicolumn{1}{c;{0.7pt/1pt}}{\cellcolor{ciano}+}            &    \cellcolor{ciano}\textbf{H}         &     \cellcolor{ciano}-        & \multicolumn{1}{c;{0.7pt/1pt}}{\cellcolor{ciano}-}            &      \cellcolor{ciano}-       & \multicolumn{1}{c;{0.7pt/1pt}}{\cellcolor{ciano}-}             &      \cellcolor{ciano}-        & \multicolumn{1}{c|}{\cellcolor{ciano}+}             &     \cellcolor{cinza}+         & \multicolumn{1}{c|}{\cellcolor{cinza}+}             \\
\multicolumn{1}{|c|}{\cellcolor{ciano}\textbf{21}}           & \cellcolor{ciano}Rockt{\"a}schel et al. \cite{rocktaschel_reasoning_2015}                  & \multicolumn{1}{r;{0.7pt/1pt}}{\textbf{\cellcolor{ciano}2015}}     &     \cellcolor{ciano}-        &  \cellcolor{ciano}+     &  \cellcolor{ciano}+    & \multicolumn{1}{c;{0.7pt/1pt}}{\cellcolor{ciano}-}            &       \cellcolor{ciano}-      & \multicolumn{1}{c;{0.7pt/1pt}}{\cellcolor{ciano}+}            &    \cellcolor{ciano}\textbf{H}         &     \cellcolor{ciano}-        & \multicolumn{1}{c;{0.7pt/1pt}}{\cellcolor{ciano}-}            &      \cellcolor{ciano}-       & \multicolumn{1}{c;{0.7pt/1pt}}{\cellcolor{ciano}-}             &      \cellcolor{ciano}-        & \multicolumn{1}{c|}{\cellcolor{ciano}+}             &     \cellcolor{cinza}-         & \multicolumn{1}{c|}{\cellcolor{cinza}+}             \\
\multicolumn{1}{|c|}{\cellcolor{ciano}\textbf{22}}           & \cellcolor{ciano}Luong et al. \cite{luong_effective_2015}                  & \multicolumn{1}{r;{0.7pt/1pt}}{\textbf{\cellcolor{ciano}2015}}     &     \cellcolor{ciano}-        &  \cellcolor{ciano}+    &   \cellcolor{ciano}+    & \multicolumn{1}{c;{0.7pt/1pt}}{\cellcolor{ciano}-}            &       \cellcolor{ciano}-      & \multicolumn{1}{c;{0.7pt/1pt}}{\cellcolor{ciano}+}            &    \cellcolor{ciano}\textbf{H}         &     \cellcolor{ciano}-        & \multicolumn{1}{c;{0.7pt/1pt}}{\cellcolor{ciano}-}            &      \cellcolor{ciano}-       & \multicolumn{1}{c;{0.7pt/1pt}}{\cellcolor{ciano}-}             &      \cellcolor{ciano}-        & \multicolumn{1}{c|}{\cellcolor{ciano}+}             &     \cellcolor{cinza}-         & \multicolumn{1}{c|}{\cellcolor{cinza}+}             \\
\multicolumn{1}{|c|}{\cellcolor{ciano}\textbf{23}}           & \cellcolor{ciano}Hermann et al. \cite{hermann2015teaching}                  & \multicolumn{1}{r;{0.7pt/1pt}}{\textbf{\cellcolor{ciano}2015}}     &     \cellcolor{ciano}-        &  \cellcolor{ciano}+    &  \cellcolor{ciano}+     & \multicolumn{1}{c;{0.7pt/1pt}}{\cellcolor{ciano}-}            &       \cellcolor{ciano}-      & \multicolumn{1}{c;{0.7pt/1pt}}{\cellcolor{ciano}+}            &    \cellcolor{ciano}\textbf{H}         &     \cellcolor{ciano}-        & \multicolumn{1}{c;{0.7pt/1pt}}{\cellcolor{ciano}-}            &      \cellcolor{ciano}-       & \multicolumn{1}{c;{0.7pt/1pt}}{\cellcolor{ciano}-}             &      \cellcolor{ciano}-        & \multicolumn{1}{c|}{\cellcolor{ciano}+}             &     \cellcolor{cinza}-         & \multicolumn{1}{c|}{\cellcolor{cinza}+}             \\
\multicolumn{1}{|c|}{\cellcolor{ciano}\textbf{24}}           & \cellcolor{ciano}DMN \cite{kumar_ask_2015}                  & \multicolumn{1}{r;{0.7pt/1pt}}{\textbf{\cellcolor{ciano}2015}}     &     \cellcolor{ciano}+        &  \cellcolor{ciano}-   &       \cellcolor{ciano}+ & \multicolumn{1}{c;{0.7pt/1pt}}{\cellcolor{ciano}-}            &       \cellcolor{ciano}-      & \multicolumn{1}{c;{0.7pt/1pt}}{\cellcolor{ciano}+}            &    \cellcolor{ciano}\textbf{H}         &     \cellcolor{ciano}-        & \multicolumn{1}{c;{0.7pt/1pt}}{\cellcolor{ciano}-}            &      \cellcolor{ciano}-       & \multicolumn{1}{c;{0.7pt/1pt}}{\cellcolor{ciano}-}             &      \cellcolor{ciano}-        & \multicolumn{1}{c|}{\cellcolor{ciano}+}             &     \cellcolor{cinza}-         & \multicolumn{1}{c|}{\cellcolor{cinza}+}             \\
\multicolumn{1}{|c|}{\cellcolor{ciano}\textbf{25}}           & \cellcolor{ciano}BiDAF \cite{seo_bidirectional_2016}                   & \multicolumn{1}{r;{0.7pt/1pt}}{\textbf{\cellcolor{ciano}2016}}     &     \cellcolor{ciano}+        &  \cellcolor{ciano}+ &      \cellcolor{ciano}-    & \multicolumn{1}{c;{0.7pt/1pt}}{\cellcolor{ciano}-}            &      \cellcolor{ciano}-      & \multicolumn{1}{c;{0.7pt/1pt}}{\cellcolor{ciano}+}            &    \cellcolor{ciano}\textbf{H}         &     \cellcolor{ciano}-        & \multicolumn{1}{c;{0.7pt/1pt}}{\cellcolor{ciano}-}            &      \cellcolor{ciano}-       & \multicolumn{1}{c;{0.7pt/1pt}}{\cellcolor{ciano}-}             &      \cellcolor{ciano}-        & \multicolumn{1}{c|}{\cellcolor{ciano}+}             &     \cellcolor{cinza}+         & \multicolumn{1}{c|}{\cellcolor{cinza}+}             \\
\multicolumn{1}{|c|}{\cellcolor{ciano}\textbf{26}}           & \cellcolor{ciano}STRAW \cite{mnih2016strategic}                   & \multicolumn{1}{r;{0.7pt/1pt}}{\textbf{\cellcolor{ciano}2016}}     &     \cellcolor{ciano}+        &  \cellcolor{ciano}+  &  \cellcolor{ciano}+       & \multicolumn{1}{c;{0.7pt/1pt}}{\cellcolor{ciano}-}            &  \cellcolor{ciano}-      & \multicolumn{1}{c;{0.7pt/1pt}}{\cellcolor{ciano}+}            &    \cellcolor{ciano}\textbf{EM}         &     \cellcolor{ciano}-        & \multicolumn{1}{c;{0.7pt/1pt}}{\cellcolor{ciano}-}            &      \cellcolor{ciano}+       & \multicolumn{1}{c;{0.7pt/1pt}}{\cellcolor{ciano}-}             &      \cellcolor{ciano}-        & \multicolumn{1}{c|}{\cellcolor{ciano}+}             &     \cellcolor{cinza}+         & \multicolumn{1}{c|}{\cellcolor{cinza}+}             \\
\multicolumn{1}{|c|}{\cellcolor{ciano}\textbf{27}}           & \cellcolor{ciano}Allamanis et al. \cite{allamanis2016convolutional}                   & \multicolumn{1}{r;{0.7pt/1pt}}{\textbf{\cellcolor{ciano}2016}}     &     \cellcolor{ciano}-        &  \cellcolor{ciano}+    &    \cellcolor{ciano}+   & \multicolumn{1}{c;{0.7pt/1pt}}{\cellcolor{ciano}-}            &       \cellcolor{ciano}-      & \multicolumn{1}{c;{0.7pt/1pt}}{\cellcolor{ciano}+}            &    \cellcolor{ciano}\textbf{H}         &     \cellcolor{ciano}-        & \multicolumn{1}{c;{0.7pt/1pt}}{\cellcolor{ciano}-}            &      \cellcolor{ciano}-       & \multicolumn{1}{c;{0.7pt/1pt}}{\cellcolor{ciano}-}             &      \cellcolor{ciano}-        & \multicolumn{1}{c|}{\cellcolor{ciano}+}             &     \cellcolor{cinza}-         & \multicolumn{1}{c|}{\cellcolor{cinza}+}             \\
\multicolumn{1}{|c|}{\cellcolor{ciano}\textbf{28}}           & \cellcolor{ciano}Lu et al. \cite{lu_hierarchical_2016}                   & \multicolumn{1}{r;{0.7pt/1pt}}{\textbf{\cellcolor{ciano}2016}}     &     \cellcolor{ciano}-        &  \cellcolor{ciano}+    &  \cellcolor{ciano}-      & \multicolumn{1}{c;{0.7pt/1pt}}{\cellcolor{ciano}-}            & \cellcolor{ciano}-      & \multicolumn{1}{c;{0.7pt/1pt}}{\cellcolor{ciano}+}            &    \cellcolor{ciano}\textbf{H}         &     \cellcolor{ciano}-        & \multicolumn{1}{c;{0.7pt/1pt}}{\cellcolor{ciano}-}            &      \cellcolor{ciano}-       & \multicolumn{1}{c;{0.7pt/1pt}}{\cellcolor{ciano}-}             &      \cellcolor{ciano}-        & \multicolumn{1}{c|}{\cellcolor{ciano}+}             &     \cellcolor{cinza}-         & \multicolumn{1}{c|}{\cellcolor{cinza}+}             \\
\multicolumn{1}{|c|}{\cellcolor{ciano}\textbf{29}}           & \cellcolor{ciano}ACT \cite{graves_adaptive_2016}                   & \multicolumn{1}{r;{0.7pt/1pt}}{\textbf{\cellcolor{ciano}2016}}     &     \cellcolor{ciano}+        &  \cellcolor{ciano}+  &     \cellcolor{ciano}+    & \multicolumn{1}{c;{0.7pt/1pt}}{\cellcolor{ciano}-}            &       \cellcolor{ciano}-      & \multicolumn{1}{c;{0.7pt/1pt}}{\cellcolor{ciano}+}            &    \cellcolor{ciano}-         &     \cellcolor{ciano}-        & \multicolumn{1}{c;{0.7pt/1pt}}{\cellcolor{ciano}-}            &      \cellcolor{ciano}-       & \multicolumn{1}{c;{0.7pt/1pt}}{\cellcolor{ciano}+}             &      \cellcolor{ciano}-        & \multicolumn{1}{c|}{\cellcolor{ciano}+}             &     \cellcolor{cinza}-         & \multicolumn{1}{c|}{\cellcolor{cinza}+}             \\
\multicolumn{1}{|c|}{\cellcolor{ciano}\textbf{30}}           & \cellcolor{ciano}Lu et al. \cite{lu_knowing_2017}                   & \multicolumn{1}{r;{0.7pt/1pt}}{\textbf{\cellcolor{ciano}2016}}     &     \cellcolor{ciano}-        &  \cellcolor{ciano}+  &    \cellcolor{ciano}+     & \multicolumn{1}{c;{0.7pt/1pt}}{\cellcolor{ciano}-}            & \cellcolor{ciano}-      & \multicolumn{1}{c;{0.7pt/1pt}}{\cellcolor{ciano}+}            &    \cellcolor{ciano}\textbf{FM/H}         &     \cellcolor{ciano}-        & \multicolumn{1}{c;{0.7pt/1pt}}{\cellcolor{ciano}\textbf{MC}}            &      \cellcolor{ciano}-       & \multicolumn{1}{c;{0.7pt/1pt}}{\cellcolor{ciano}-}             &      \cellcolor{ciano}-        & \multicolumn{1}{c|}{\cellcolor{ciano}+}             &     \cellcolor{cinza}-         & \multicolumn{1}{c|}{\cellcolor{cinza}+}             \\
\multicolumn{1}{|c|}{\cellcolor{ciano}\textbf{31}}           & \cellcolor{ciano}HAN \cite{yang2016hierarchical}                   & \multicolumn{1}{r;{0.7pt/1pt}}{\textbf{\cellcolor{ciano}2016}}     &     \cellcolor{ciano}-        &  \cellcolor{ciano}+  &     \cellcolor{ciano}+    & \multicolumn{1}{c;{0.7pt/1pt}}{\cellcolor{ciano}-}            &       \cellcolor{ciano}-      & \multicolumn{1}{c;{0.7pt/1pt}}{\cellcolor{ciano}+}            &    \cellcolor{ciano}\textbf{H}         &     \cellcolor{ciano}-        & \multicolumn{1}{c;{0.7pt/1pt}}{\cellcolor{ciano}-}            &      \cellcolor{ciano}-       & \multicolumn{1}{c;{0.7pt/1pt}}{\cellcolor{ciano}-}             &      \cellcolor{ciano}-        & \multicolumn{1}{c|}{\cellcolor{ciano}+}             &     \cellcolor{cinza}-         & \multicolumn{1}{c|}{\cellcolor{cinza}+}             \\
\multicolumn{1}{|c|}{\cellcolor{ciano}\textbf{32}}           & \cellcolor{ciano}Excitation Backprop \cite{zhang2018top}                   & \multicolumn{1}{r;{0.7pt/1pt}}{\textbf{\cellcolor{ciano}2016}}     &     \cellcolor{ciano}-        &  \cellcolor{ciano}+  &    \cellcolor{ciano}-      & \multicolumn{1}{c;{0.7pt/1pt}}{\cellcolor{ciano}-}            &       \cellcolor{ciano}-      & \multicolumn{1}{c;{0.7pt/1pt}}{\cellcolor{ciano}+}            &    \cellcolor{ciano}\textbf{N}        &     \cellcolor{ciano}-        & \multicolumn{1}{c;{0.7pt/1pt}}{\cellcolor{ciano}-}            &      \cellcolor{ciano}-       & \multicolumn{1}{c;{0.7pt/1pt}}{\cellcolor{ciano}-}             &      \cellcolor{ciano}-        & \multicolumn{1}{c|}{\cellcolor{ciano}+}             &     \cellcolor{cinza}-         & \multicolumn{1}{c|}{\cellcolor{cinza}+}             \\
\multicolumn{1}{|c|}{\cellcolor{ciano}\textbf{33}}           & \cellcolor{ciano}DCN \cite{xiong2016dynamic}                   & \multicolumn{1}{r;{0.7pt/1pt}}{\textbf{\cellcolor{ciano}2016}}     &     \cellcolor{ciano}-        &  \cellcolor{ciano}+  &     \cellcolor{ciano}-    & \multicolumn{1}{c;{0.7pt/1pt}}{\cellcolor{ciano}-}            &       \cellcolor{ciano}-      & \multicolumn{1}{c;{0.7pt/1pt}}{\cellcolor{ciano}+}            &    \cellcolor{ciano}\textbf{H}         &     \cellcolor{ciano}-        & \multicolumn{1}{c;{0.7pt/1pt}}{\cellcolor{ciano}-}            &      \cellcolor{ciano}-       & \multicolumn{1}{c;{0.7pt/1pt}}{\cellcolor{ciano}-}             &      \cellcolor{ciano}-        & \multicolumn{1}{c|}{\cellcolor{ciano}+}             &     \cellcolor{cinza}+         & \multicolumn{1}{c|}{\cellcolor{cinza}+}             \\
\multicolumn{1}{|c|}{\cellcolor{ciano}\textbf{34}}           & \cellcolor{ciano}GCA-LSTM \cite{liu_global_2017}                 & \multicolumn{1}{r;{0.7pt/1pt}}{\textbf{\cellcolor{ciano}2017}}     &    \cellcolor{ciano}-         &  \cellcolor{ciano}+  &  \cellcolor{ciano}+    & \multicolumn{1}{c;{0.7pt/1pt}}{\cellcolor{ciano}-}            &      \cellcolor{ciano}-       & \multicolumn{1}{c;{0.7pt/1pt}}{\cellcolor{ciano}+}            &       \cellcolor{ciano}\textbf{H}      &    \cellcolor{ciano}-         & \multicolumn{1}{c;{0.7pt/1pt}}{\cellcolor{ciano}-}            &     \cellcolor{ciano}-        & \multicolumn{1}{c;{0.7pt/1pt}}{\cellcolor{ciano}-}             &      \cellcolor{ciano}-     & \multicolumn{1}{c|}{\cellcolor{ciano}+} &   \cellcolor{cinza}-   & \multicolumn{1}{c|}{\cellcolor{cinza}+} \\ 
\multicolumn{1}{|c|}{\cellcolor{ciano}\textbf{35}}           & \cellcolor{ciano}Reed et al. \cite{reed_few-shot_2017}                 & \multicolumn{1}{r;{0.7pt/1pt}}{\textbf{\cellcolor{ciano}2017}}     &    \cellcolor{ciano}-         &     \cellcolor{ciano}+   &   \cellcolor{ciano}+    & \multicolumn{1}{c;{0.7pt/1pt}}{\cellcolor{ciano}-}            &      \cellcolor{ciano}-       & \multicolumn{1}{c;{0.7pt/1pt}}{\cellcolor{ciano}+}            &       \cellcolor{ciano}\textbf{H}      &    \cellcolor{ciano}-         & \multicolumn{1}{c;{0.7pt/1pt}}{\cellcolor{ciano}-}            &     \cellcolor{ciano}-        & \multicolumn{1}{c;{0.7pt/1pt}}{\cellcolor{ciano}-}             &      \cellcolor{ciano}-     & \multicolumn{1}{c|}{\cellcolor{ciano}+} &   \cellcolor{cinza}-   & \multicolumn{1}{c|}{\cellcolor{cinza}+} \\
\multicolumn{1}{|c|}{\cellcolor{ciano}\textbf{36}}           & \cellcolor{ciano}Seo et al. \cite{seo2017visual}                 & \multicolumn{1}{r;{0.7pt/1pt}}{\textbf{\cellcolor{ciano}2017}}     &    \cellcolor{ciano}-         &     \cellcolor{ciano}+  &    \cellcolor{ciano}+  & \multicolumn{1}{c;{0.7pt/1pt}}{\cellcolor{ciano}-}            &      \cellcolor{ciano}-       & \multicolumn{1}{c;{0.7pt/1pt}}{\cellcolor{ciano}+}            &       \cellcolor{ciano}\textbf{FM}      &    \cellcolor{ciano}-         & \multicolumn{1}{c;{0.7pt/1pt}}{\cellcolor{ciano}-}            &     \cellcolor{ciano}-        & \multicolumn{1}{c;{0.7pt/1pt}}{\cellcolor{ciano}-}             &      \cellcolor{ciano}-     & \multicolumn{1}{c|}{\cellcolor{ciano}+} &   \cellcolor{cinza}-   & \multicolumn{1}{c|}{\cellcolor{cinza}+} \\
\multicolumn{1}{|c|}{\cellcolor{ciano}\textbf{37}}           & \cellcolor{ciano}SAB \cite{ke_sparse_2018}\cite{ke2017sparse}                 & \multicolumn{1}{r;{0.7pt/1pt}}{\textbf{\cellcolor{ciano}2017}}     &    \cellcolor{ciano}-         &     \cellcolor{ciano}+   &  \cellcolor{ciano}+   & \multicolumn{1}{c;{0.7pt/1pt}}{\cellcolor{ciano}-}            &      \cellcolor{ciano}-       & \multicolumn{1}{c;{0.7pt/1pt}}{\cellcolor{ciano}+}            &       \cellcolor{ciano}\textbf{H}      &    \cellcolor{ciano}-         & \multicolumn{1}{c;{0.7pt/1pt}}{\cellcolor{ciano}-}            &     \cellcolor{ciano}-        & \multicolumn{1}{c;{0.7pt/1pt}}{\cellcolor{ciano}-}             &      \cellcolor{ciano}-     & \multicolumn{1}{c|}{\cellcolor{ciano}+} &   \cellcolor{cinza}-   & \multicolumn{1}{c|}{\cellcolor{cinza}+} \\ 
\multicolumn{1}{|c|}{\cellcolor{ciano}\textbf{38}}           & \cellcolor{ciano}ACF Network \cite{choi_attentional_2017}                 & \multicolumn{1}{r;{0.7pt/1pt}}{\textbf{\cellcolor{ciano}2017}}     &    \cellcolor{ciano}+         &     \cellcolor{ciano}-    &  \cellcolor{ciano}+   & \multicolumn{1}{c;{0.7pt/1pt}}{\cellcolor{ciano}-}            &      \cellcolor{ciano}-       & \multicolumn{1}{c;{0.7pt/1pt}}{\cellcolor{ciano}+}            &       \cellcolor{ciano}-      &    \cellcolor{ciano}-         & \multicolumn{1}{c;{0.7pt/1pt}}{\cellcolor{ciano}-}            &     \cellcolor{ciano}+        & \multicolumn{1}{c;{0.7pt/1pt}}{\cellcolor{ciano}-}             &      \cellcolor{ciano}-     & \multicolumn{1}{c|}{\cellcolor{ciano}+} &   \cellcolor{cinza}+   & \multicolumn{1}{c|}{\cellcolor{cinza}-} \\
\multicolumn{1}{|c|}{\cellcolor{ciano}\textbf{39}}           & \cellcolor{ciano}Kim et al. \cite{kim_structured_2017}              & \multicolumn{1}{r;{0.7pt/1pt}}{\textbf{\cellcolor{ciano}2017}}     &  \cellcolor{ciano}-           & \cellcolor{ciano}+  & \cellcolor{ciano}-      & \multicolumn{1}{c;{0.7pt/1pt}}{\cellcolor{ciano}-}            &     \cellcolor{ciano}-        & \multicolumn{1}{c;{0.7pt/1pt}}{\cellcolor{ciano}+}            &      \cellcolor{ciano}\textbf{H}       &   \cellcolor{ciano}-          &  \multicolumn{1}{c;{0.7pt/1pt}}{\cellcolor{ciano}-}            &      \cellcolor{ciano}-       & \multicolumn{1}{c;{0.7pt/1pt}}{\cellcolor{ciano}-}             &      \cellcolor{ciano}-        & \multicolumn{1}{c|}{\cellcolor{ciano}+}             &      \cellcolor{cinza}-        & \multicolumn{1}{c|}{\cellcolor{cinza}+}             \\
\multicolumn{1}{|c|}{\cellcolor{ciano}\textbf{40}}           & \cellcolor{ciano}BAN \cite{kim_bilinear_2018}                 & \multicolumn{1}{r;{0.7pt/1pt}}{\textbf{\cellcolor{ciano}2018}}     &    \cellcolor{ciano}-         &     \cellcolor{ciano}+   &  \cellcolor{ciano}-   & \multicolumn{1}{c;{0.7pt/1pt}}{\cellcolor{ciano}-}            &      \cellcolor{ciano}-       & \multicolumn{1}{c;{0.7pt/1pt}}{\cellcolor{ciano}+}            &     \cellcolor{ciano}-      &    \cellcolor{ciano}-         & \multicolumn{1}{c;{0.7pt/1pt}}{\cellcolor{ciano}\textbf{V}}            &     \cellcolor{ciano}-        & \multicolumn{1}{c;{0.7pt/1pt}}{\cellcolor{ciano}-}             &      \cellcolor{ciano}-     & \multicolumn{1}{c|}{\cellcolor{ciano}+} &   \cellcolor{cinza}-   & \multicolumn{1}{c|}{\cellcolor{cinza}+} \\
\multicolumn{1}{|c|}{\cellcolor{ciano}\textbf{41}}           & \cellcolor{ciano}AG \cite{schlemper_attention_2018}                 & \multicolumn{1}{r;{0.7pt/1pt}}{\textbf{\cellcolor{ciano}2018}}     &    \cellcolor{ciano}-         &     \cellcolor{ciano}+ &  \cellcolor{ciano}-     & \multicolumn{1}{c;{0.7pt/1pt}}{\cellcolor{ciano}-}            &      \cellcolor{ciano}-       & \multicolumn{1}{c;{0.7pt/1pt}}{\cellcolor{ciano}+}            &       \cellcolor{ciano}\textbf{FM}      &    \cellcolor{ciano}-         & \multicolumn{1}{c;{0.7pt/1pt}}{\cellcolor{ciano}-}            &     \cellcolor{ciano}-        & \multicolumn{1}{c;{0.7pt/1pt}}{\cellcolor{ciano}-}             &      \cellcolor{ciano}-     & \multicolumn{1}{c|}{\cellcolor{ciano}+} &   \cellcolor{cinza}-   & \multicolumn{1}{c|}{\cellcolor{cinza}+} \\
\multicolumn{1}{|c|}{\cellcolor{ciano}\textbf{42}}           & \cellcolor{ciano}Perera et al. \cite{perera2020lstm}                 & \multicolumn{1}{r;{0.7pt/1pt}}{\textbf{\cellcolor{ciano}2018}}     &    \cellcolor{ciano}-         &     \cellcolor{ciano}+  &   \cellcolor{ciano}+    & \multicolumn{1}{c;{0.7pt/1pt}}{\cellcolor{ciano}-}            &      \cellcolor{ciano}-       & \multicolumn{1}{c;{0.7pt/1pt}}{\cellcolor{ciano}+}            &       \cellcolor{ciano}\textbf{H}      &    \cellcolor{ciano}-         & \multicolumn{1}{c;{0.7pt/1pt}}{\cellcolor{ciano}\textbf{MC}}            &     \cellcolor{ciano}-        & \multicolumn{1}{c;{0.7pt/1pt}}{\cellcolor{ciano}-}             &      \cellcolor{ciano}-     & \multicolumn{1}{c|}{\cellcolor{ciano}+} &   \cellcolor{cinza}-   & \multicolumn{1}{c|}{\cellcolor{cinza}+} \\
\multicolumn{1}{|c|}{\cellcolor{ciano}\textbf{43}}           & \cellcolor{ciano}Deng et al. \cite{deng_latent_2018}                 & \multicolumn{1}{r;{0.7pt/1pt}}{\textbf{\cellcolor{ciano}2018}}     &    \cellcolor{ciano}-         &     \cellcolor{ciano}+   &   \cellcolor{ciano}-   & \multicolumn{1}{c;{0.7pt/1pt}}{\cellcolor{ciano}-}            &      \cellcolor{ciano}-       & \multicolumn{1}{c;{0.7pt/1pt}}{\cellcolor{ciano}+}            &       \cellcolor{ciano}\textbf{H}      &    \cellcolor{ciano}-         & \multicolumn{1}{c;{0.7pt/1pt}}{\cellcolor{ciano}-}            &     \cellcolor{ciano}-        & \multicolumn{1}{c;{0.7pt/1pt}}{\cellcolor{ciano}-}             &      \cellcolor{ciano}-     & \multicolumn{1}{c|}{\cellcolor{ciano}+} &   \cellcolor{cinza}-   & \multicolumn{1}{c|}{\cellcolor{cinza}+} \\
\multicolumn{1}{|c|}{\cellcolor{ciano}\textbf{44}}           & \cellcolor{ciano}HAN \cite{kim2020hypergraph}                 & \multicolumn{1}{r;{0.7pt/1pt}}{\textbf{\cellcolor{ciano}2020}}     &    \cellcolor{ciano}-         &     \cellcolor{ciano}+   &  \cellcolor{ciano}-   & \multicolumn{1}{c;{0.7pt/1pt}}{\cellcolor{ciano}-}            &      \cellcolor{ciano}-       & \multicolumn{1}{c;{0.7pt/1pt}}{\cellcolor{ciano}+}            &       \cellcolor{ciano}\textbf{M}      &    \cellcolor{ciano}-         & \multicolumn{1}{c;{0.7pt/1pt}}{\cellcolor{ciano}-}            &     \cellcolor{ciano}-        & \multicolumn{1}{c;{0.7pt/1pt}}{\cellcolor{ciano}-}             &      \cellcolor{ciano}-     & \multicolumn{1}{c|}{\cellcolor{ciano}+} &   \cellcolor{cinza}-   & \multicolumn{1}{c|}{\cellcolor{cinza}+} \\
\multicolumn{1}{|c|}{\cellcolor{ciano}\textbf{45}}           & \cellcolor{ciano}IMRAM \cite{chen2020imram}                 & \multicolumn{1}{r;{0.7pt/1pt}}{\textbf{\cellcolor{ciano}2020}}     &    \cellcolor{ciano}-         &     \cellcolor{ciano}+   &    \cellcolor{ciano}-    & \multicolumn{1}{c;{0.7pt/1pt}}{\cellcolor{ciano}-}            &      \cellcolor{ciano}-       & \multicolumn{1}{c;{0.7pt/1pt}}{\cellcolor{ciano}+}            &       \cellcolor{ciano}\textbf{H}      &    \cellcolor{ciano}-         & \multicolumn{1}{c;{0.7pt/1pt}}{\cellcolor{ciano}-}            &     \cellcolor{ciano}-        & \multicolumn{1}{c;{0.7pt/1pt}}{\cellcolor{ciano}-}             &      \cellcolor{ciano}-     & \multicolumn{1}{c|}{\cellcolor{ciano}+} &   \cellcolor{cinza}-   & \multicolumn{1}{c|}{\cellcolor{cinza}+} \\
\multicolumn{1}{|c|}{\cellcolor{ciano}\textbf{46}}           & \cellcolor{ciano}Lekkala et al. \cite{lekkala2020attentive}                 & \multicolumn{1}{r;{0.7pt/1pt}}{\textbf{\cellcolor{ciano}2020}}     &    \cellcolor{ciano}-         &     \cellcolor{ciano}+  &  \cellcolor{ciano}-     & \multicolumn{1}{c;{0.7pt/1pt}}{\cellcolor{ciano}-}            &      \cellcolor{ciano}-       & \multicolumn{1}{c;{0.7pt/1pt}}{\cellcolor{ciano}+}            &       \cellcolor{ciano}-      &    \cellcolor{ciano}-         & \multicolumn{1}{c;{0.7pt/1pt}}{\cellcolor{ciano}\textbf{V}}            &     \cellcolor{ciano}-        & \multicolumn{1}{c;{0.7pt/1pt}}{\cellcolor{ciano}-}             &      \cellcolor{ciano}-     & \multicolumn{1}{c|}{\cellcolor{ciano}+} &   \cellcolor{cinza}-   & \multicolumn{1}{c|}{\cellcolor{cinza}+} \\ \hline

\multicolumn{15}{l}{\textbf{Hybrid}} \\ \hline

\multicolumn{1}{|c|}{\cellcolor{ciano}\textbf{47}}           & \cellcolor{ciano}Transformer \cite{vaswani_attention_2017}                 & \multicolumn{1}{r;{0.7pt/1pt}}{\textbf{\cellcolor{ciano}2017}}     &    \cellcolor{ciano}-         &     \cellcolor{ciano}+   &  \cellcolor{ciano}+    & \multicolumn{1}{c;{0.7pt/1pt}}{\cellcolor{ciano}-}            &      \cellcolor{ciano}-       & \multicolumn{1}{c;{0.7pt/1pt}}{\cellcolor{ciano}+}            &       \cellcolor{ciano}\textbf{H}      &    \cellcolor{ciano}-         & \multicolumn{1}{c;{0.7pt/1pt}}{\cellcolor{ciano}-}            &     \cellcolor{ciano}-        & \multicolumn{1}{c;{0.7pt/1pt}}{\cellcolor{ciano}-}             &      \cellcolor{ciano}-     & \multicolumn{1}{c|}{\cellcolor{ciano}+} &   \cellcolor{cinza}-   & \multicolumn{1}{c|}{\cellcolor{cinza}+} \\
\multicolumn{1}{|c|}{\cellcolor{ciano}\textbf{48}}           & \cellcolor{ciano}DiSAN \cite{shen2018disan}                 & \multicolumn{1}{r;{0.7pt/1pt}}{\textbf{\cellcolor{ciano}2018}}     &    \cellcolor{ciano}-         &     \cellcolor{ciano}+   &    \cellcolor{ciano}- & \multicolumn{1}{c;{0.7pt/1pt}}{\cellcolor{ciano}-}            &      \cellcolor{ciano}-       & \multicolumn{1}{c;{0.7pt/1pt}}{\cellcolor{ciano}+}            &       \cellcolor{ciano}\textbf{H}      &    \cellcolor{ciano}-         & \multicolumn{1}{c;{0.7pt/1pt}}{\cellcolor{ciano}\textbf{L}}            &     \cellcolor{ciano}-        & \multicolumn{1}{c;{0.7pt/1pt}}{\cellcolor{ciano}-}             &      \cellcolor{ciano}-     & \multicolumn{1}{c|}{\cellcolor{ciano}+} &   \cellcolor{cinza}-   & \multicolumn{1}{c|}{\cellcolor{cinza}+} \\
\multicolumn{1}{|c|}{\cellcolor{ciano}\textbf{49}}           & \cellcolor{ciano}ANP \cite{kim2019attentive}                 & \multicolumn{1}{r;{0.7pt/1pt}}{\textbf{\cellcolor{ciano}2019}}     &    \cellcolor{ciano}-         &     \cellcolor{ciano}+   &   \cellcolor{ciano}-   & \multicolumn{1}{c;{0.7pt/1pt}}{\cellcolor{ciano}-}            &     \cellcolor{ciano}+       & \multicolumn{1}{c;{0.7pt/1pt}}{\cellcolor{ciano}+}            &       \cellcolor{ciano}\textbf{O}      &    \cellcolor{ciano}-         & \multicolumn{1}{c;{0.7pt/1pt}}{\cellcolor{ciano}-}            &     \cellcolor{ciano}-        & \multicolumn{1}{c;{0.7pt/1pt}}{\cellcolor{ciano}-}             &      \cellcolor{ciano}-     & \multicolumn{1}{c|}{\cellcolor{ciano}+} &   \cellcolor{cinza}-   & \multicolumn{1}{c|}{\cellcolor{cinza}+} \\
\multicolumn{1}{|c|}{\cellcolor{ciano}\textbf{50}}           & \cellcolor{ciano}BRIMs \cite{mittal2020learning}                 & \multicolumn{1}{r;{0.7pt/1pt}}{\textbf{\cellcolor{ciano}2020}}     &    \cellcolor{ciano}-         &     \cellcolor{ciano}+  &     \cellcolor{ciano}+  & \multicolumn{1}{c;{0.7pt/1pt}}{\cellcolor{ciano}-}            &      \cellcolor{ciano}-       & \multicolumn{1}{c;{0.7pt/1pt}}{\cellcolor{ciano}+}            &       \cellcolor{ciano}\textbf{H}      &    \cellcolor{ciano}-         & \multicolumn{1}{c;{0.7pt/1pt}}{\cellcolor{ciano}-}            &     \cellcolor{ciano}-        & \multicolumn{1}{c;{0.7pt/1pt}}{\cellcolor{ciano}-}             &      \cellcolor{ciano}-     & \multicolumn{1}{c|}{\cellcolor{ciano}+} &   \cellcolor{cinza}-   & \multicolumn{1}{c|}{\cellcolor{cinza}+} \\
\multicolumn{1}{|c|}{\cellcolor{ciano}\textbf{51}}           & \cellcolor{ciano}MSAN \cite{kim2020modality}                 & \multicolumn{1}{r;{0.7pt/1pt}}{\textbf{\cellcolor{ciano}2020}}     &    \cellcolor{ciano}+         &     \cellcolor{ciano}+    &  \cellcolor{ciano}+   & \multicolumn{1}{c;{0.7pt/1pt}}{\cellcolor{ciano}-}            &      \cellcolor{ciano}-       & \multicolumn{1}{c;{0.7pt/1pt}}{\cellcolor{ciano}+}            &       \cellcolor{ciano}\textbf{H}      &    \cellcolor{ciano}-         & \multicolumn{1}{c;{0.7pt/1pt}}{\cellcolor{ciano}-}            &     \cellcolor{ciano}+        & \multicolumn{1}{c;{0.7pt/1pt}}{\cellcolor{ciano}-}             &      \cellcolor{ciano}-     & \multicolumn{1}{c|}{\cellcolor{ciano}+} &   \cellcolor{cinza}+   & \multicolumn{1}{c|}{\cellcolor{cinza}+}            \\ \hline
\end{tabular}

\end{center}
\caption{Summary of main neural attention models. Factors in order are: Selective (f1), divided (f2), oriented (f3), sustained (f4), selective perception (f5), selective cognition (f6), location-based (f7), object-based (f8), feature-based (f9), task-oriented (f10), time-oriented (f11), stateless (f12), stateful (f13), hard (f14), soft (f15). In the location-based column (f7) column: hidden states/data embeddings (H), external memory cells (EM), feature maps (FM), input data (I), and others (O). In the feature-based column (f9) column: visual (V), linguistic (L), memory cell (MC), hidden states (H), and others (O). In the other columns, the presence of the feature is indicated by the + symbol and absence by the - symbol.} 

\label{table:main_models}

\end{figure*}

\subsection{Components: Selective, Divided, Oriented and Sustained}
\label{sec:components}

Each attention subsystem can have a selective, divided, oriented, or sustained property. Regarding the number of elements that we pay attention to simultaneously, we can classify subsystems as selective or divided attention. Regarding attention between instants of time, attention can be oriented or sustained. \textbf{Selective attention} chooses only one stimulus among all the others. In contrast, \textbf{divided attention} is the highest level of attention and refers to responding simultaneously to multiple stimuli or tasks. Biologically, divided attention can only operate two tasks simultaneously if one of them is mediated by automatic processes and the other by a cognitive \cite{eysenck2004psychology} so that only one task requires much intellectual effort.

\textbf{Oriented attention} can shift the focus by leaving a stimulus for another through three processes: 1) leave the current focus, 2) change the focus to the expected stimulus, and 3) locate the target and maintain attention. Such a component represents the ability to coordinate simultaneous tasks to be interrupted and resumed temporarily. This function is usually linked to the central executive who coordinates and manages the information processing activities of the brain \cite{oberauer2019working}. At working memory (WM), oriented attention mechanisms are extremely useful in determining which perceptual and long-term memory information is loaded and deleted at each time step $t$ among all existing information. Oriented components often select a single item or set of items from all the information available to feed some mental process input. \textbf{Sustained attention} or vigilance allows the maintenance of the goal over time through directly focusing on specific stimuli to complete a planned activity. This component plays a fundamental role in learning, performing daily tasks, maintaining dialogues and social relationships, among many other skills that affect mental health \cite{esterman2019models}. Despite this, sustained attention is generally less studied than transient aspects of attention, such as shifting, dividing, and attentional selection. What distinguishes sustained attention is the focus on performance on a single task over time. There are fluctuations within the individual's overall ability to maintain stable performance on the task, with a trade-off between exposure to stimuli and a recovery period.

In our framework, \textbf{selective attention} can be understood as the ability of the subsystem to select only one stimulus from the focus target among all stimuli or select only a subset of stimuli from the target with the same attentional weight and completely inhibit the response of the unselected stimuli. On the other hand, in \textbf{divided attention} the attentional mask is distributed over the entire input focus target so that no stimulus is completely inhibited, only modulated with weights greater than or less than its original values. In \textbf{oriented attention}, at time $t_{1}$, the attentional focus is on a non-empty subset of elements of the focus target $\tau_{t}$, at time $t_{2}$ the focus is shifted to a new subset of elements from the same target or another target, at time $t_{3}$ the focus may be a different target or initial target. However, if the target does not change over time and the attentional mask is always the same, or if the target changes over time but the attentional mask remains on the same semantic elements as the target, the system is sustained attention. For example, if the target is a sequence of images and the attentional mask remains centered around the same object throughout the sequence, or if the mask remains on the same features as the image, the system is \textbf{sustained attention}. If the attentional system can choose to change its focus in time or remain in the same focus, the system is also considered oriented attention because, at some point, there is the possibility of switching.

Oriented or sustained systems are present only in sequential/recurring architectures. The first oriented systems emerged through visual search engines in images and encoder-decoder frameworks. Visual search architectures based on \textbf{RAM} \cite{mnih_recurrent_2014} use the oriented attention to control the perceptual sensors' focus. At each time step $t$, the sensors act as retina selecting only a portion of the input image for subsequent processing. The selected image portion receives the same attentional weight while the other regions are completely inhibited from further processing, presenting selective characteristics of attention. A \textbf{DRAW} \cite{draw} uses a very similar oriented component for image generation. At each time step $t$, two attentional components shift the focus from a reading head and a writing head to small portions of the image, allowing sequential generation and refinement of the portions previously generated. However, at the same time $t$, attentional masks explicitly divide attention between all pixels in the image. However, the masking computation process on targets occurs to select only desired patches, excluding other regions in subsequent processing, such as a selective mechanism. \textbf{Spatial Transformer} \cite{jaderberg2015spatial} features a selected visual search engine that samples only one region of interest in the image, undoes spatial deformations, and passes the result on to subsequent processing.
This mechanism is very flexible regarding the temporal aspect, and its characteristics depend on the architecture to which it is attached. If it is in non-sequential architectures, it is only selected, but if it is in sequential architectures, it can act as an oriented or sustained attention mechanism. If the mechanism can track the same region of interest, it is sustained. Otherwise, it is oriented.

Most of the encoder-decoder-attention frameworks for reasoning \cite{hudson_compositional_2018} \cite{rocktaschel_reasoning_2015}\cite{nam_dual_2017}, machine comprehension \cite{hermann_teaching_2015}, neuralne and machine translation \cite{bahdanau_neural_2014} are divided and oriented attention. At each time step $t $, the attentional mask can simultaneously hide hidden states from the encoder to compose a dynamic context vector with the weighting of all information, such as a reasoning structure in a working memory that selects and modulates memories to meet some mental process. The same hidden states remain the target in the next time step, but the system can assign attentional weights in a completely different way, featuring an oriented attention system. Some attentional systems that operate on external memories are also divided and oriented. The classic \textbf{Neural Turing Machine} \cite{graves_neural_2014} \cite{mao2019aspect} divides attention on all external memory cells to retrieve a modulated representation of the stored memories. At each time step, oriented attention mechanisms are guided by a different contextual input that defines how the new distribution of attention on memories will be. \textbf{STRAW} \cite{mnih2016strategic} uses a divided and oriented read/write heads system to find and update regions with a greater focus on the action plan of a trained virtual agent via reinforcement learning. Another selected and oriented attentional system decides when the heads should act or if only advances in the time of the action-plan and commitment-plan should be made. Perera et al. \cite{perera2020lstm} uses two divided and oriented mechanisms \textbf{attention in LSTMs cells}. The first is external to the LSTM cell to aggregate historical information from the previous hidden states in just one memory vector $h_{A}^{t-1} $. The second acts inside the cell as a gate layer to update the current cell vector proportional to the importance of each portion of $h_{A}^{t-1}$. The mechanism receives the memory vector $h_{A}^{t-1}$ as the target, dividing attention over the entire vector, giving more weight to the most relevant portions and less weight to less relevant ones. The current cell vector is then updated.

The \textbf{GCA-LSTM Network} \cite{liu_global_2017} uses a divided and oriented attentional mechanism to update, at each time step $t $, the cell state in LSTM units. The mechanism generates an attentional mask for the $i_{j,t} $input gate and LSTM cell's spatial / temporal contextual information, based on previous hidden states. At \textbf{dasNet} \cite{stollenga_deep_2014} attentional mechanisms coupled in CNNs communicate the static convolutional structures with each other generating a sequential processing structure, in which oriented and divided attention systems receive as a context a set of feature maps from the previous time step. It then generates attentional masks on the feature maps of the current time step, choosing between maintaining or not the same attentional weights on the new feature maps. However, most convolutional structures are only divided attention, mainly targeting feature maps. \textbf{Attention gated networks} \cite{schlemper_attention_2018} presents a classic example of purely divided mechanisms. They receive a set of feature maps as input and return a spatial attentional mask over all the maps simultaneously. In this case, the attentional system is focused on spatial characteristics, weighting the same regions of different maps with the same weight, but on each map, the mask weighs each pixel with different weights. Some mechanisms are also quite flexible concerning architecture and temporal characteristics. For example, \textbf{Structured Attention Networks} \cite{kim_structured_2017} features divided mechanisms that can be coupled in sequential architectures operating as oriented or sustained attention.

Some approaches use the oriented component to shift the focus on neural structures and not directly on the data. The \textbf{Attentional Correlation Filter Network} \cite{choi_attentional_2017} uses an oriented and select component to choose each time step a different set of filtering validation strategies for the input image. \textbf{Modality Shifting Attention} \cite{kim2020modality} uses an oriented and select component to decide between using the visual modality or the linguistic modality. Few models have sustained attention. To our knowledge, only Tang et al. \cite{tang_learning_2014} proposed an attentional system \textbf{selected and sustained} capable of tracking a specific face in a scene amid occlusion situations, sudden changes in rotation, scale, and perspective.

\subsection{Selective Perception versus Selective Cognition} 
\label{sec:selectiveperception}

Attention can focus on things other than the sensory stimuli that come through the senses. It can address mental processes, such as memories, thoughts, mental calculations, etc. When the focus is on the external environment, it can also be called selective perception, and when focused on the internal environment, it can be called selective cognition. We consider that perceptual selection occurs when the attentional subsystem receives external sensory stimuli. In this case, the attention acts between the raw data and the neural network reinforcing the perception. Selective cognition, the set $\tau_{t} $is information in the latent space (i.e., memory data, data embeddings, feature data, or hidden states). Despite the classic studies of attention in sensory perception, Deep Learning approaches focus on cognitive selection on hidden states/embedding vectors, external memory, and feature maps, as shown in Figure \ref{table:main_models} in f4 and f5. The first attentional mechanism in the area, proposed for RNNSearch \cite{bahdanau_neural_2014}, is cognitive selection on encoder hidden states. The goal is to weight a dynamic context vector based on the words previously generated. Following this line, countless other cognitive mechanisms have been developed to deal with long-distance dependencies between internal memory structures in the encoder-decoder frameworks and even access memories external to the network. Subsequently, mechanisms for hierarchical alignment appeared \cite{yang2016hierarchical}, multimodal alignment \cite{xu_show_2015}\cite{kim2020hypergraph}, boost features \cite{hu_squeeze-and-excitation_2017}\cite{chen20182}, feature embedding \cite{chen2016latent} , and fusion information \cite{liu_global_2017} all using internal information from the neural network. The main existing perceptual selection mechanisms are focused on computer vision tasks, bringing some inspirations from the theories of human visual attention \cite{mnih_recurrent_2014}.

\subsection{Stimulus' Nature}
\label{sec:space_based}

According to the nature of the stimulus, attention can be task-oriented or time-oriented if the target is a program (i.e., neural network), and it can be space-based versus object-based if the target is a data set.

\subsubsection{Space-Based versus Object-Based Models} There is no consensus on the perceptual scale served by attention: Do we attend to stimulus locations, features, or objects? In the last 50 years, behavioral studies have broadly demonstrated the modulation of attention in several perceptual domains, including space, features, objects, and sensory modalities. The current belief is that attention can be deployed for each of these units, implying that there is no single attentional unit. However, classical studies' main domain is spatial (i.e., location-based attention), which has been the focus of intense research since 1970. This focus is not accidental - vision is an inherently spatial sense, and the first cortical stages of visual representation are spatially organized. Many important studies have documented spatial attention in modulating neural activities in the extra-striated cortex \cite{yantis2008neural}.

Subsequently, other attentional selection domains that are not strictly spatial - feature-based and object-based became the focus of investigation in classical literature. Feature-based attention refers to selecting stimuli based on the values expressed within a specific feature dimension (e.g., yellow in the color dimension and left in the movement dimension). Saenz et al. \cite{saenz2002global} did experiments on humans using functional magnetic resonance imaging (fMRI) and observed that the magnitude of neural responses depends on the stimulus features in conjunction with spatial aspects to select spatial and non-spatial sensory information relevant to the task. While location and feature-based attention are widely studied, object-based has been the focus of behavioral research only for the past 25 years \cite{yantis2008neural}. Studies have revealed that attention can be directed to one or two spatially overlapping objects. O'Craven et al. \cite{o1999fmri} showed observers a spatial overlap between house and face. At any time, the test subjects should look at the house or face. The authors observed brain activity in selective cortical regions of faces and selective at home that depended on which of the two stimuli was attended to. They found that the magnitude of the motion-triggered signal in the MT area also depended on whether the assisted object was moving or not, suggesting that all assisted objects' features were selected.

In our framework, location-based attention is an subsystem focused on stimulus localization existindo um peso atencional para cada estímulo do focus target, ou seja $a_{t,j}^{k} \in \mathbb {R}$. The feature-based attention is a subsystem focused on the target features, ou seja the focus output $a_{t,j}^{k} \in \mathbb {R}^{F_{\tau k}}$. In object-based attention an subsystem capable of focusing on the focus target's semantic elements. For example, the focus target $ \tau_{t}^{k} $ at time $\emph{t}$ maybe represent the set of $n_{\tau^{k}}$ pixels $px$, and attention subsystem select only the object. The focus output set $A = \left \{x \in \mathbb{Z}: 0 \leq x \leq 1 \right \} $, $a_{t}^{k} \in \mathbb {R}^{n_{\tau k}}$, and each position of $a_{t, j}^{k}$ is 1 only if the pixels are inside object.

In this context, most models are data-driven and are exclusively location-based, as shown in Figure \ref{table:main_models} in f6. Its main targets are the hidden states/embedding vectors, feature maps, external memory cells, and raw data from the neural network input stimuli. When the mechanisms are focused on hidden states or embedding vectors, the attentional weights assist in the construction of dynamic context vectors minimizing information bottleneck problems in multimodal approaches \cite{xu_show_2015} \cite{kim2020hypergraph}, encoder-decoder structures \cite{bahdanau_neural_2014}, and embedding representation problems \cite{chen2016latent}. Similarly, when applied to external memory cells, the mechanisms were able to different memory locations to build a dynamic vector of summarizing past experiences from the task's current context. When location-based mechanisms are applied on feature-maps, they adjust the output of feature extractors to make target regions stand out in the presence of disturbing backgrounds seeking to imitate mechanisms in some regions of the human visual cortex. However, few location-based mechanisms focus directly on input stimuli from the neural network. Although there is a wide range of research in psychology and neuroscience focusing on spatial aspects of human vision, only \textbf{RAM} \cite{mnih_recurrent_2014} (Section \ref{sec:ram}), \textbf{DRAW} \cite{draw} (Section \ref{sec:draw_network}), \textbf{Spatial Transformer} \cite{jaderberg2015spatial} and similar approaches explore location-based attention models on raw input image data. Spatial Transformer presents a particularly interesting approach, based on feature maps, the \textit{localization network} determines the transformation parameters, which act as a context for the attentional system to select a local grid on the feature maps and apply the learned transformation minimizing the deformations of the focus region for the convolutional network in a classification task.

There are few purely feature-based, object-based, or hybrid approaches. Purely feature-based mechanisms are more common on feature maps in convolutional neural networks \cite{hu_squeeze-and-excitation_2017} \cite{chen20182} and in graph neural networks \cite{velickovic_graph_2018} \cite{wang_heterogeneous_2019} seeking to adjust the properties of features from global information or the vicinity of the target stimuli, in an attempt to highlight those that are most relevant to the target task. There are still few hybrid approaches with location and feature-based mechanisms in two stages of processing: 1) In the first stage, the features are weighted on the target, building useful context vectors for the second stage; 2) In the second stage, location-based mechanisms use the iteration over the features to guide the attentional focus on the location of the stimuli. There are also very few object-based approaches in the field. To the best of our knowledge, only Tang et al. \cite{tang_learning_2014} proposed an object-centered visual attention approach to generative models inspired by \textbf{Shifter Circuit Model} \cite{olshausen1993neurobiological}. It is a biologically plausible visual attention model to form invariant representations of scale and position of objects in the world. By controlling neurons driven by memory, it controls synaptic forces that guide the flow of spatially organized information in the primary visual cortex (V1) to upper cortical regions, allowing assisted objects to be represented in an invariant way in position and scale. Similarly, the model uses an attentional system with a retina-like representation to search for faces in a scene guided by signals present in associative memory. A series of canonical transformations learned during training help guide the focus on the image, selecting the patch of the face of interest. In a second stage, auxiliary mechanisms propagate the resulting signals to an associative memory represented by a Gaussian Deep Belief Networks (DBN) \cite{hinton2009deep}.

\subsubsection{Task-oriented versus Time-oriented Models}

When the target is a program, the attentional mechanism intuitively seeks to answer the following questions: Among all these networks, which should be chosen to perform an answer/task? How much computing time should be spent for each neural structure ?. The task-oriented selection subsystem chooses one (or more) programs to be executed next. We can consider the target set $T$ to be the set of possible $N$ programs and attention to select the most appropriate for the task in time $t$. Time-oriented selection attention chooses how much computing time to allocate to each program given a computational time budget. For example, the framework contains several neural networks $\tau_{t} = \{\tau_{t}^{1}, \tau_{t}^{2}, \tau_{t}^{3} \}$to be executed. The attention subsystem must decide how much computation to spend, from the budget $B$, on each neural network $\tau_{t}^{k}$. The focus output is $a_{t} = \{a_{t}^{1}, a_{t}^{2}, a_{t}^{3} \}$, $a_{t}^{1} + a_{t}^{2} + a_{t}^{3} = 1$, and the amount of computation for each program can be calculated as $a_{t}^{k} B$.

Few architectures in the area target a neural network (Figure \ref{table:main_models} in f9 and f10). \textbf{Adaptive Computation Time (ACT)} \cite{graves_adaptive_2016} chooses how many auxiliary computing substeps will be performed by a recurring structure in a $t $time frame. The structure decides how a computing budget will be allocated, controlling when the recurring structure should stop and generate the final output $y_{t} $based on the auxiliary outputs. The \textbf{Attentional Correlation Filter Network} \cite{choi_attentional_2017} directs the attentional focus on a different set of feature extractors based on previous deep regression network validation scores, determining which set of extractors each time step $t$ must be activated to receive the input image stream. The previous validation scores work as context information giving feedback to the attentional system about the performance of the general system in the task, from that feedback attention can regulate the focus points for the next iteration. \textbf{Modality Shifting Attention} \cite{kim2020modality} has a task-oriented mechanism responsible for shifting attention between the neural network that captures visual sensory stimuli and the neural network that captures linguistic stimuli. The system is guided by a context represented by the question about a sequence of images and captions for question-answering tasks. In \textbf{STRAW} \cite{mnih2016strategic} a task-oriented attentional mechanism controls the activation of read, write and the time-advance structure over the action-plan and the commitment-plan controlling when the data-oriented attentional mechanisms must on updating a trained agent's plans via Reinforcement Learning.

\subsection{Bottom-Up versus Top-Down Models}
\label{sec:bottom-up}


A big difference between the models is whether they depend on
bottom-up influences, top-down influences, or a combination of both. \textbf{Bottom-up} or involutory attention is determined by the characteristics of the input stimuli (ie, \textbf{stimulus-driven}), while \textbf{top-down} cues are determined by cognitive phenomena such as knowledge, expectation, reward and current goals (ie, \textbf{goal-driven}). Stimuli that attract attention in a bottom-up manner are sufficiently distinct concerning the surrounding characteristics. As a result, attention is exogenous, automatic, reflective, and feed-forward. A typical example of
bottom-up attention is to look at a scene with just one
horizontal bar between several vertical bars where attention
it is immediately directed to the horizontal bar. On the other hand, top-down attention to be a voluntary process deliberated by the individual, in which a particular location, feature, or object is relevant to current behavioral goals. Such a process is guided by elements of a high semantic level, such as motivation, expectation, private interests, rewards, and social motivations \cite{colombini2014attentional}. In our framework, bottom-up attention is guided by the focus target's discrepancies, whether or not there is a contextual entry $c_{t} $. However, the context comes from information from the target's own set of stimuli. If the attention is guided by contextual information from other sensory sources, external or previous memories, the attention is top-down.

Additionally, we consider an additional classification regarding the presence of context or the previous inner state of attention. In this sense, the bottom-up and top-down mechanisms can still be \textbf{stateful} or \textbf{stateless}. In stateful the attentional subsystem considers context and inner information as part of the input set (ie, $i_{t-1} $$\neq $$\emptyset $or $c_{t} $$\neq $$\emptyset $), otherwise the subsystem implements the stateless selection. The \textbf{bottom-up stateless} mechanisms have no previous context or inner states of attention as part of the input, so the attentional focus is assigned only through the internal extraction of the target's discrepancies. The \textbf{bottom-up stateful} mechanisms can have contextual and previous inner state-input simultaneously, or just one of the options. However, the context extracted externally comes from the current target. In the inner state, it represents the previous state of attention on the same target. That is, the target does not change over time. There are no \textbf{top-down stateless} mechanisms since a condition for the existence of top-down influences is the presence of a context external to the current target. For the existence of \textbf{top-down stateful} mechanisms, the presence of context as input is mandatory, with the presence of the previous inner state being optional. Besides, the context refers to previous memories, external memories, or elements from other sensory sources other than the target.

The mechanisms explore top-down and bottom-up influences seeking to answer one of two questions: 1 intuitively) Where to look at the target given the alignment observed between it and the context? 2) In the absence of context, where should I look, given the discrepancies and similarities between the target elements? Some models answer the first question by extracting the context directly from the target, that is, using bottom-up influences. The encoder stack of \textbf{Neural Transformer} \cite{vaswani_attention_2017} (Section \ref{sec:neural_transformer}) models a bottom-up attentional system completely guided by low-level contexts extracted directly from the target of the attentional system. In each encoder, an input $I$ is decomposed into a set of queries, keys, and values extracted in parallel. In the initial stage of attention, queries act as a low-level context and keys as a target, producing attentional masks. In the second stage, these masks and the values make up the context vector to guide the attentional focus on the original $ I $ input that will be input to a new attentional encoder, resulting in a chain of bottom-up attentional systems with a low-level context.

Typically, attention-based graph neural networks use a combination of bottom-up mechanisms to guide attentional focus with and without context.
\textbf{Graph Attention Networks} \cite{velickovic_graph_2018} (Section \ref{sec:gats}) targets a set of neighboring nodes and from the discrepancies between that set defines different attentional weights for each feature. In the second stage, this answer is used as a context by another bottom-up and location-based subsystem on the set of neighboring input nodes. The final composition between the attentional map and all nodes generates an embedding with the neighborhood's representativeness for a specific node. Similarly, \textbf{Heterogeneous Graph Attention Network} \cite{wang_heterogeneous_2019} uses two bottom-up attention mechanisms at the node level and bottom-up mechanisms at the semantic level to capture various types of semantic information between heterogeneous graphs.

In convolutional networks, the bottom-up mechanisms are present when the target is a set of feature maps. Usually, only inter-channel or intra-channel discrepancies guide attention by promoting boost features or recalibrating channels. The attentional mechanism of \textbf{Squeeze-and-Excitation Networks} \cite{hu_squeeze-and-excitation_2017}, a pioneering approach in the area, receives as a focus target a set of feature maps, generates channel-wise statistics using global average pooling. Based on these statistics, it captures channel-wise dependencies capable of learning non-linear iterations between channels. Finally, re-scale each input channel in a feature-based approach. \textbf{Double Attention Networks} \cite{chen20182}, in a similar but location-based approach, uses as feature context maps from the same convolutional layer (i.e., bottom-up) or different layers (i.e., top-down) as a context for defining relations between the elements, whose main objective is based on the extracted relations to ponder the regions of the feature maps of a convolutional layer. Finally, the \textbf{SNAIL} \cite{mishra_simple_2017} via bottom-up captures temporal relationships between feature maps in a meta-learner approach using self-attention mechanisms similar to Transformer.

However, most area approaches are top-down stateful, as shown in Figure \ref{table:main_models}. The first mechanism was proposed for \textbf{RNNSearch} \cite{bahdanau_neural_2014} in mid-2014, in which the context received as input is directly responsible for the dynamic change of the context vector $c_{t} $received by the decoder at each time step $t $. Similarly, Xu et al. \cite{xu_show_2015} used information from the decoder's previous hidden state to guide a location-based mechanism through regions of the input image in a multimodal question-answering approach. Networks with internal and external memory are usually top-down structures. \textbf{End-to-End Memory Networks} \cite{sukhbaatar2015end} uses an external query for a question as a context to search and focus on memory cells more related to the question's content. \textbf{Neural Turing Machine} \cite{graves_neural_2014} uses as context parameters issued by the network controller that define content-based and location-based addressing in external memory. \textbf{Sparse Attentive Backtracking} \cite{ke_sparse_2018}\cite{ke2017sparse} applies an attentional sparse memory recovery method to build the next hidden state of an RNN, using the provisional hidden state $\widehat{h}^{(t)}$ and target all previously processed memories. The mechanism acts as a method capable of blaming or giving credit to previous memories, similarly to what human beings do, without the need to repeat all events from the present until the credited event, managing to capture long-distance dependencies between states efficiently. Zhang et al. \cite{zhang2018top} used top-down attention to propose \textbf{Excitation Backprop} - a new backpropagation scheme based on biological evidence from \textbf{Winner-Take-All (WTA) competition} \cite{lee1999attention} among visual filters and the \textbf{Selective Tuning Model} \cite{tsotsos1995modeling} selectively adjust a visual processing system through a top-down hierarchy of winner-take-all processes embedded in the visual processing pyramid. Similarly, Excitation Backprop uses a probabilistic WTA on CNNs to promote excitatory or inhibitory connections between neighboring neurons from top-down influences from previously visited neurons.

Co-attention structures are also typically top-down. There are attentional mechanisms propagating attention in this structure in two ways: from the query to the context and from the context to the query. \textbf{Dynamic Coattention Networks} \cite{xiong2016dynamic} computes attentional scores in this way. The mechanisms produce weights for each word in the question based on document words as context, just as they produce weights for each word in the document based on question words as context. Similarly, \textbf{Hypergraph Attention Networks} \cite{kim2020hypergraph} computes attention between two hypergraphs comparing the semantics between two symbolic representations of different sensory sources.

Some structures are hybrid and use both levels of influence to guide attention. Neural Transformer \cite{vaswani_attention_2017} (Section \ref{sec:neural_transformer}) stands out as the main hybrid structure in the area. It has a completely bottom-up encoder and a hybrid decoder so that the first attentional system is bottom-up and processes the translated words. In contrast, the second is top-down for using the last encoder's attentional information as a context to guide attention on the target, represented by the words previously translated. Following the structure of Transformer, recently \textbf{BRIMs} \cite{mittal2020learning} presented hybrid mechanisms to carry out communication between RIMs modules. Bottom-up attentional subsystems communicate between modules of the same layer, as well as the composition of hidden states in initial layers using the entry $x_{t}$ as the target, and via top-down attention modules in different layers communicate with each other requesting information about hidden states of previous and posterior layers to compose the current hidden state.

\subsection{Continuity: Soft versus Hard}
\label{sec:hard_soft}

The selection may occur either by choosing a discrete subset of the possible choices or by performing a soft (or continuous) giving real-valued scores to the possible choices. The different types of selection can be implemented with modules by appropriately choosing the focus output set$A$. If $A = \left \{x \in \mathbb{R}:0 <x <1 \right \}$ the selection is soft, and$A = \left \{x \in \mathbb{Z}:0 \leq x \leq 1 \right \}$ the selection is hard.

In Deep Learning, the mechanisms are mainly divided into the following categories:1) \textbf{hard attention} determines whether a part of the mechanism's input should be considered or not, reflecting the interdependence between the input of the mechanism and the target of the deep neural network. The weight assigned to an input port is either 0 or 1; 2) \textbf{soft attention} divides attentional weights between 0 and 1 for each input element so that the sum of all weights is equal to 1. It decides how much attention should be focused on each element, considering the interdependence between the input of the deep neural network's mechanism and target; 3) \textbf{self-attention} quantifies the interdependence between the input elements of the mechanism. This mechanism allows the inputs to interact with each other "self" and determine what they should pay more attention to. There are also some secondary categories:1) \textbf{global attention} is a simplification of the classic soft attention proposed for encoder-decoder frameworks; 2) \textbf{local attention} is a tradeoff between hard and soft attention; 3) \textbf{co-attention} assigns attention from both the context to the target and from the target to the context, and 4) \textbf{hierarchical attention} presents mechanisms adapted to deal with hierarchical structures at different levels of granularity. Concerning continuity, the mechanisms classified as hard attention have hard continuity according to our taxonomy, and all other mechanisms in the area have soft continuity.

There is still no systematic study to determine the advantages and disadvantages of hard and soft continuity mechanisms. However, there is a wide range of soft continuity mechanisms, as shown in the Figure \ref{sub:main_architectures} in f13 and f14. One justification is that the hard mechanisms, in most cases, make the architecture non-differentiable, requiring more elaborate training strategies, such as Reinforcement Learning (RL) or even hybrid supervised and RL \cite{mnih_recurrent_2014} approaches. Usually, these strategies are still little explored in computer vision and natural language processing - where the main neural attention models are - because they require the well-designed design of the reward functions, which is not always intuitive or necessary when there is ground truth, available for training via supervised learning. Currently, few architectures in the area use both mechanisms simultaneously. \textbf{Pointer Networks} \cite{vinyals2015pointer} features a soft mechanism for distributing attention over all input elements, followed by a hard mechanism for choosing one as an output at each stage of the decoder. \textbf{Modality Shifting Attention} \cite{kim2020modality} uses a hard mechanism to switch between different sensory modalities and soft continuity mechanisms to reason about the final prediction of the network.

\section{Neural Attention Models}
\label{sub:main_architectures}

\begin{figure*}[htb]
  \centering
  \includegraphics[width=135mm]{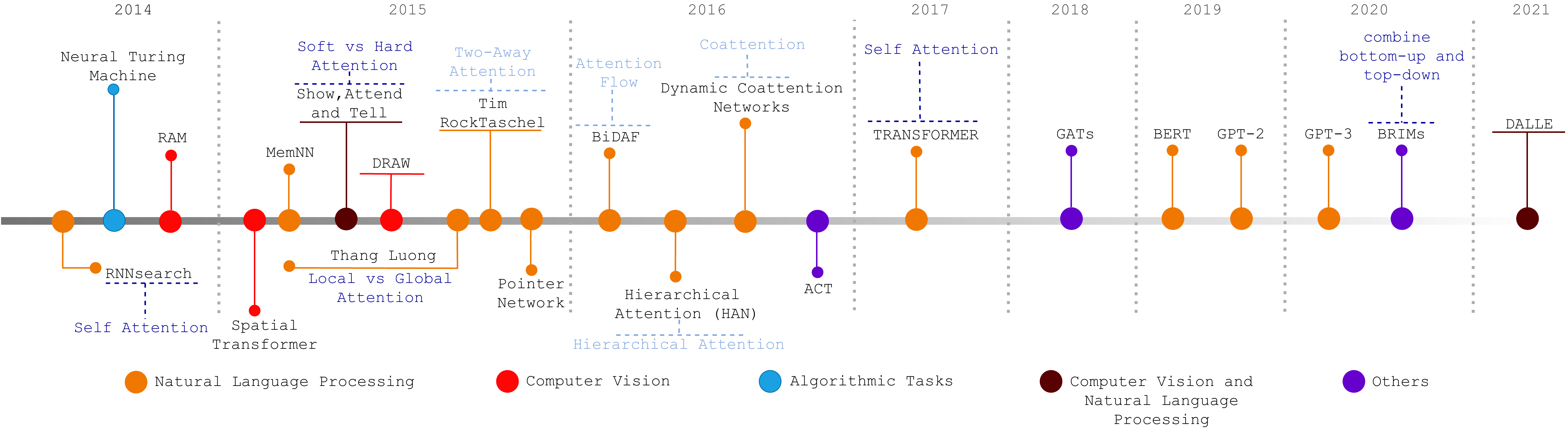}
  \caption{Timeline illustrating the main key developments from 2014 to the present day. RNNSearch \cite{bahdanau_neural_2014} presented the first attention mechanism. Neural Turing Machine \cite{graves_neural_2014} and Memory Networks \cite{weston_2014_memory} introduced memory and dynamic flow control. RAM \cite{mnih_recurrent_2014} and DRAW \cite{draw} learned to combine multi-glimpse, visual attention, and sequential processing. Spatial Transformer \cite{jaderberg2015spatial} introduced a module to increase the robustness of CNNs to variations. Show, attend and tell \cite{xu_show_2015} created attention for multimodality. Pointer Networks \cite{vinyals2015pointer} presented attention as a pointer. BiDAF \cite{seo_bidirectional_2016}, HAN \cite{yang2016hierarchical}, and DCN \cite{xiong2016dynamic} presented attentional techniques to align data with different hierarchical levels. ACT \cite{graves_adaptive_2016} introduced the computation time topic. Neural Transformer \cite{vaswani_attention_2017} was the first self-attentive neural network with an end-to-end attention approach. GATs \cite{velickovic_graph_2018} introduced attention in GNNs. BERT \cite{devlin_bert:_2018}, GPT-2 \cite{radford2019language}, GPT-3 \cite{brown2020language}, and DALLE \cite{unpublished2021dalle} are state of the art in language models and text-to-image generation. Finally, BRIMs \cite{mittal2020learning} learned to combine bottom-up and top-down signals.   
}
  \label{fig:timeline}
\end{figure*}

In this section, we discuss some of the main neural attention models from the theoretical perspective of attention. A timeline summarizing the main developments and their main contributions is shown in Figure \ref{fig:timeline}. In the \ref{sec:rnn_search} section we discuss \textbf{RNNSearch} \cite{bahdanau_neural_2014}, in the \ref{sec:neural_turing_machine} section we discuss \textbf{Neural Turing Machine} \cite{graves_neural_2014}, in the \ref{sec:ram} we discussed \textbf{RAM} \cite{mnih_recurrent_2014}, in the \ref{sec:end_to_end_memory} section we discussed the \textbf{End-to-End Memory Networks} \cite{sukhbaatar2015end}, in the \ref{sec:show_attend_tell} we discussed the \textbf{Show, Attend and Tell} \cite{xu_show_2015}, in the \ref{sec:draw_network} section we discussed the \textbf{DRAW} \cite{draw}, in the \ref{sec:bidaf} section we discussed the \textbf{BiDAF} \cite{seo_bidirectional_2016}, in the \ref{sec:neural_transformer} section a \textbf{Neural Transformer} \cite{vaswani_attention_2017}, finally in the section \ref{sec:gats} we discussed a \textbf{GATs} \cite{velickovic_graph_2018}.

\subsection{RNN Search: the beginning}
\label{sec:rnn_search}

\begin{figure}[htb]
  \centering
  \includegraphics[width=\linewidth]{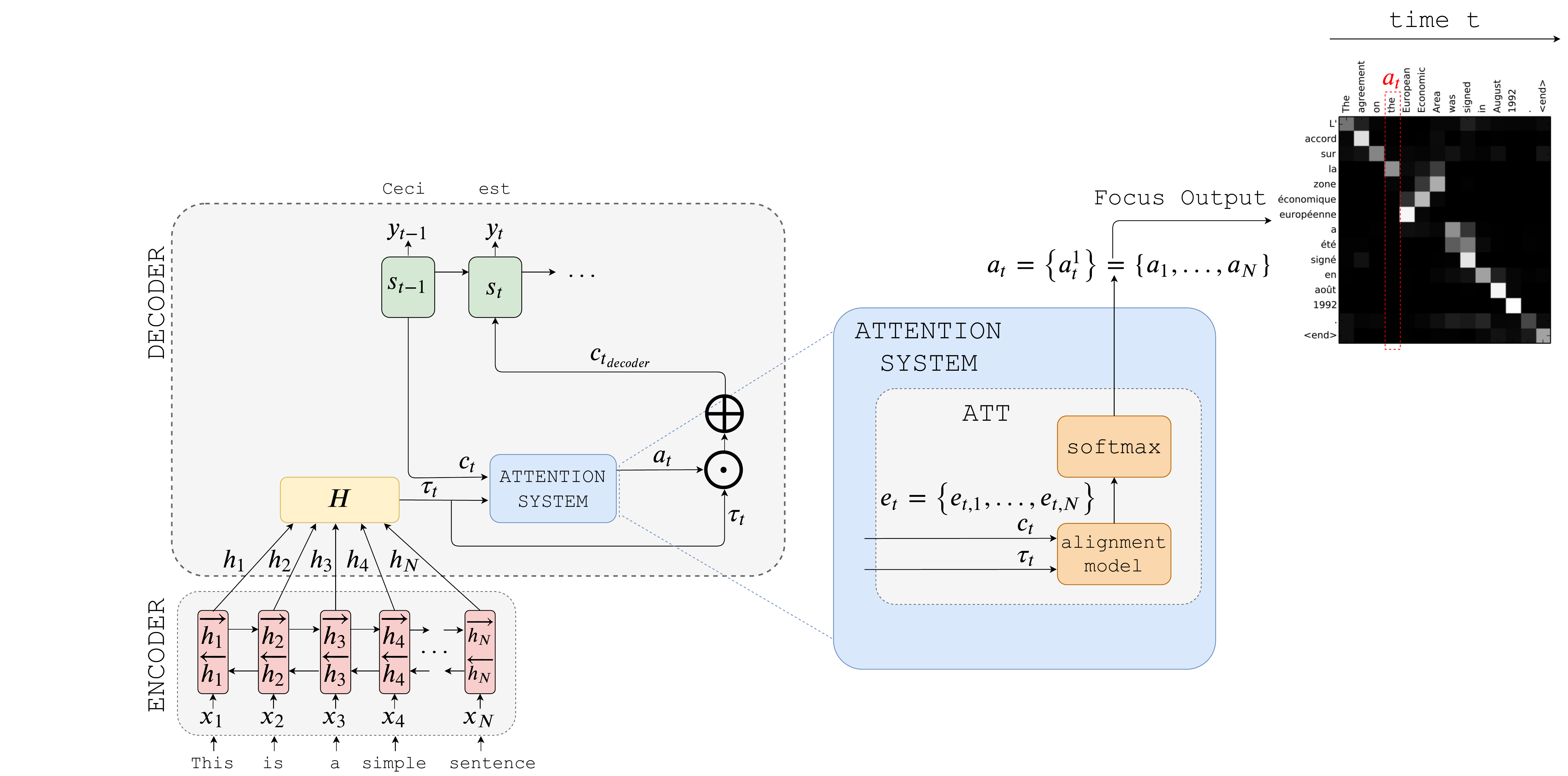}
  \caption[rnn search]{RNNSearch \cite{bahdanau_neural_2014} architecture illustration\footnotemark. The encoder generates a set of hidden states, which are the input for the only attention system. The divided attention shares the focus between the different stimuli, and in a top-down way, it generates a dynamic context vector for a decoder.}
  \label{fig:rnn_search_model}
\end{figure}

RNNSearch \cite{bahdanau_neural_2014} uses attention for machine translation. The purpose is compute an output sequence \textbf{\textit{y}} that is translation from input sequence \textbf{\textit{x}}. The architecture consists of an encoder followed by a decoder, as shown in Figure \ref{fig:rnn_search_model}. The encoder is a bidirectional RNN (BiRNN) that consist of forward and backward RNN's for compute an annotation term $h_{j}$. The forward RNN $\overrightarrow{f}$ reads the input sequence in the order of $x_{1}$ to $x_{N} $ and calculates the forward hidden state sequence ($ \overrightarrow{h_{1}}, .. ., \overrightarrow{h_{N}}$). The backward RNN reads the sequence in the reverse order $\overleftarrow{f}$, (from $x_{N}$ to $x_{1}$), resulting in the backward hidden states sequence ($ \overleftarrow{h_{1}}, .. ., \overleftarrow{h_{N}}$). The annotation $h_{j}$, for each word $x_{j}$, is the concatenation of the $ \overrightarrow{h_{j}} $ and $ \overleftarrow{h_{j}} $ as follows

\begin{equation}
    (h_{1}, ..., h_{N}) = Encoder(x_{1}, ..., x_{N}) 
    \label{encoder_rnn_search}
\end{equation}

\begin{equation}
    h_{j} = [\overrightarrow{h_{j}};\overleftarrow{h_{j}}]^{T}
    \label{annotation_bi_rnn}
\end{equation}

The decoder consists of classic RNN and attention system. The classic RNN calculates from a context vector $c_{t_{decoder}}$ a probability distribution for all possible output symbols:

\begin{equation}
    p(y_{t}|y_{1}, ..., y_{t-1}, x) = RNN_{decoder}(c_{t_{decoder}})
    \label{decoder_rnn}
\end{equation}


The attentional system has only one subsystem (Figure \ref{fig:rnn_search_model}) that receives information from a single sensory modality (i.e., textual). At each time step \textit{t}, the subsystem takes as input a contextual $c_{t} = \left \{ c_{t}^{1} \right \} = \left \{ c_{t,1}^{1} \right \} = \left \{ s_{t-1} \right \}$, a focus target $\tau_{t} = \left \{ \tau_{t}^{1} \right \} = \left \{ \tau_{t,1}^{1}, ..., \tau_{t,N}^{1} \right \} = \left \{ h_{1}, ..., h_{N} \right \}$, and produces attention weights $a_{t} = \left \{ a_{t}^{1} \right \} = \left \{ a_{t,1}^{1}, ..., a_{t,N}^{1} \right \} = \left \{ a_{1}, ..., a_{N} \right \}$ as output, where $s_{t-1} \in \mathbb{R}^{1 \times d}$ is the decoder's previous hidden state, $h_{j} \in \mathbb{R}^{1 \times d}$ is encoder annotation vector, $a_{t}^{1} \in \mathbb{R}^{1 \times N}$ are the weights of attention over all the encoder's annotation vectors. The focus target is processed by alignment function $e_{t,j} = a(s_{t-1}, h_{j})$ to obtain a set scores $e_{t,j}$ that reflects the importance of $h_{j}$ with respect $s_{t-1}$ in deciding the next state $s_{t}$ and generating $y_{t}$. The alignment function $a$ is a feedforward neural network which is jointly trained with the framework. The scores are normalized through a softmax function to obtain attention weights $a_{j} = \frac{e^{e_{t,j}}}{\sum_{j = 1}^{N}e^{e_{t,j}}}$. Finally, a weighted sum over enconder's hidden states generates the dynamic context vector $c_{t_{decoder}} = \sum_{j=1}^{N} a_{j}h_{j}$ $ \in \mathbb{R}^{1 \times d}$.

Intuitively, attention decides which parts of the source sentence pay attention. Allowing the decoder to have this mechanism is unnecessary to encode a context vector of fixed size. Instead, the information is spread throughout the annotation sequence and can be selectively retrieved by the decoder as needed. 
By assigning attention simultaneously and continuously to each $h_{j}$, the \textbf{selection is soft and divided}. \textbf{Top-down stateful} since a previous decoder's state represents a context. \textbf{Location-based} since the purpose of attention is to weigh the stimuli $h_{j}$ by assigning the same weight to all features. Finally, the
the system is \textbf{cognitive} and \textbf{oriented} for differently selecting the same focus target in latent space at each time step $t$.

\footnotetext{\url{https://github.com/larocs/attention_dl/blob/master/imgs}}

\subsection{Neural Turing Machine: An attention-augmented memory approach}
\label{sec:neural_turing_machine}

\begin{figure}[htb]
  \centering
  \includegraphics[width=\linewidth]{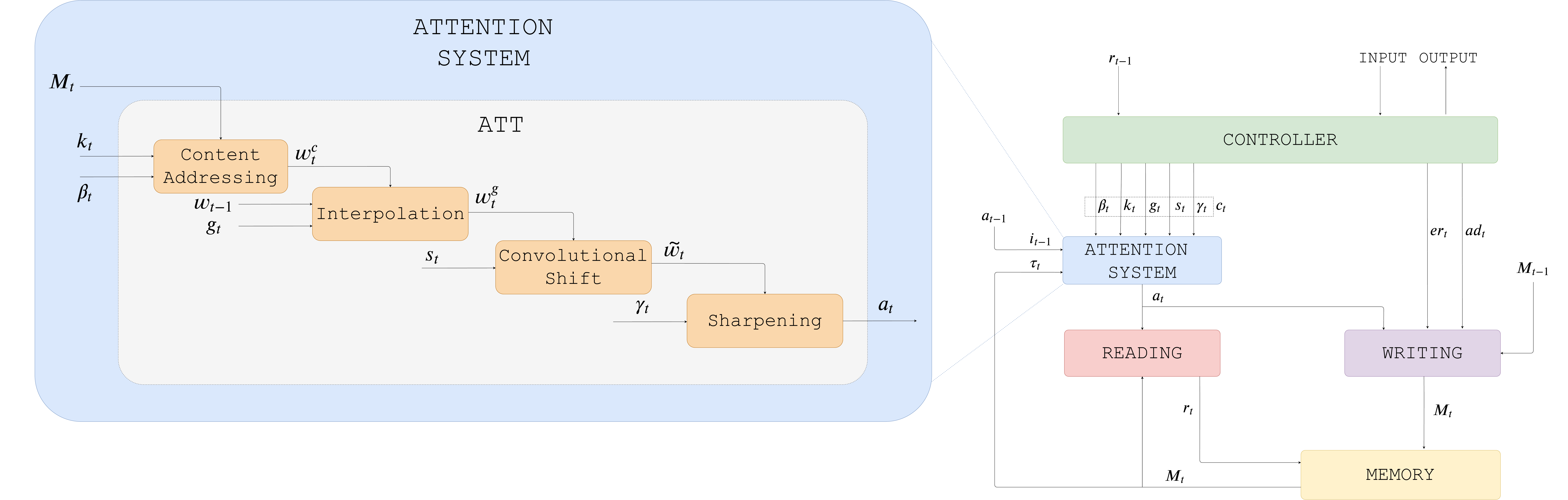}
  \caption[turing]{Neural Turing Machine \cite{graves_neural_2014} architecture illustration. The architecture has a controller, an external memory, read/write heads, and an attentional system. At each time step $t$, the attentional system, guided by the controller's parameters, defines the memory locations that will be read and written. The system is oriented, divided, location-based, top-down stateful, cognitive and soft.}
  \label{fig:turing_model}
\end{figure}

Neural Turing Machine \cite{graves_neural_2014} uses attention for algorithmic tasks. The architecture consists of a controller, heads, external memory, and attention system, as shown in Figure \ref{fig:turing_model}. The controller is a feedforward or recurrent network, which interacts with the outside world and the attentional system to manipulate memory through reading and writing heads. The memory $M_{t} = \left \{M_{t,0}, ..., M_{t,N-1} \right \}$ $\in$ $\mathbb{R}^{N \times M}$ is an matrix, where $N$ is amount number of memory cells, and $M$ is amount number of memory cell's features. The attentional system uses the controller parameters to define each reading/writing operation's focus determining the importance degree at each location. So, a single head can attend an individual cell or weakly in several cells.

At each time step $t$, the system receives as input a focus target $\tau_{t} = \left \{ \tau_{t}^{1} \right \} = \left \{ \tau_{t,1}^{1}, ..., \tau_{t,N}^{1} \right \} = \left \{ M_{t,0} , ..., M_{t,N - 1} \right \}$ with all memory content, and a contextual input $c_{t} = \left \{ c_{t}^{1} \right \} = \left \{ c_{t,1}^{1}, c_{t,2}^{1}, c_{t,3}^{1}, c_{t,4}^{1}, c_{t,5}^{1} \right \} = \left \{ k_{t}, \beta_{t}, g_{t}, s_{t}, \gamma_{t} \right \}$ with controller outputs, where $k_{t} \in \mathbb{R}^{M}$ is the key vector, $\beta_{t} \in \mathbb{R}$ is the key strength, $g_{t} \in \mathbb{R}$ is the interpolation gate, $s_{t} \in \mathbb{R}^{2k+1}$ is the shift weight, and $\gamma_{t} \geq 1$ is the sharpening weight. Unlike most of the attentional systems, this system receives as input the past inner state $i_{t-1} = \left \{ a_{t-1} \right \} $, which is the attentional mask in the previous time. The attentional system acts as an interface between the controller and read/write heads. It has two addressing steps - the first focuses on content, and the second is focusing on location. This structure is very similar to VOCUS \cite{frintrop2006vocus}, an classic visual attention model proposed by Fintrop in 2006. Content addressing resembles the bottom-up step, and the subsequent processing is similar VOCUS top-down step.

Content addressing is inspired by the Hopfield Network, but with a simple retrieval mechanism. It is based on similarity between the memories and an approximation vector issued by the controller. Specifically, the key vector $k_{t}$ is compared to each vector $M_{t,i}$ by a similarity function $ K\left ( k_{t}, M_{t,i} \right ) = \frac{k_{t}M_{t,i}}{\left \| k_{t} \right \|\left \| M_{t,i} \right \|} $, producing content addressing weights $w_{t}^{c} \in \mathbb{R}^{N}$, composed by $w_{t,i}^{c} = \frac{e^{\beta_{t}K \left ( k_{t},M_{t,i}\right )}}{\sum_{j=0}^{N-1}e^{\beta_{t}K\left (k_{t},M_{t,j}\right )}}$, where $K$ is the cosine similarity.

Content addressing is very efficient, but in some tasks, it needs a recognizable spatial address. Location-based addressing facilitates simple iteration with three main steps - interpolation, convolutional shift, and sharpening. Interpolation controls the use of the content-based addressing mask. The gate $g_{t} \in \left [ 0,1 \right ]$ combines the past inner state $a_{t-1}$ with the $w_{t}^{c}$. If the gate is zero, the content weighting is ignored. If the gate is one, the previous attentional mask is ignored, and the system uses only content-based addressing. After interpolation, the convolutional shift allows the current focus to change and serve adjacent memory locations. This mechanism, is a one-dimensional circular convolution, where the shifting weight $s_{t}$ is the kernel to be convolved on the output's interpolation:

\begin{equation}
    w_{t}^{g} = g_{t}w_{t}^{c} + \left ( 1 - g_{t} \right )a_{t-1}
    \label{reward_3}
\end{equation}

\begin{equation}
\tilde{w_{t,i}} = \sum_{j=0}^{N-1}w_{t}^{g}\left ( j \right )s_{t}\left ( i-j \right )
    \label{reward_4}
\end{equation}

\noindent where $g_{t} \in \mathbb{R}$ is the interpolation gate, $a_{t-1} \in \mathbb{R}^{N}$ are the attention weights of the previous time step, $s_{t}$ is a normalized distribution over the allowed integer displacements. 

Intuitively, $s_{t}$ represents the displacement instructions. If only one offset position is allowed (i.e, $k=1$), $s_{t}$ will be a vector consisting of 3 elements, which can be interpreted by following instructions $\left \{ \right.$ \textit{shift 1 forward}, \textit{maintain focus}, \textit{shift 1 backward} $\left.  \right \}$. In the general case, $s_{t}$ will have $2k + 1$ elements, where $k$ is the highest absolute displacement value. To avoid very high dispersions, the sharpening step takes $\tilde{w_{t}} $ and $\gamma_{t}$ to adjust the sharpness of the weights generating the final attention mask $a_{t} \in \mathbb{R}^{N}$, composed by weights $a_{t,i} = \frac{\tilde{w_{t,i}}^{\gamma^{t}}}{\sum_{j=0}^{N-1}\tilde{w_{t,j}}^{\gamma^{t}}}$ for each memory cell $i$.

The reading head takes the attentional reading mask and the memory for generate as output $r_{t} \leftarrow \sum_{i}^{N} a_{t,i}M_{t,i} \in \mathbb{R}^{M}$, defined as a convex combination of memory cells. Similarly, the writing head takes the attentional writing mask to erases and adds data in memory. The elements in memory cell $i$ are reset to zero if $a_{t,i}$ and the erase element are one; if $a_{t,i}$ or the erase is zero, the memory will not be changed. When multiple heads are present, erasures can be performed in any order, as multiplication is commutative. The combined erase and add operations of all writing heads produce the final memory contents in time $t$. Specifically,

\begin{equation}
    \tilde{M}_{t,i} \leftarrow M_{t-1,i}\left [ 1 - a_{t,i}e_{t} \right ]
    \label{reward}
\end{equation}

\begin{equation}
    M_{t,i} \leftarrow \tilde{M}_{t,i} + a_{t,i}add_{t} 
    \label{reward}
\end{equation}

\noindent where $M_{t,i} \in \mathbb{R}^{M}$ is the memory cell $i$, $\tilde{M}_{t,i} \in \mathbb{R}^{M}$ is the memory cell $i$ with the content deleted, $a_{t,i} \in \mathbb{R}$ is the attentional weight for memory cell $i$, $e_{t} \in \mathbb{R}^{M}$, such that $e_{t} = \left \{ x \in \mathbb{R}: 0 \leq x \leq 1\right \}$ is the erasure vector, $add_{t} \in \mathbb{R}^{M}$ is the vector with content to be added in memory, 1 is a row-vector of all 1-s, and the multiplication against the memory cell is point-wise. Note that both the delete and add vectors have independent $M$ components, allowing for refined control over which elements in each cell location are modified.

NTM's attentional system has \textbf{location}, \textbf{soft}, \textbf{oriented} and \textbf{divided} properties, since weights attentional mask respect the constraint $\sum_{i=1}^{N-1}a_{t,i} = 1$ e suas intensidades em cada local se alteram no tempo. The system is \textbf{top-down stateful} por ser influenciado por parâmetros estimados pelo controlador, and \textbf{cognitive} for atuar sobre os heads. Este mecanismo introduziu duas características importantes da cognição humana: estruturas de ligação variável e o processamento procedural. 
\subsection{Recurrent Attention Model (RAM): A visual attention system for image classification}
\label{sec:ram}

\begin{figure}[h]
  \centering
  \includegraphics[width=\linewidth]{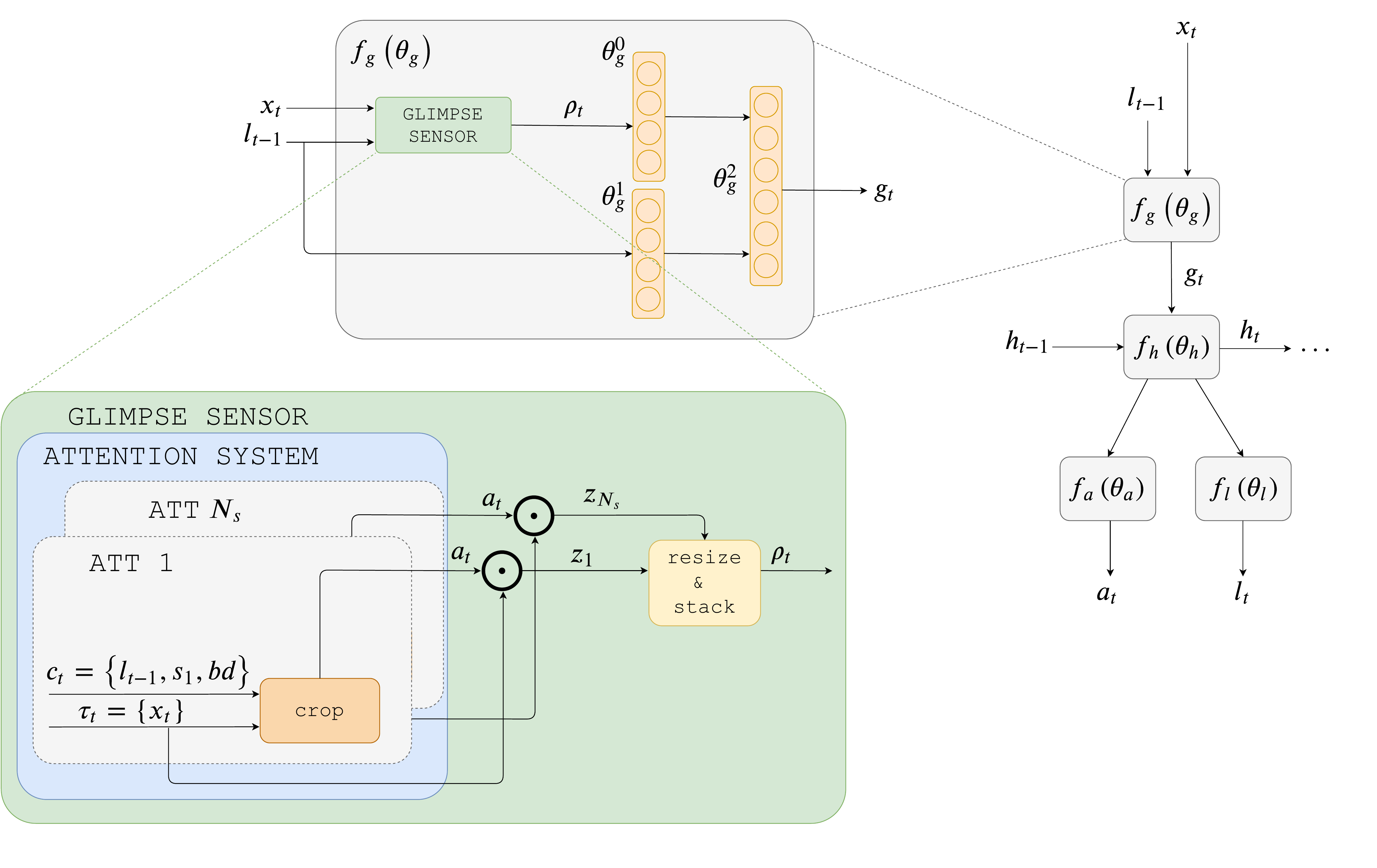}
  \caption[ram]{RAM \cite{mnih_recurrent_2014} architecture illustration. The architecture presents a perceptual selection system within the glimpse sensor to generate a retina-like representation. The core network captures this representation and synthesizes a historical composition between the current and previous steps. The action and location network uses this historical summarization to determine the next action and focus via the reinforcement learning paradigm.}
  \label{fig:ram_model}
\end{figure}

RAM \cite{mnih_recurrent_2014} uses attention for image classification. The architecture consists of the attention system, glimpse sensor, glimpse network, location network, core network, and action network (Figure \ref{fig:ram_model}). First, the glimpse sensor extracts a retina-like representation $\rho_{t}$ around localization $l_{t-1}$. It encodes the region around $l_{t-1}$ in high-resolution, and progressively uses a low-resolution representation for the farthest pixels of $l_{t-1}$. At each time step $t$, the attention system inside glimpse sensor selects $N_{s}$ square patches centered in $l_{t-1}$. The first path being $g_{w} \times g_{w}$ pixels in size, and each subsystem produces successive patches having twice the width of the previous.

The attention system is similar classic visual attention approches with programmed microsaccades designed in the 1990s \cite{itti1998model}. Itti et al. \cite{itti1998model} presented the first practical approache to microsaccades based on a competitive structure -- \textbf{Winner-takes-all (WTA)} mechanisms jointly \textbf{inhibition and return} strategy \cite{itti1998model}. Differently, RAM uses reinforcement learning in a sequential structure to determine the best policy for microsaccades, and uses only \textbf{top-down stateful} attention to modulate the focus, while the classic systems have even explored bottom-up, top-down and hybrid approaches. The RAM is an exception with \textbf{selective perception}, \textbf{hard}, \textbf{selective} and \textbf{oriented} attention simultaneously. It adopts a \textbf{location-based} system while hybrid approches with feature-based and location-based are biologically plausible and widely used in historical models. In most classic visual attention systems was common an feature-based approach to modulating low-level features (i.e., color, intensity, and orientation), and after merging all stimulus an location-based attention finds the regions to be attended.

With some divergences from the classic visual attention models, RAM presents several attentional subsystems in parallel, where each takes the same focus target as input $\tau_{t} = \left \{ \tau_{t}^{1} \right \} = \left \{ \tau_{t,1}^{1}, ..., \tau_{t,N}^{1} \right \} = \left \{ x_{t} \right \}$, where $x_{t}$ is input image, and a different scale factor in each contextual input $c_{t} = \left \{ c_{t}^{1} \right \} = \left \{ c_{t,1}^{1}, c_{t,2}^{1}, c_{t,3}^{1} \right \} = \left \{ l_{t-1}, s_{i}, bd \right \}$, where $l_{t-1} \in \mathbb{R}^{2}$, $s_{i} \in \mathbb{R}$ is the scale factor $i$, $bd \in \mathbb{R}$ is the sensor bandwidth. And produces as output an attention mask  
$a_{t}$ = $\left \{ a_{t}^{1} \right \}$ = $\left \{ a_{t,1}^{1}, ..., a_{t,N}^{1} \right \}$ = $\left \{ a_{1}, ..., a_{N} \right \}$, where $a_{i}$ is one if the pixel inside in focus patch and zero otherwise. The masks $a_{t}$ select the patches and then they are scaled to the sensor bandwidth dimensions and stacked to produce retina-like representation $\rho_{t}$. The glimpse network $f_{g}$ combines $\rho_{t}$ and $l_{t-1}$ to produce the glimpse feature vector $g_{t} = Rect\left ( Linear\left ( h_{g} \right ), Linear\left ( h_{l} \right ) \right )$, where $h_{g}$, $h_{l} \in \mathbb{R}^{128}$, $g_{t} \in \mathbb{R}^{256}$, 
$h_{l} = Rect\left ( Linear\left ( l_{t-1} \right ) \right )$, and 
$h_{g} = Rect\left ( Linear\left ( \rho\left ( x_{t},l_{t-1} \right ) \right ) \right )$. Let 
$Linear\left ( x \right ) = Wx+b$  for some weight matrix $W$ and bias vector $b$, 
$Rect\left ( x \right ) = max\left ( x,0 \right )$ be the rectifier nonlinearity.

The core network $f_{h}$ receives as input the glimpse encoding $g_{t}$, the previous internal state $h_{t-1}$, and outputs the current internal state $h_{t} = f_{h}\left ( h_{t-1}, g_{t}, \theta_{h}\right )$. The internal state $h_{t}$ summarizes the history of the information seen to decide how to act and to deploy the sensor. The location network $f_{l}\left ( h_{t}, \theta_{l}\right )$ and action network $f_{a}\left ( h_{t}, \theta_{a}\right )$ uses $h_{t}$ to generate next location $l_{t}$, and the classification $action_{t}$, respectively. For simple classification experiments by Graves \cite{mnih_recurrent_2014}, $f_{h}$ was a network of rectifier
units defined as $h_{t} = f_{h}\left ( h_{t-1}, g_{t} \right ) = Rect\left ( Linear\left ( h_{t-1} \right ) + Linear\left ( g_{t} \right ) \right )$, and on a dynamic environment was used LSTM units. The location networks generate as output the average's location policy, given by $f_{l}\left ( h_{t}, \theta_{l}\right ) = Linear(h_{t})$. The location $l_{t}$ is chosen stochastically from a distribution parameterized by the location network $f_{l}(h_{t},\theta_{l})$, ie, $l_{t} \sim p(\cdot | f_{l}(h_{t},\theta_{l}))$, where $p$ is two-component Gaussian with a fixed variance. 

For classification decisions, the action network $f_{a}\left ( h_{t}, \theta_{a}\right ) = \frac{e^{Linear\left ( h_{t} \right )}}{Z}$, conditions a distribution to generate the output $action_{t} \sim p\left ( \cdot | f_{a}\left ( h_{t},\theta_{a} \right ) \right )$, where $p$ is a softmax. After executing an action, the agent receives a new visual observation of the environment $x_{t+1}$ and reward signal $r_{t+1}$. The goal is to maximize the sum $R = \sum_{t=1}^{T}r_{t}$ of the reward signal. For image classification, for $r_{T} = 1$ if the object is classified correctly after 
$T$ steps and $0$ otherwise. This setup is instance of a Partially Observable Markov Decision Process (POMDP). The true state of the environment is unobserved, and the model needs to learning a stochastic policy $\pi \left ( \left ( l_{t},a_{t} \right ) | s_{1:t};\theta \right )$ with parameters $\theta = \left \{ \theta_{g}, \theta_{a}, \theta_{h} \right \}$, that, at each time step $t$, maps the the environment's history $s_{1:t} = x_{1}, l_{1}, a_{1}, . . ., x_{t−1}, l_{t−1}, a_{t−1}, x_{t}$ to a distribution over actions. 
\subsection{End-To-End Memory Networks (EMNet): A memory-based end-to-end attention system}
\label{sec:end_to_end_memory}

\begin{figure}[h]
  \centering
  \includegraphics[width=\linewidth]{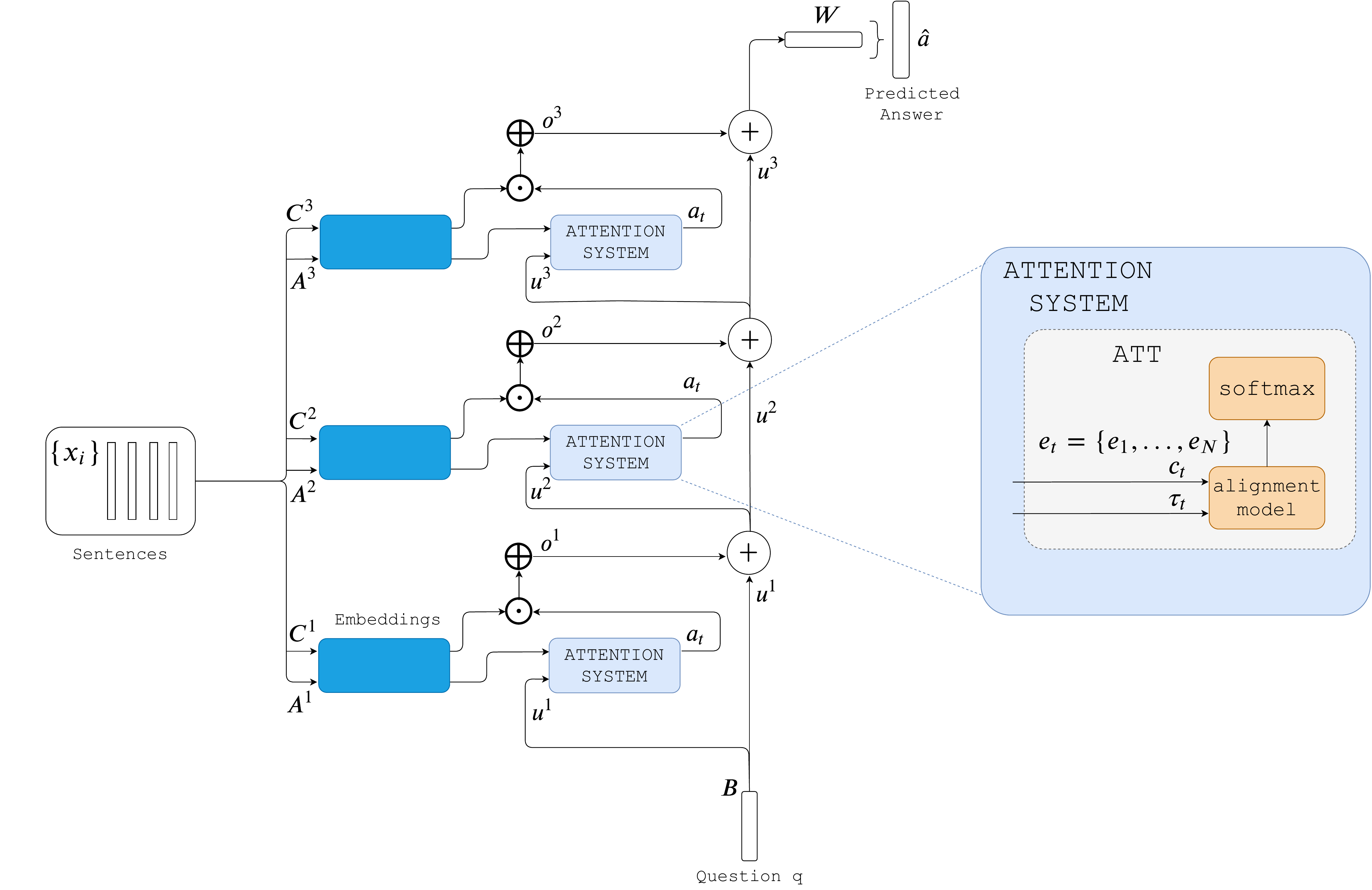}
  \caption[memory]{End-to-End Memory Networks \cite{sukhbaatar2015end} architecture illustration. The network is a stack of attentional systems interconnected with each other and with external memory. All subsystems are a top-down and cognitive selection. This architecture shows the attention distributed throughout the network, in which the selection of inferior stimuli guides selection at higher levels through the interconnection between the modules. Such a structure is closer to the biological mechanisms since there is not only an attentional center in the brain.}
  \label{fig:end_to_end_arquitetura}
\end{figure}

End-To-End Memory Networks \cite{sukhbaatar2015end} uses attention for question answering, and language modeling. The architecture is a form of Memory Networks \cite{weston_2014_memory} but unlike, it is trained end-to-end. EMNet consists of a memory and a stack of identical attentional systems, as shown in Figure \ref{fig:end_to_end_arquitetura}. 
Each layer $i$ takes as input set $\left \{ x_{1}, ..., x_{N} \right \}$ to be stored in memory. The input set is converted in memory vectors $\left \{ m_{1}, ..., m_{N} \right \}$ and $\left \{ h_{1}, ..., h_{N} \right \}$ using the embedding matrix $A^{i} \in \mathbb{R}^{d \times V}$ to generate each $m_{i} \in \mathbb{R}^{d}$, and the matrix $C^{i} \in \mathbb{R}^{d \times V}$ to generate each $h_{i} \in \mathbb{R}^{d}$. In the first layer, the question $q$ is also embedded by $B^{1}$ to obtain an internal state $u^{1}$. From the second layer, the internal state $u^{i+1} = u^{i} + o^{i} \in \mathbb{R}^{d}$ is the sum of the $i$ layer output and the internal state $u^{i}$. Finally, last layer generates the prediction $\hat{a} = \frac{e^{Wu^{i+1}}}{\sum_{j = 1}^{N}e^{Wu^{i+1}}}$, where $\hat{a} \in \mathbb{R}^{V}$ is the predicted label, and $W \in \mathbb{R}^{V \times d}$ is the matrix of weights.    

The attention system is the architecture's core. In each layer $i$, it takes as focus target $\tau_{t} = \left \{ \tau_{t}^{1} \right \} = \left \{ \tau_{t,1}^{1}, ..., \tau_{t,N}^{1} \right \} = \left \{ h_{1}, ..., h_{N} \right \}$ memories' embeddings, and contextual input $c_{t} = \left \{ c_{t}^{1} \right \} = \left \{ c_{t,1}^{1}, .., c_{t,N+1}^{1} \right \} = \left \{ u^{i}, m_{1}, ..., m_{N}  \right \}$. Through an alignment function $e_{i,j} = (u^{i})^{T}m_{j}$ the attentional system computes the match between $u^{i}$ and each memory $m_{i}$, generating as output a mask of importance $a_{t} = \left \{ a_{t}^{1} \right \} = \left \{ a_{t,1}^{1}, ..., a_{t,N}^{1} \right \} = \left \{ a_{1}, ..., a_{N} \right \}$ for each $h_{i}$, where $a_{j} = \frac{e^{e_{i,j}}}{\sum_{j = 1}^{N}e^{e_{i,j}}} \in \mathbb{R}$. The output $o^{i}$ is a sum over the transformed inputs $h_{i}$, weighted by the attention mask. Intuitively, attention looks for the memory elements most related to question $q$ using a simple alignment function dispensing traditional RNNs. This system can also be seen as version of attention in RNNSearch \cite{bahdanau_neural_2014} with mutiple computational steps per output symbol, and with similar selection characteristics -- \textbf{soft}, \textbf{divided}, \textbf{top-down stateful}, \textbf{cognitive}, and \textbf{location-based}. Note that the alignment function is differentiable. Therefore, during training, all architecture elements are jointly learned by minimizing a standard cross-entropy loss between $\hat{a}$ and the true label $a$ using stochastic gradient descent.

\subsection{Show, Attend and Tell: A multimodal approach}
\label{sec:show_attend_tell}

\begin{figure}[h]
  \centering
  \includegraphics[width=\linewidth]{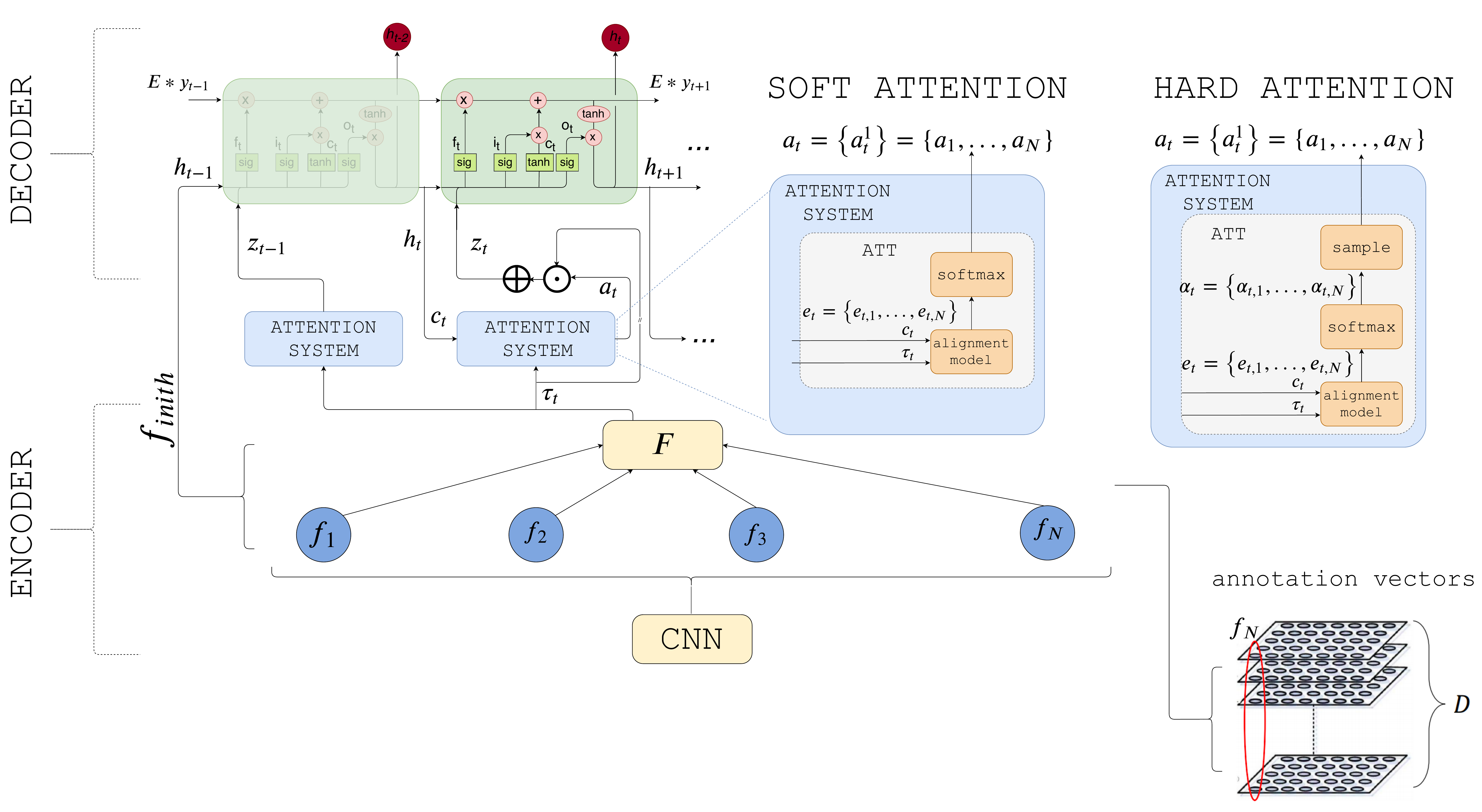}
  \caption[show]{Show, Attend and Tell \cite{xu_show_2015} architecture illustration. Attention unifies different sensory experiences to decide a task. A single attentional subsystem, which can be hard or soft, aligns high-level image and text representations to determine the next word in the decoder.}
  \label{fig:show_attend_and_tell_model}
\end{figure}

Show, Attend and Tell \cite{xu_show_2015} uses attention for image caption. The architecture consists of an encoder, decoder, and an attention system (Figure \ref{fig:show_attend_and_tell_model}). The encoder is a CNN to extract features from image $I$. And at time step $t$, the decoder uses LSTM units for generating one word $y$ for a caption. Similar to RNNSearch \cite{bahdanau_neural_2014}, the decoder calculates from a context vector $z_{t}$ a probability distribution for all possible caption symbols, as follows

\begin{equation}
    (f_{1}, ..., f_{N}) = Encoder(I) 
    \label{annotations}
\end{equation}

\begin{equation}
    p(y_{t}|y_{1}, ..., y_{C}, I) = Decoder(z_{t})
    \label{decoder_rnn}
\end{equation}

where $C$ is caption size.

The attention system has a subsystem (Figure \ref{fig:show_attend_and_tell_model}) that generates the context vector $z_{t}$ for the decoder. It receives a contextual input $c_{t} = \left \{ c_{t}^{1} \right \} = \left \{ c_{t,1}^{1} \right \} = \left \{ h_{t-1} \right \}$ with language content, a focus target $\tau_{t} = \left \{ \tau_{t}^{2} \right \} = \left \{ \tau_{t,1}^{2}, ..., \tau_{t,N}^{2} \right \} = \left \{ f_{1}, ..., f_{N} \right \}$ with visual content, and produces attention weigths $a_{t} = \left \{ a_{t}^{2} \right \} = \left \{ a_{t,1}^{2}, ..., a_{t,N}^{2} \right \} = \left \{ a_{1}, ..., a_{N} \right \}$ as output, where $h_{t-1} \in \mathbb{R}^{1 \times d}$ is a previous decoder state, $f_{j} \in \mathbb{R}^{1 \times d_{I}}$ is encoder annotation vector, $a_{t}^{2} \in \mathbb{R}^{1 \times N}$ are attention weights for all annotation vectors. This structure receives as input different sensory modalities (i.e, visual in focus target and language in context), inspired by perceptual theories \cite{o2012perception}. Nos seres vivos, a experiência perceptual não é desarticulada e fragmentada, mas está intimimente ligada à uma unidade objectual comum. Por exemplo, para agarrar uma bola, uma pessoa precisa vê-la se aproximando, ou para decidir o gosto de uma comida, elementos do tato e do cheiro ponderam a decisão. Similarmente, esta arquitetura implementa atenção entre duas fontes sensoriais diferentes para decidir uma tarefa. The attention system is equal to the RNNSearch \cite{bahdanau_neural_2014}, and has same selection features -- \textbf{soft}, \textbf{divided}, \textbf{location-based}, \textbf{top-down stateful}, \textbf{oriented}, and \textbf{cognitive} -- if implements soft attention. In contrast, if the mechanism is hard attention has an additional sampling block $a_{t} \sim Multinoulli_{N}(\alpha_{t,j})$ parameterized by the scores $\alpha_{t,j} = \frac{e^{e_{t,j}}}{\sum_{j = 1}^{N}e^{e_{t,j}}}$, where $e_{t,j} = a(h_{t-1}, f_{j})$. The sampling mechanism makes the system stochastic with \textbf{hard continuity}.

\subsection{Deep Recurrent Attentive Writer (DRAW)}
\label{sec:draw_network}

\begin{figure}[htb]
  \centering
  \includegraphics[width=\linewidth]{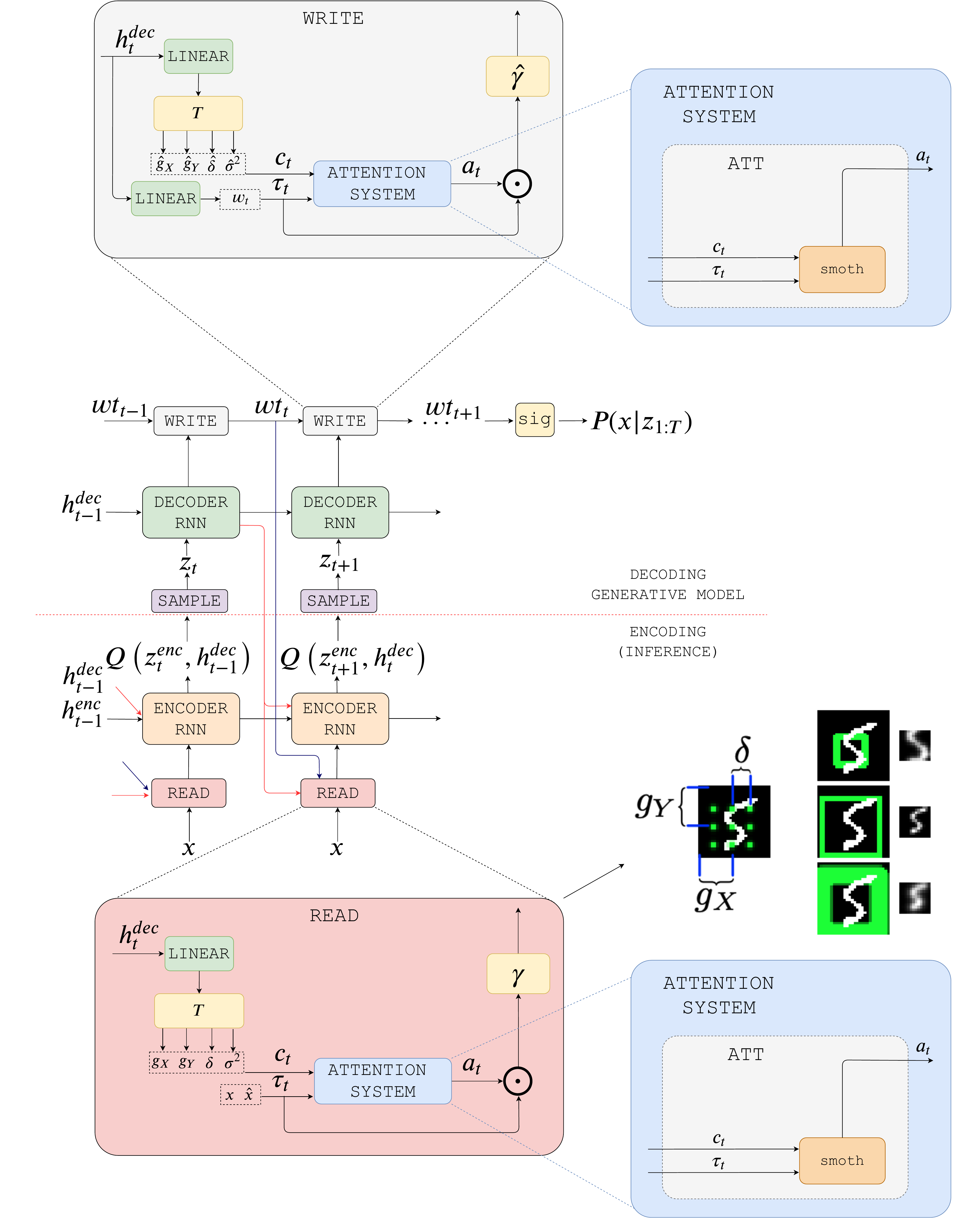}
  \caption[draw]{DRAW \cite{draw} architecture illustration. The architecture has an encoder, a decoder, and attention-controlled read/write heads. \textit{Read} generates a superimposed filter grid on an image and extract an patch $N \times N$. Similarly, \textit{write} head defines which image patch will be drawn. Both systems are divided, selective, oriented, location-based, top-down stateful, soft and hard.}
  \label{fig:draw_model}
\end{figure}

DRAW \cite{draw} uses attention for image generation. The architecture is similar variational autoencoders (VAEs) \cite{doersch2016tutorial} with some differences (Figure \ref{fig:draw_model}). Firstly, the encoder/decoder are recurrent neural networks, and the encoder receives the previous outputs from the decoder. Secondly, the decoder outputs are added successively to the distribution generating the data, instead of generating that distribution in a single step. And thirdly, the attention system dynamically updates the network, restricting the input region observed by the encoder and the output region modified by the decoder. The attention make decisions about \textit{which} regions are input for the network, \textit{which} regions are modified in the generated image, and also \textit{what} needs to be modified.

At each time-step $t$, the encoder receives as input the read vector $r_{t} = read(x_{t},\hat{x_{t}}, h_{t-1}^{dec})$, and the previous decoder's state $h_{t-1}^{dec}$ for generate current state $ h_{t}^{enc} = RNN^{enc}(h_{t-1}^{enc}, [r_{t},h_{t-1}^{dec}])$, where $x_{t}$ is input image, $\hat{x_ {t}} = x - \sigma(wt_{t-1})$ is error image, and $\sigma$ is the sigmoid function. The $h_{t}^{enc}$ parameterize the $Q$ distribution over the latent space vector $z_{t} \sim Q(Z_{t}|h_{t}^{enc})$, which is input to the decoder. The $Q \left (Z_{t}|h_{t}^{enc}\right)$ distribution is a diagonal Gaussian $N \left (Z_{t}|\mu_{t},\sigma_{t} \right ) $ in latent space, where $\mu_{t} = W(h_{t}^{enc})$, $\sigma_{t} = e^{W(h_{t}^{enc})}$, \textit{W} are weight matrices of a linear transformation layer. The decoder output $h_{t}^{dec} = RNN^{dec}(h_{t-1}^{dec},z_{t})$ is added by the \textit{write} function for reconstruct the image through the cumulative \textit{canvas} matrix $wt_{t} = wt_{t-1} + write(h_{t}^{dec})$. After \textit{T} iterations the \textit{canvas} matrix $wt_{T}$  parameterize $D(X|wt_{T})$ for generate image $\tilde{x}$:

\begin{equation}
    \tilde{z_{t}} \sim P(Z_{t})
    \label{eq_10}
\end{equation}

\begin{equation}
    \tilde{h}_{t}^{dec} = RNN^{dec}(\tilde{h}_{t-1}^{dec},\tilde{z}_{t})
    \label{eq_11}
\end{equation}

\begin{equation}
    \tilde{wt}_{t} = \tilde{wt}_{t-1} + write(\tilde{h}_{t}^{dec})
    \label{eq_12}
\end{equation}

\begin{equation}
    \tilde{x} \sim D(X|\tilde{wt}_{T})
    \label{eq_13}
\end{equation}

\noindent where \textit{P} is a prior, $\tilde{z_{t}}$ is a sample of latent space. \textit{D} is a probability distribution, if the input is binary it is a Bernoulli distribution with a mean given by $\sigma(wt_{T})$.

The \textit{read} and \textit{write} functions are controlled by the attention systems for decide which image path will be processed, and which image region will be modified in the output (Figure \ref{fig:draw_model}). The attention system has only one subsystem and a single sensory modality (i.e, visual). At each time step \textit{t}, the subsystem takes as input a contextual $c_{t} = \left \{ c_{t}^{1} \right \} = \left \{ c_{t,1}^{1}, c_{t,2}^{1}, c_{t,3}^{1}, c_{t,4}^{1} \right \} = \left \{ g_{X}, g_{Y}, \delta, \sigma^{2} \right \}$ with grid properties, a focus target
$\tau_{t} = \left \{ \tau_{t}^{1} \right \} = \left \{ \tau_{t,1}^{1}, ..., \tau_{t,2N}^{1} \right \} = \left \{ x, \hat{x} \right \}$, and produces attention weights $a_{t} = \left \{ a_{t}^{1} \right \} = \left \{ a_{t,1}^{1}, ..., a_{t,2N}^{1} \right \} = \left \{ F_{x}, F_{y} \right \}$. The pair $(g_{X}, g_{Y})$ is grid center coordinates, $\delta$ is width of the step, $\sigma^{2}$ is isotropic variance of the Gaussian filters, $x$ $\in \mathbb{R}^{B \times A}$ is a input image, $\hat{x} \in \mathbb{R}^{A \times B}$ is the error image, $a_{t}^{1} \in \mathbb{R}^{1 \times 2N}$ is the weight of attention to $\tau_{t}$. The parameters $g_{X}$, $g_{Y}$, $\delta$, $\sigma^{2}$, $\gamma$ are determined dynamically using a linear transformation of the $h^{dec}$, as 

\begin{equation}
    (\tilde{g_{X}},\tilde{g_{Y}},log\sigma^{2},log\tilde{\delta},log\gamma) = W(h^{dec})
    \label{saida_decoder}
\end{equation}

\begin{equation}
    g_{X} = \frac{A+1}{2}(\tilde{g}_{X} + 1)
    \label{g_x}
\end{equation}

\begin{equation}
    g_{y} = \frac{B+1}{2}(\tilde{g}_{Y} + 1)
    \label{g_y}
\end{equation}

\begin{equation}
    \delta = \frac{max(A,B) - 1}{N-1}\tilde{\delta}
    \label{delta_formula}
\end{equation}

\noindent where variance, step, and intensity are given on a log scale to ensure positive values. The scale of $g_{X}, g_{Y}, \delta$ is chosen to ensure that the initial path, with an initialized network, covers approximately the entire input image.

The attention system in \textit{read} is \textbf{divided}, \textbf{selective}, \textbf{oriented}, \textbf{perceptive}, \textbf{location-based}, \textbf{top-down stateful}, \textbf{soft}, and \textbf{hard}. Despite being completely differentiable, this system applies the soft mask over the entire image as a divided selection. However, unlike other approaches in the literature, targeting mask computation only generates patches from one region, such as hard and selective attention. In the \textit{write}, the features are similar, but the attention is \textbf{cognitive}. Specifically, the attention functions generate as output $a_{t}$ matrices of the horizontal and vertical Gaussian filter bank $F_{X}[i,a] = \frac{1}{Z_{X}}e^{\left ( - \frac{(a - \mu_{X}^{i})^{2}}{2\sigma^{2}} \right )}$, and $ F_{Y}[j,b] = \frac{1}{Z_{Y}}e^{\left ( - \frac{(b - \mu_{Y}^{i})^{2}}{2\sigma^{2}} \right )}$, where $F_{X} \in \mathbb{R}^{N \times A}$, $F_{Y}$ $\in \mathbb{R}^{N \times B}$, $(i, j)$ is a point of the attention mask, $\left ( a, b \right )$ is a point in the input image, $Z_{X}$ and $Z_{Y}$ are normalization constants for $\sum_{a} F_{X}[i, a] = 1$ and $\sum_{b} F_{Y}[j, b] = 1$. Finally, \textit{read} operation returns concatenation of two patches $N \times N$ of the input image and the error image. For the \textit{write} operation, a specific set of parameters $\hat{\delta}$, $\hat{F_{X}}$, $\hat{F_{Y}} $ are output from $ h_{t}^{dec}$:

\begin{equation}
    read(x,\hat{x}_{t},h_{t-1}^{dec}) = \gamma[F_{Y}xF_{X}^{T},F_{Y}\hat{x}F_{X}^{T}]
    \label{read}
\end{equation}

\begin{equation}
    w_{t} = W(h_{t}^{dec})
    \label{write_1}
\end{equation}

\begin{equation}
    write(h_{t}^{dec}) = \frac{1}{\hat{\gamma}}\hat{F_{Y}^{T}}w_{t}\hat{F_{X}}
    \label{write_2}
\end{equation}

\noindent where $w_{t}$ is $\mathbb{R}^{N \times N}$ \textit{writing} patch emited by $h_{t}^{dec}$.

\subsection{Bi-Directional Attention Flow (BiDAF)}
\label{sec:bidaf}

\begin{figure}[h]
  \centering
  \includegraphics[width=\linewidth]{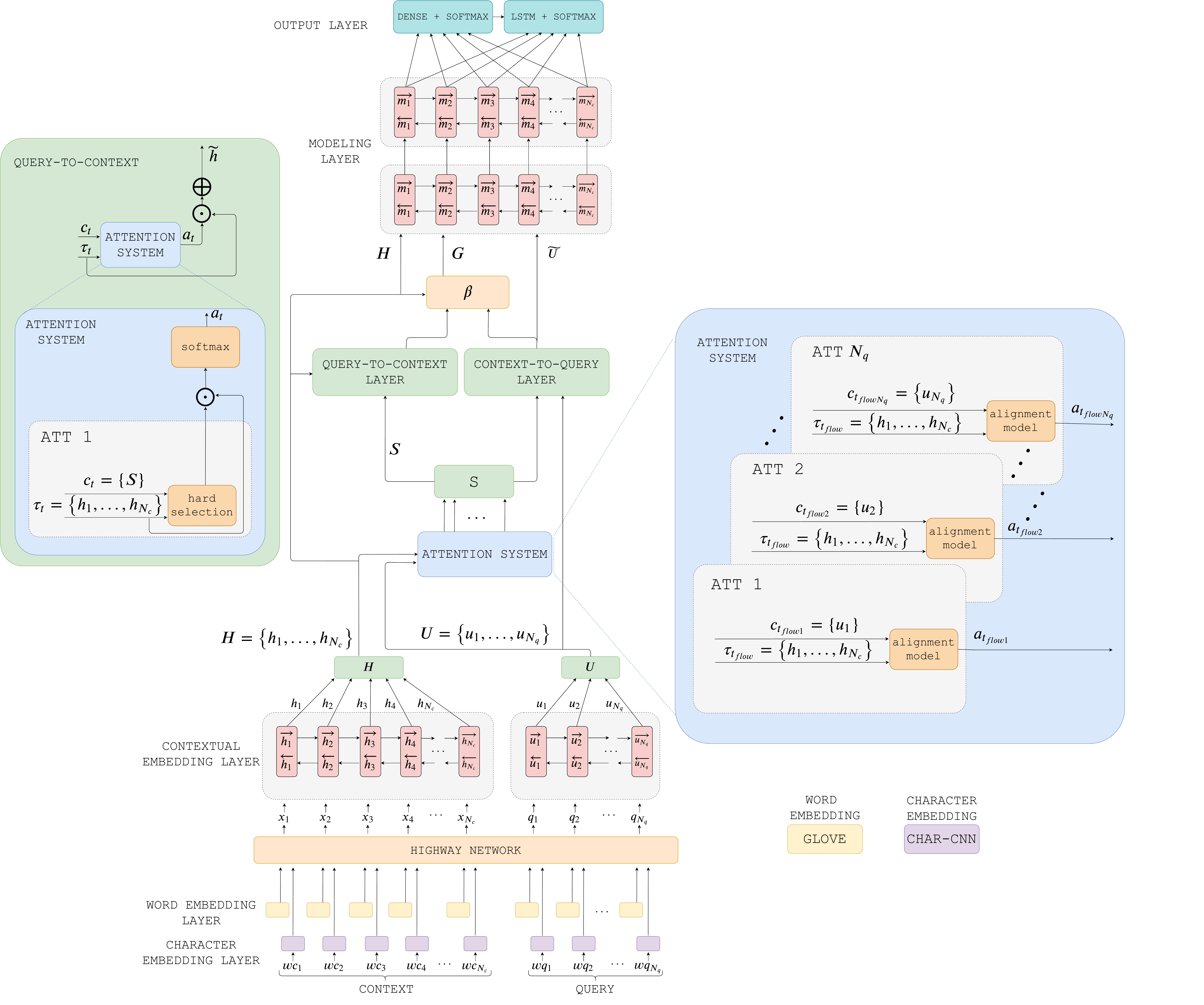}
  \caption[bidaf]{BiDAF \cite{seo_bidirectional_2016} architecture illustration. The attention system has small independent modules that communicate different queries with the target. In sequence, a hierarchically superior subsystem recalibrates previous attentional signals changing original attention focus.}
  \label{fig:bidaf_arquitetura}
\end{figure}

BiDAF \cite{seo_bidirectional_2016} uses attention for machine comprehension. The architecture consists of eight main components -- character embedding layer, word embedding layer, contextual embedding layer, attention system, context-to-query, query-to-context, modeling layer and output layer (Figure \ref{fig:bidaf_arquitetura}). The character embedding layer maps each word to a high-dimensional vector space. Let $\left \{ wc_{1}, ..., wc_{N_{c}} \right \}$ and $\left \{ wq_{1}, ..., wq_{N_{q}} \right \}$ represent the words in the input paragraph and the question, respectively. Characters are embedded into vectors using Convolutional Neural Network (CNN). The word embedding layer maps each word to a high-dimensional vector space using pre-trained Glove \cite{pennington2014glove}. The concatenation of the character and word embedding vectors is passed through two-layer of the Highway Network \cite{srivastava2015highway}, and generate context set $X = \left \{ x_{1}, ..., x_{N_{c}} \right \}$, $x_{i} \in \mathbb{R}^{d}$, and query set $Q = \left \{ q_{1}, ..., q_{N_{c}} \right \}$, $q_{i} \in \mathbb{R}^{d}$.

The contextual embedding layer takes as input a set $X$ and $Q$ and uses an bidirectional LSTM \cite{bi_lstm} to model temporal iterations between words, generating as output a set of context vectors $H = \left \{ h_{1}, ..., h_{N_{c}} \right \}$, $h_{i} \in \mathbb{R}^{2d}$, and query vectors $U = \left \{ u_{1}, ..., u_{N_{q}} \right \}$, $u_{i} \in \mathbb{R}^{2d}$. Each vector of $H$ and $U$ is 2d dimensional due to the concatenation of the forward and backward LSTMs outputs. Note that the first three layers are feature extractors for context and queries in different granularity levels, similarly CNNs. The next layer is the attentional system, whose function is to link and disseminate information from the context and query. Unlike the classic mechanisms, the attentional system does not summarize the query and the context in vectors of unique features. Instead, the attention vector and the embedding's previous layers can flow to the subsequent modeling layer to reduce information loss by the early summary.

The attentional system has several subsystems in parallel, which receive as input the same focus target and different contextual information, that is, $\tau_{t} = \left \{ \tau_{t}^{1} \right \} = \left \{ \tau_{t,1}^{1}, ..., \tau_{t,N_{c}}^{1} \right \} = \left \{ h_{1}, ..., h_{N_{c}} \right \}$, $c_{t} = \left \{ c_{t}^{1} \right \} = \left \{ c_{t,1}^{1} \right \} = \left \{ u_{j} \right \}$. Each subsystem has an alignment function that relates all context vectors to a given query $u_{j}$, and outputs masks that together make up an attention matrix $S \in \mathbb{R}^{N_{c} \times N_{q}}$, ie, $a_{t} = \left \{ a_{t}^{1} \right \} = \left \{ \tau_{t,1}^{1}, ..., \tau_{t,N_{c}}^{1} \right \} = \left \{ S_{1,1}, ..., S_{N_{c},j} \right \}$. 
Each subsystem presents the same selection features -- \textbf{soft}, \textbf{divided}, \textbf{location-based}, \textbf{top-down stateful}, and \textbf{cognitive}. Specifically,  

\begin{equation}
    S_{i,j} = a(h_{i}, u_{j})
    \label{S}
\end{equation}

\begin{equation}
    a(h_{i}, u_{j}) = w_{(S)}^{T}[h;u;h \odot u]
    \label{alp}
\end{equation}

\noindent where $S_{i,j} \in \mathbb{R}$ indicates the similarity between the $i-th$ word of the context vector and the $j-th$ query word, $h_{i}$ is the $i-th$ context vector, and $u_{j}$ is the $j-th$ query vector, and $a$ is alignment function that encodes the similarity between two input vectors, $w_{(S)} \in \mathbb{R}^{6D}$ is a trainable weight vector, $\odot$ is point-to-point multiplication, [;] is vector concatenation across row.

The $S$ matrix is input to the context-to-query layer (C2Q), which normalizes the weights and applies them to each query vector $u_{j}$ to produce the set
$\widetilde{U} = \left \{ \widetilde{u_{1}}, ..., \widetilde{u}_{N_{c} } \right \}$ with the most relevant query words for each context word. $S$ is also input to the query-to-context layer (Q2C), which has an attention subsystem \textbf{hard}, \textbf{selective}, \textbf{location-based}, \textbf{top-down stateful}, and \textbf{cognitive}. The attention subsystem uses a row's $S$ matrix as contextual information and the $H$ set as focus target $\tau_{t}$. It uses a \textbf{hard selection} to change the focus of attention produced in previous layers using the $max_{col}$ function. The output's attention mask searches for which words in the context are essential to answer the query generating a dynamic context vector $\widetilde{h}$. Specifically,

\begin{equation}
    \alpha_{t} = softmax(S_{t,1:N_{q}})
    \label{S}
\end{equation}

\begin{equation}
    \widetilde{u_{t}} = \sum_{j=1}^{N_{q}} \alpha_{t,j}u_{j}
    \label{softmax_bidaf}
\end{equation}

\begin{equation}
    b = softmax(max_{col}(S))
    \label{softmax_bidaf_2}
\end{equation}

\begin{equation}
    \widetilde{h} = \sum_{j=1}^{N_{c}} b_{j}h_{j}
    \label{softmax_bidaf_2}
\end{equation}

\noindent where $\alpha_{t} \in \mathbb{R}^{N_{q}}$ and $\sum \alpha_{t} = 1$ $\forall t$, $\widetilde{u_{t}} \in \mathbb{R}^{2d}$ is an array containing the query vectors served by the entire context, $b \in \mathbb {T}$, and the $ max_{col}$ is performed across the column, and $\widetilde{h} \in \mathbb{R}^{2d}$.

Finally, the contextual embeddings $h_{i}$, $\widetilde{u}_{i}$, and $\widetilde{h}_{i}$ are combined to generate $G = \left \{ g_{1}, ..., g_{N_{c}} \right \}$, where each vector can be considered as the query-aware representation of each context word. Next, $G$ is input for the modeling layer, which captures the interaction among the context words conditioned on the query. This layer uses two bidirectional LSTMs \cite{bi_lstm}, with output dimension \textit{d} for each direction, getting the matrix $\beta$ which is passed to the output layer to predict the response. The output layer is application-specific, predicting the initial and final indices of the paragraph sentence. For this, the probability distributions of the initial and final index are obtained throughout the entire paragraph, as follows

\begin{equation}
    g_{i} = \beta(h_{i}, \widetilde{u_{i}}, \widetilde{h_{i}})
    \label{G_t}
\end{equation}

\begin{equation}
    \beta(h_{i}, \widetilde{u_{i}}, \widetilde{h_{i}}) = [h_{i}; \widetilde{u_{i}}; h_{i} \odot \widetilde{u_{i}}; h_{i} \odot \widetilde{h_{i}}]
    \label{beta_t}
\end{equation}

\begin{equation}
    p^{1} = Softmax(w_{(p^{1})}^{T}[G;M])
    \label{p_1}
\end{equation}

\begin{equation}
    p^{2} = Softmax(w_{(p^{2})}^{T}[G;M_{2}])
    \label{p_2}
\end{equation}

\noindent where $g_{i} \in \mathbb{R}^{d_{G}}$ corresponding to the \textit{i-th} context word, $\beta \in \mathbb{R}^{d_{G}} (d_{G} = 8d)$ can be an trainable neural network but in this example is a function that merges three vectors, and $d_{G}$ is the output dimension of the $ \beta $ function, $w_{(p^{1})} \in \mathbb{R}^{10d} $ are vectors of trainable weights, $M$ is the output of the first modeling layer, $ M_{2} \in \mathbb{R}^{2d \times N_{c}}$ is the output of \textit{M} after going through another layer of bidirectional LSTM.

\subsection{Neural Transformer}
\label{sec:neural_transformer}

\begin{figure}[h]
  \centering
  \includegraphics[width=\linewidth]{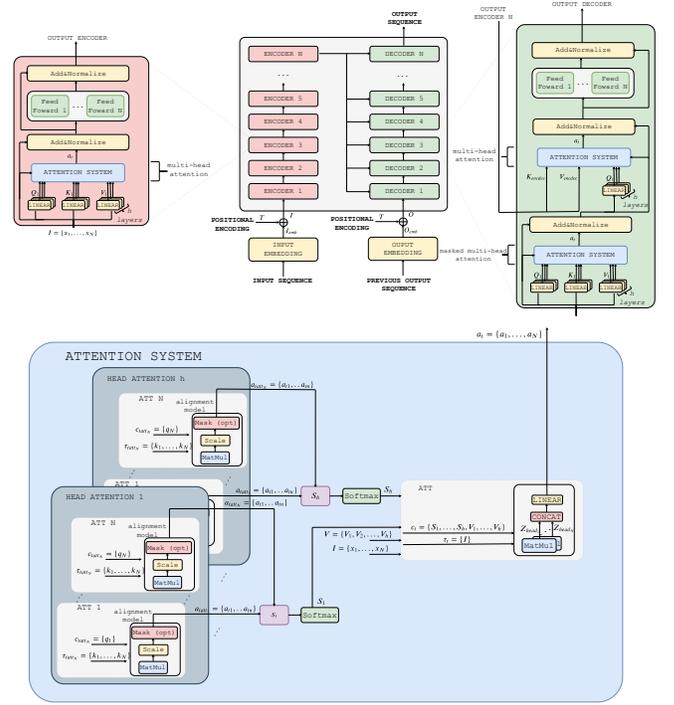}
  \caption[ntm]{The Neural Transformer \cite{vaswani_attention_2017} architecture illustration. The attention system has several parallel heads. Each head has $N$ attention subsystems that perform the same task but have different contextual inputs. The encoder is purely bottom-up stateful attention. Each query communicates each stimulus in focus target with the other. The decoder is hybrid, in which the first system is bottom-up and the second is top-down.}
  \label{fig:transformer_arquitetura}
\end{figure}

Neural Transformer \cite{vaswani_attention_2017} uses attention for machine translation. The architecture consists of an arbitrary amount of stacked encoders/decoders, as shown in Figure \ref{fig:transformer_arquitetura}. Each encoder has linear layers, an attention system, feed-forward neural networks, and normalization steps. The attention system is architecture's core. It has several parallel heads. Each head has $N$ attention subsystems that perform the same task, but with different contextual inputs. In the first layer, the encoder receives as input an embedding matrix $I = I_{emb} + T$ $\in \mathbb{R}^{N \times d_{emb}}$ for $N$ words, where $I_{emb} = \left \{ e_{1}, e_{2}, e_{3}, ..., e_{N} \right \}, e_{i} \in \mathbb{R}^{1 \times d_{emb}}$ is an embedding matrix, and $T = \left \{ PE_{0}, PE_{1}, PE_{2}, PE_{3}, ..., PE_{N} \right \}$ is composed by positional encodings $PE_{i} = \left \{ pe_{0}, pe_{1}, p_{2}, ..., p_{d_{emb}} \right \}$:

\begin{equation}
    pe_{i} \in \left\{\begin{matrix}
sin(\frac{pos}{10000^{\frac{2i}{d_{emb}}}}) &  i = 0, 2, ..., d_{emb}\\ 
cos(\frac{pos}{10000^{\frac{2i}{d_{emb}}}}) & i = 1, 3, ..., d_{emb-1}
\end{matrix}\right.
    \label{pe_vector}
\end{equation}

\noindent where $d_{emb} = 512$ is the word embedding dimension. $T$ is the matrix of positions encoders. $PE_{i} \in \mathbb{R}^{1 \times d_{emb}}$ is a position encoder vector for a word, \textit{pos} is the position of the word in the sequence, \textit{i} $ \in [0, 255] $, and refers to the dimension value.

The input $I$ goes through linear layers and generates, for each word, a \textbf{query vector} ($q_{i}$), a \textbf{key vector} ($k_{i}$), and a \textbf{value vector} ($v_{i}$), as follows

\begin{equation}
    Q_{i} = X \times W^{Q}_{i}, Q_{i} \in \mathbb{R}^{N \times d_{k}}
    \label{matriz_q}
\end{equation}

\begin{equation}
    K_{i} = X \times W^{K}_{i}, K_{i} \in \mathbb{R}^{N \times d_{k}}
    \label{matriz_k}
\end{equation}

\begin{equation}
    V_{i} = X \times W^{V}_{i}, V_{i} \in \mathbb{R}^{N \times d_{k}}
    \label{matriz_k}
\end{equation}

\noindent where \textit{i} is the head index, $W^{Q}_{i}$, $W^{K}_{i}$, and $W^{V}_{i}$ $\in \mathbb{R}^{d_{emb} \times d_{k}}$ are trainable weights. The value $d_{k}$ is $d_{emb}/h = 64$, and $h = 8$ is the number of parallel heads. $Q_{i} = \left \{ q_{1}, q_{2}, ... , q_{N} \right \}, q_{i} \in \mathbb{R}^{1 \times d_{k}}$ is the queries matrix, $K_{i} = \left \{ k_{1}, k_{2}, ... , k_{N} \right \}, k_{i} \in \mathbb{R}^{1 \times d_{k}}$ is the keys matrix, and $V_{i} = \left \{ v_{1}, v_{2}, ... , v_{N} \right \}, v_{i} \in \mathbb{R}^{1 \times d_{k}}$ is the values matrix.

The attention system receives as input all $Q$, $K$, and $V$ arrays in several parallel attention heads. The multi-head structure explores multiple subspaces, getting different projections of the data. Having multiple heads on the Transformer is similar to having multiple filters on CNNs. Each head has $N$ attention subsystems which receive as input the same focus target $\tau_{t} = \left \{ \tau_{t}^{1} \right \} = \left \{ \tau_{t,1}^{1}, ..., \tau_{t,N}^{1} \right \} = \left \{ k_{1}, ..., k_{N} \right \}$, different contextual inputs $c_{t_{att_{i}}} = \left \{ c_{t}^{1} \right \} = \left \{ c_{t,1}^{1} \right \} = \left \{ q_{i} \right \}$, and outputs an attention mask that relates all keys to specific query $q_{i}$. 
Mathematically, the operations performed by a head (Figure \ref{fig:transformer_arquitetura}) are represented by matrix multiplication between all queries and keys, as follows

\begin{equation}
    S_{i} = softmax(\frac{Q_{i}*K_{i}^{T}}{\sqrt{d_{k}}})
    \label{self_attention_matrix}
\end{equation}

\noindent where $S_{i} \in \mathbb{R}^{N \times N}$ is self-attention matrix, \textit{i} is the head index, Q $\in \mathbb{R}^{N \times d_{k}}$, $K^{T}$ $\in \mathbb{R}^{d_{k} \times N}$. For gradient stability, $d_{k}$ is the normalization factor, and an softmax function turns the attention scores into probabilities.

The similarity between $q_{i}$ and $k_{i}$ is a score of importance that the elements of the sequence have among themselves. This structure characterizes the most important \textbf{bottom-up stateful} system in the field, in which contexts come simultaneously from the discrepancies between the different stimuli of the focus target itself. Intuitively, queries act as small units that allow parallel communication between all the target's stimuli. In sequence, other \textbf{bottom-up stateful} attention receives as context input $c_{t} = \left \{ c_{t}^{1} \right \} = \left \{ S_{1}, ..., S_{h}, V_{1}, ..., V_{h} \right \}$ all self-attention matrices $S_{i}$ and values $V_{i}$, and as focus target $\tau_{t} = \left \{ \tau_{t}^{1} \right \} = \left \{ \tau_{t,1}^{1}, ..., \tau_{t,N}^{1} \right \} = \left \{ x_{1}, ..., x_{N} \right \}$ the original input $I$ composed by $x_{i} \in \mathbb{R}^{d_{emb}}$ vectors. This system create final attention mask $a_{t}$ for original encoder input $I$:

\begin{equation}
    Z_{i} = S_{i}*V_{i}
    \label{self_attention_1}
\end{equation}

\begin{equation}
    a_{t} = Concat(Z_{1},Z_{2},...,Z_{h})*W^{O}
    \label{matriz_z}
\end{equation}

\noindent where $a_{t} = \left \{ a_{t,1}, ..., a_{t,N} \right \}$ $\in \mathbb{R}^{N \times d_{emb}}$ is a linear combination from each \textit {head}, and $W^{O}$ $\in \mathbb{R}^{h*d_{k} \times d_{emb}}$ are weights learned during training.

The mask $a_{t}$ modulates focus target $I$ through simple sum and layer-normalization \cite{ba2016layer} step, resulting updated $I = \left \{ x_{1}, ..., x_{N} \right \}$. After, each updated $x_{i}$ goes through feedfoward neural network composed by linear transformations and ReLU activation function. Finally, residual input $I$, and feedfoward results are input for last layer-normalization step:

\begin{equation}
    I = LayerNorm(I + a_{t})
    \label{layer_1}
\end{equation}

\begin{equation}
    f_{i} = FFN(x_{i}) = max(0,x_{i}W_{1}+b_{1})W_{2}+b_{2}
    \label{feed_foward_rede}
\end{equation}

\begin{equation}
    I = LayerNorm(I + F)
    \label{layer_2}
\end{equation}

\noindent where the hidden layer has a dimensionality $d_{ff} = 2048$, and $F$ $\in \mathbb{R}^{N \times d_{emb}}$ is feedfoward matrix, and updated $I$ $\in \mathbb{R}^{N \times d_{emb}}$ is output encoder.

The last encoder output are transformed into the contextual inputs $K_{encdec}$, $V_{encdec}$ for new attention systems in decoder levels. This data can also be understood as short-term memories that help the decoder focus on the input sequences' appropriate information. All decoders are similar to a single encoder structure, but with some important differences: 1) Step-by-step processing; and 2) Two attention systems (Figure \ref{fig:transformer_arquitetura}). The first has \textbf{masked multi-head} structure to mask future translate words with $-\infty$ values. The second receives as focus target the previously translated word embeddings and as contextual input $K_{encdec}$, $V_{encdec}$ and queries from the layer below it. This system operates as \textbf{top-down stateful} attention, given keys/values and queries/target from different sources. This structure acts as \textbf{oriented attention}, given change the focus target e attention masks at each time step $t$. After all attention systems, a linear layer followed by the softmax function transform the last decoder output stack into a probability vector to predict the correct word.

Neural Transformer represents the primary end-to-end attention approach. Communication between bottom-up and top-down structures makes it the \textbf{first hybrid attention model}. The simultaneously location-based and feature-based attention is also innovative, given the most models are only location-based. In the first stage of attention, parallel heads are \textbf{location-based}, while the second stage, composed of a single attention subsystem, is \textbf{feature-based}. Besides, the sequence of multiple stacked attention modules eliminates the need for recurrence in tasks where RNNs are typically used.

\subsection{Graph Attention Networks (GATs)}
\label{sec:gats}

\begin{figure}[h]
  \centering
  \includegraphics[width=\linewidth]{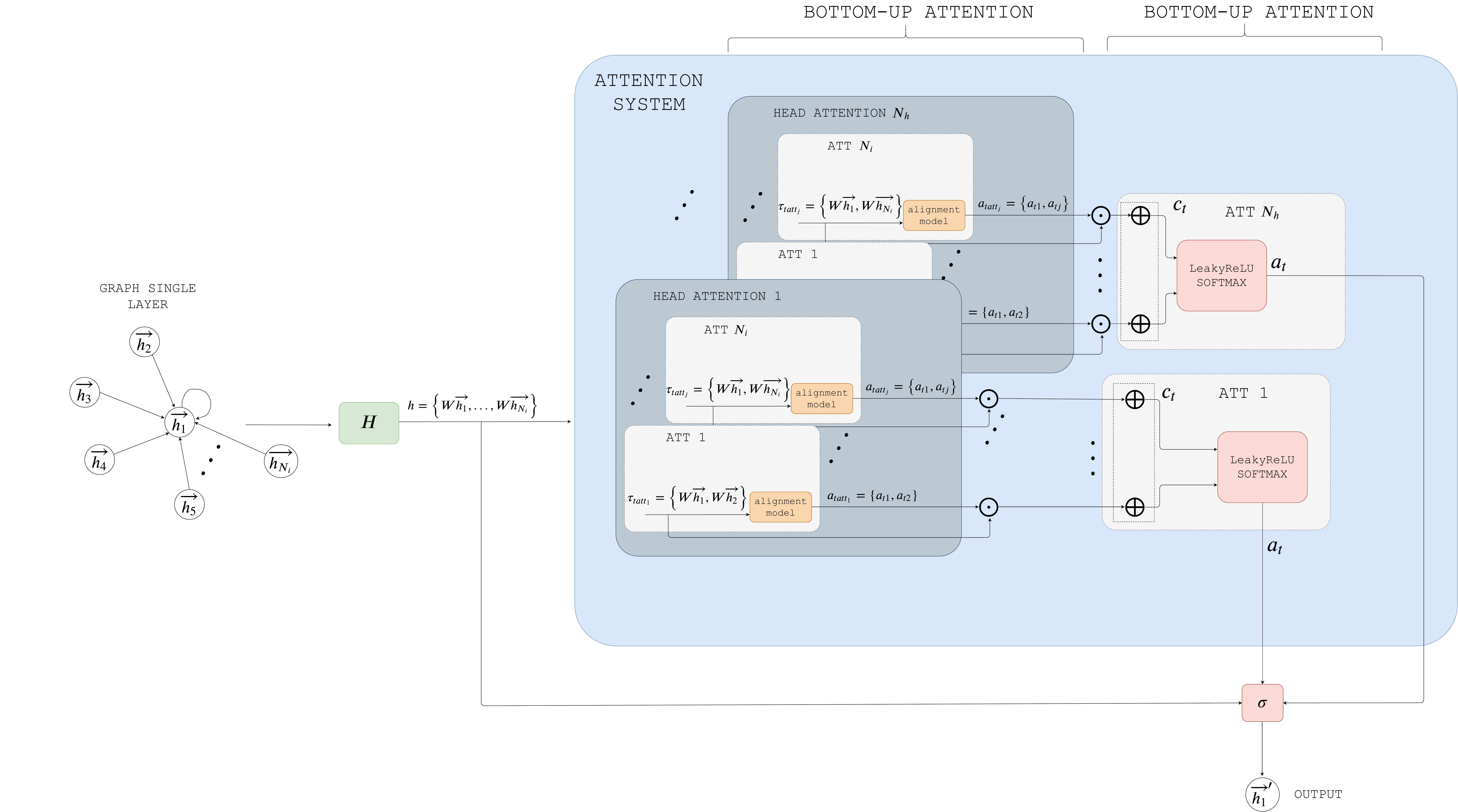}
  \caption[graph]{Graph Attentional Layers (GAL) \cite{velickovic_graph_2018} architecture illustration. The bottom-up and feature-based attention captures the discrepancies between the focus target features and generates embedding vectors for each node as output.}
  \label{fig:gats_arquitetura}
\end{figure}

GATs \cite{velickovic_graph_2018} uses attention to operate on graph-structured data. The architecture consists of stacking several graph attentional layers (GAL) (Figure \ref{fig:gats_arquitetura}). The GAL input is the nodes' features $h = \left \{ \overrightarrow{h_{1}}, \overrightarrow{h_{2}}, \overrightarrow{h_{3}}, ... ,\overrightarrow{h_{N_{i}}} \right \}$, where $ \overrightarrow {h_ {i}} \in \mathbb{R} ^ {1 \times F}$, $N$ is the number of nodes, and $F$ is the number of features in each node. The outputs are a new set of features $h^{'} = \left \{ \overrightarrow{h_{1}^{'}}, \overrightarrow{h_{2}^{'}}, \overrightarrow{h_{3}^{'}}, ... ,\overrightarrow{h_{N}^{'}} \right \}$ with a potentially different cardinality $F^{'}$. This embeddings are combination of small bottom-up subsystems. First, $N_{h}$ heads with $N_{i}$ \textbf{bottom-up stateless} subsystems operating in parallel to modulate all stimulus' features based on 
neighborhood. In sequence, $N_{h}$ \textbf{bottom-up stateful} subsystems transform the heads outputs into the attention masks for $h$.

Each \textbf{bottom-up stateless} subsystem takes as input a focus target $\tau_{t} = \left \{ \tau_{t}^{1} \right \} = \left \{ \tau_{t,1}^{1}, \tau_{t,2}^{1} \right \} = \left \{ W\overrightarrow{h_{i}}, W\overrightarrow{h_{j}} \right \}$, $\tau_{t} \in \mathbb{R}^{2F}$. An alignment function $a$ generates the attentional weights for each input feature, ie, $a_{t} = \left \{ a_{t}^{1} \right \} = \left \{ a_{t,1}^{1}, a_{t,2}^{1} \right \} = \left \{ a^{T} \right \}$, where $a_{t} \in \mathbb{R}^{2F}$, and $a$ is a single-layer feedforward neural network, parametrized by a weight vector $a^{T}$. Each subsystem has the same selection characteristics -- \textbf{divided}, \textbf{cognitive}, \textbf{feature-based}, and \textbf{soft}. The result of attention mask over $\tau_{t}$ is contextual input $c_{t} = \left \{ c_{t}^{1} \right \} = \left \{ c_{t,1}^{1}, ..., c_{t,N_{i}}^{1} \right \}$ for \textbf{bottom-up stateful} attention subsystems: 

\begin{equation}
    c_{t,j}^{1} = \overrightarrow{a^{T}}[W\overrightarrow{h_{i}} || W\overrightarrow{h_{j}}]
    \label{graph_laye_att}
\end{equation}

\noindent where $.^{T}$ represents transposition and $\left |  \right |$ is the concatenation operation.

The \textbf{bottom-up stateful} attention subsystems based on heads output, and focus target $\tau_{t} = \left \{ \tau_{t}^{1} \right \} = \left \{ \tau_{t,1}^{1}, ..., \tau_{t,N_{i}}^{1} \right \} = \left \{ W\overrightarrow{h_{i}}, W\overrightarrow{h_{N_{i}}} \right \}$ generates attention mask $a_{t} = \left \{ a_{t}^{1} \right \} = \left \{ a_{t,1}^{1}, ...,  a_{t,N_{i}}^{1} \right \} = \left \{ a_{i,1}, ..., a_{i,N_{i}} \right \}$, where $a_{t} \in \mathbb{R}^{N_{i}}$. These subsystems apply, rectification and normalization layers over $c_{t}$ to produce \textbf{divided}, \textbf{cognitive}, \textbf{location-based}, and \textbf{soft selection} over all nodes. This subsystems' objective is to generate a embedding for each node $i$ resulting from a combination of the all neighborhood, as follows  

\begin{equation}
    \alpha_{i,j} = LeakyReLU(c_{t})
    \label{graph_laye_att_2}
\end{equation}

\begin{equation}
    a_{i,j} = \frac{\alpha_{i,j}}{\sum_{k \in N_{i}} \alpha_{i,j}}
    \label{graph_laye_norm}
\end{equation}

\begin{equation}
    \overrightarrow{h_{i}^{'}} = \sigma(\frac{1}{N_{h}}\sum_{k=1}^{N_{h}}\sum_{j \in N_{i}} a_{i,j}^{k}W^{k}\overrightarrow{h_{j}})
    \label{graph_laye_output_2}
\end{equation}

\noindent where $N_{h}$ is amount of heads, $N_{i}$ is the set of nodes belonging to a certain neighborhood of the $i$ node, and LeakyReLU is non-linear function with negative input slope $\alpha=0.2$. $\overrightarrow {h_{i}^{'}} \in \mathbb{R}^{F^{'}}$ is the output from the GAT layer to node $i$.

\section{Discussion}
\label{sec:trends}

In this section, we discuss outstanding issues with attention models around some main topics.

\subsection{Perception, multimodality, and sensors sharing}

Historically, several aspects and characteristics of attention have been mapped by observing external stimuli' selection, resulting in a wide range of theories to address attention's perceptual selection. Some researchers consider attention as a fundamental process for the existence of perception. In classical theories, sensory memories store information about the input stimuli for a period. 


The stimuli have activation and inhibition dynamics within the system. Such dynamics are entirely dependent on exposure to the same input stimulus (i.e., inhibition and return), and the construction of the internal representation of the world occurs sequentially through the exploration of stimuli that fire with greater intensity, guided by top-down and bottom-up influences. However, few attentional systems in Deep Learning have been explored from this perspective. In CNNs, the main focus of the systems is feature maps instead of the raw input image. In RNNs and generative models, the models are mainly cognitive selection over previous memories. Usually, the few existing perceptual selection systems present all the input stimuli to the network. In parallel, computation is made over the data in one pass, without the sequential dynamics and without considering the same stimulus's activation degree depending on the time exposure.


Similarly, attentional systems also explore few aspects of multimodality. Multiple inter and intra modal attentional stages for sensory fusion are still poorly explored. Typically, most of the current systems relate two different modalities without sensory fusion. One modality is only supported for the selection of stimuli of the other already in latent space. However, classical theories indicate that many areas of the brain share the processing of information from different senses. Some evidence indicates that large parts of the visual cortex are multisensory, with different fusion stages ranging from perception to higher levels of processing.

\subsection{Attention over data}


Most attentional systems occur over the data that flows through neural networks. Systems are almost exclusively location-based, disregarding features and object properties. However, this approach does not fully corroborate theories and cognitive models of the 90s. The theory of integration of characteristics by Treisman \cite{treisman1980feature} points to the initial stages of perception as purely feature-based, to the point where the merging of the different features occurs, and a WTA process determines the location of the most prominent stimulus by inhibiting the missing stimuli. Backer et al. \cite{backer2001data} presented a computational model in three stages, the first feature-based stage, the second location normally detecting four protruding regions, and finally, an object-based inhibition of return. Besides, there is strong experimental evidence showing that these properties are not multi-exclusive, present mainly in the human view \cite{frintrop2010computational}. We believe that attentional systems with all three aspects can greatly benefit computer vision applications.

\subsection{Attention over program}


Most attentional systems are oriented towards data selection. So far, task-oriented or time-oriented systems do not exist. Task-oriented structures are important to increase the modularity of classical neural networks and facilitate interpretability, given that the behavior of the modules can be indirectly investigated through the decisions made by the attentional system. Similarly, time-oriented structures are important for the consistent distribution of computation times between different structures. Such a structure is particularly interesting for deciding between the exposure time of stimuli in perceptual stages.

\subsection{Dual-Process Theory: Connectionism versus Symbolism}

According to the dual-process theory \cite{kahneman2003maps}, human reasoning has two distinct systems. The first is unconscious, fast, intuitive, automatic, non-linguistic, and deals only with implicit knowledge. The second is conscious, evolutionarily recent, exclusive to humans, slow, conscious, linguistic, algorithmic, incorporating rules-based reasoning and explicit knowledge forms. Semantic variables in conscious thinking are often causal, controllable, and relate to thoughts so that concepts can be recombined to form new and unknown concepts. For years, in AI, these two types of reasoning have been studied separately, creating two distinct research branches: connectionist models represented mainly by neural networks and symbolic models represented by highly declarative and recursive algorithms. Although different, these two systems are complementary in complex reasoning. Currently, the fusion between the symbolic and the connectionist (i.e., neuro-symbolic approaches) is one of the main AI challenges for the coming years. In this context, attention is a key element as an interface between the two types of representation. One of the main issues to be solved by attention is how to link two systems with such different representations of knowledge. Few systems have acted in this perspective. Some initiatives have emerged through external memory networks \cite{weston_2014_memory} \cite{graves_neural_2014}, in RL \cite{mnih2016strategic} and recently, in Hypergraph Attention Networks \cite{kim2020hypergraph}, in which attentional systems reason about symbolic graphs in a same semantic space and build representative embeddings for a connectionist model to make the prediction.

\subsection{Human Memories}


One of the main questions of connectionist models is how to represent, organize and manage knowledge external to the prediction structure to guarantee reusability of the knowledge acquired previously. In psychology and neuroscience, the relationship between attention and memory is intrinsic and fundamental to our cognitive system. However, in Deep Learning, most attentional systems focus on selecting and representing data from the current input. Few systems are directed to external memories or internal storage structures with specific characteristics. Research, still in its initial stages, focuses on the management of working memories \cite{graves_neural_2014} or episodic \cite{xiong_dynamic_2016}, with the absence of sensory, motor, procedural, and semantic memories.

\subsection{Hybrid Models}

Most attentional systems receive only top-down influences, coming from memories of the previous internal states of neural networks. High-level contextual elements such as rewards, motivation, emotional and sensory status are still absent. Besides, few systems have bottom-up influences, although there are numerous theories and models discussed in the classical literature, mainly focused on sensory perception. We believe that bottom-up attention can bring significant learning benefits in the early stages of perception or unsupervised/self-supervised approaches. Likewise, the area still lacks hybrid neural attention models that address both levels of attention. Only Neural Transformer \cite{vaswani_attention_2017}, and recently BRIMs \cite{mittal2020learning} have brought some significant insights to the area. However, a major research question still little explored is how to iterate between the two levels of attention and how to associate the two modulations in attentional processes.

According to some neurophysiological evidence these mechanisms are concentrated in different brain areas, but they interact with each other constantly and are simultaneously present from the initial stages of perception. At some stages, the bottom-up mechanisms can completely overlap top-down influences in a competitive process, or both can be aligned, and their effects are combined in some way. For example, when reading this article, if your senses notice an imminent danger, you will stop reading for safety, where bottom-up influences have completely suppressed top-down influences. On the other hand, you may be invited to search for a specific target in an image, and discrepant features not related to the target can interfere with the visual search process but do not fully influence the point of stopping the search for the desired target. In contrast, there is still little iteration between these mechanisms in neural networks so that they act separately at different time stages. At Neural Transformer, for example, only bottom-up attention acts on perception in the first stage of time. The perceptual stage is over, and the top-down attention helps generate the translated words, with still weak iterations between the two mechanisms.
\section{Conclusions}
\label{sec:conclusion}

In this survey, we presented a review of attention in Deep Learning from a theoretical point of view. We propose a general framework of attention and taxonomy based on theoretical concepts that date back to the pre-Deep Learning era. From a set of more than 650 papers critically analyzed via systematic review, we identified the main neural attention models in the literature, discussed, and classified their main aspects using the different perspectives of attention presented in our taxonomy. We have identified models that corroborate classical theories and biological evidence discussed for years by psychologists and neuroscientists. Finally, we present the key developments in the area in detail, formulating and explaining their attention systems from our framework and taxonomy perspective. Finally, we present a critical discussion of the models analyzed, presenting some relevant research opportunities. We hope that this survey will provide a better understanding of how attentional mechanisms work in neural networks from the theoretical point of view of care and that our taxonomy will help systematize the mechanisms facilitating their understanding and provide a better understanding of different directions of attention helping to guide the future development of the area.

\section*{Acknowledgments}

The research was supported by the Coordination for the Improvement of Higher Education Personnel (CAPES). This work was carried out within the scope of PPI-Softex with support from the MCTI, through the Technical Cooperation Agreement [01245.013778/2020-21].

%
\bibliographystyle{IEEEtran}
\bibliography{ref_survey_new_new}

\end{document}